\pdfoutput=1
\PassOptionsToPackage{table}{xcolor}
\documentclass[10pt,twocolumn,letterpaper]{article}
\def\SUBMISSION{}  %
\usepackage[pagenumbers]{cvpr}

\usepackage{cuted}    %
\usepackage{currfile} %
\usepackage{caption}  %
\usepackage{graphicx}
\usepackage{amsmath,amssymb}
\usepackage{booktabs,multirow,tabularx,array}
\usepackage{xspace}
\usepackage{enumitem}
\usepackage{pifont}
\usepackage[normalem]{ulem} %
\usepackage{adjustbox}  %
\usepackage[most]{tcolorbox}
\usepackage{tikz}
\usetikzlibrary{arrows.meta,positioning,calc,fit,backgrounds,decorations.pathreplacing,shapes.geometric}
\usepackage{microtype}

\newif\ifdraftmarks
\ifdefined\SUBMISSION\draftmarksfalse\else\draftmarkstrue\fi  %
\definecolor{litecolor}{HTML}{1F5FAD}
\definecolor{maxacolor}{HTML}{0E7C6B}
\ifdraftmarks
  \newcommand{\TODO}[1]{\textbf{\color{red}[TODO: #1]}}
  \newcommand{\lite}[1]{#1}  %
  \newcommand{\maxa}[1]{#1}
\else
  \newcommand{\TODO}[1]{}
  \newcommand{\lite}[1]{#1}
  \newcommand{\maxa}[1]{#1}
\fi
\newcommand{\maxfinal}[1]{#1}
\newcommand{\Track}{\lite{Lite}\xspace}

\newcommand{\maxbench}{\textsc{CALVIN-MAX}\xspace}
\newcommand{\lmax}{\textsc{LIBERO-MAX}\xspace}
\newcommand{\Base}{\textsf{Base}\xspace}
\newcommand{\Pre}{\textsf{Pre}\xspace}
\newcommand{\Mid}{\textsf{Mid}\xspace}
\newcommand{\dtot}{\Delta_{\text{total}}}
\newcommand{\dconf}{\Delta_{\text{config}}}
\newcommand{\dtim}{\Delta_{\text{timing}}}

\newcommand{\cmark}{\textcolor{black!80}{\ding{51}}}
\newcommand{\xmark}{\textcolor{black!35}{\ding{55}}}
\newcommand{\pmark}{\textcolor{black!55}{$\circ$}}
\newcolumntype{Y}{>{\raggedright\arraybackslash}X}
\newcolumntype{Z}{>{\centering\arraybackslash}X}

\definecolor{perc}{HTML}{2A8F8A}
\definecolor{phys}{HTML}{D08A1E}
\definecolor{emb}{HTML}{7B5EA7}
\definecolor{intent}{HTML}{6B8A2E}
\definecolor{evgold}{HTML}{F2B705}
\newcommand{\tci}[3]{\makebox[2.45em][r]{#1}\;{\color{black!55}[\makebox[2.6em][r]{#2}, \makebox[2.6em][r]{#3}]}}
\newcommand{\tcis}[3]{\makebox[1.5em][r]{#1}\;{\color{black!55}[\makebox[1.75em][r]{#2}, \makebox[1.75em][r]{#3}]}}
\newcommand{\tciint}[2]{{\color{black!55}[\makebox[2.6em][r]{#1}, \makebox[2.6em][r]{#2}]}}
\newcommand{\trange}[2]{\makebox[2.6em][r]{#1} to \makebox[2.6em][r]{#2}}

\newcommand{\evse}[2]{\makebox[2.3em][r]{#1}\,{\scriptsize\color{black!55}$\pm$\makebox[1.2em][r]{#2}}}
\newcommand{\evhead}[2]{\begin{tabular}[b]{@{}c@{}}\bfseries #1\\\bfseries #2\end{tabular}}
\definecolor{armpre}{HTML}{C8553D}
\definecolor{armmid}{HTML}{2F5F8F}
\definecolor{inkgray}{HTML}{52514E}
\definecolor{rowgray}{HTML}{F3F3F1}
\newcommand{\rowgray}{\rowcolor{rowgray}}
\newcommand{\tableheader}{\bfseries}

\definecolor{takeaway}{HTML}{2F5F8F}
\newtcolorbox{takeawaybox}[1]{
  enhanced, breakable=false,
  title={#1}, fonttitle=\bfseries\small, coltitle=black,
  colback=white, colframe=takeaway!45, colbacktitle=takeaway!9,
  titlerule=0.4pt, titlerule style={takeaway!35},
  boxrule=0.6pt, arc=1mm, left=4pt, right=4pt, top=2.5pt, bottom=2.5pt,
  toptitle=1.5pt, bottomtitle=1.5pt, before skip=6pt, after skip=6pt, fontupper=\small}

\renewcommand{\paragraph}[1]{\vspace{.25em}\noindent\textbf{#1}\hspace{0.45em}}

\newcommand{\NPolWord}{\maxfinal{ten}\xspace}
\newcommand{\PoolPerc}{\maxfinal{+5.4}\xspace}
\newcommand{\PoolPercCI}{\maxfinal{[+4.3, +6.4]}\xspace}
\newcommand{\PoolPhys}{\maxfinal{$-$0.6}\xspace}
\newcommand{\PoolPhysCI}{\maxfinal{[$-$2.2, +0.9]}\xspace}
\newcommand{\PoolInter}{\maxfinal{+6.0}\xspace}
\newcommand{\PoolInterCI}{\maxfinal{[+4.1, +7.9]}\xspace}
\newcommand{\InterMin}{\maxfinal{+1.3}\xspace}
\newcommand{\InterMax}{\maxfinal{+12.8}\xspace}
\newcommand{\PercMin}{\maxfinal{+1.6}\xspace}
\newcommand{\PercMax}{\maxfinal{+7.8}\xspace}

\newcommand{\DTotMin}{\maxfinal{$-$9.9}\xspace}
\newcommand{\DTotMax}{\maxfinal{$-$4.4}\xspace}
\newcommand{\DConfMin}{\maxfinal{$-$13.1}\xspace}
\newcommand{\DConfMax}{\maxfinal{$-$6.0}\xspace}

\newcommand{\PoolPercMag}{\maxfinal{5.4}\xspace}
\newcommand{\DTotLoMag}{\maxfinal{4.4}\xspace}
\newcommand{\DTotHiMag}{\maxfinal{9.9}\xspace}
\newcommand{\AftLoMag}{\maxfinal{5.0}\xspace}
\newcommand{\AftHiMag}{\maxfinal{16.8}\xspace}
\renewcommand{\Track}{\maxfinal{Max-A}\xspace}

\newcommand{\NPolMaxWord}{\maxfinal{ten}\xspace}

\newcommand{\GainMin}{\maxfinal{+3.0}\xspace}
\newcommand{\GainMax}{\maxfinal{+20.4}\xspace}

\newcommand{\PushTot}{\maxfinal{$-$4.2}\xspace}
\newcommand{\PushTotCI}{\maxfinal{[$-$6.5, $-$2.2]}\xspace}
\newcommand{\PushTim}{\maxfinal{$-$0.3}\xspace}
\newcommand{\PushTimCI}{\maxfinal{[$-$3.4, +2.8]}\xspace}
\newcommand{\GraspTot}{\maxfinal{$-$1.6}\xspace}
\newcommand{\GraspTotCI}{\maxfinal{[$-$2.9, $-$0.4]}\xspace}



\usepackage{placeins}

\definecolor{cvprblue}{rgb}{0.21,0.49,0.74}
\usepackage[pagebackref,breaklinks,colorlinks,allcolors=cvprblue]{hyperref}
\hypersetup{pdftitle={CALVIN-MAX: When Does a Mid-Execution Change Hurt a Robot Policy?},
  pdfauthor={Sajjad Pakdamansavoji}}
\usepackage[capitalize]{cleveref}
\crefname{section}{Sec.}{Secs.}
\Crefname{section}{Section}{Sections}
\crefname{table}{Tab.}{Tabs.}
\Crefname{table}{Table}{Tables}

\def\confName{CVPR}
\def\confYear{2027}

\title{\maxbench: When Does a Mid-Execution Change Hurt a Robot Policy?}

\author{Sajjad Pakdamansavoji}

\makeatletter
\let\kit@maketitle\@maketitle
\def\@maketitle{%
  \setbox\z@\vbox{\def\@author{Anonymous \confName~submission}\kit@maketitle}%
  \@tempdima\dimexpr\ht\z@+\dp\z@\relax
  \setbox\tw@\vbox{%
    \centering
    \vskip 6pt
    {\Large\bf\@title\par}%
    \vskip 18pt
    {\large\@author\\Forall Labs\par}%
    \vskip 3pt
    {\small\tt sj.pakdaman.edu@gmail.com\\
      \href{https://forall-labs.github.io/CALVIN-MAX/}{https://forall-labs.github.io/CALVIN-MAX}\par}%
  }%
  \typeout{arXiv title block: content \the\dimexpr\ht\tw@+\dp\tw@\relax, review block \the\@tempdima,
    space before the teaser \the\dimexpr\@tempdima-\ht\tw@-\dp\tw@\relax}%
  \ifdim\dimexpr\ht\tw@+\dp\tw@\relax>\@tempdima
    \PackageError{arxiv}{Title block taller than the review version's}{}\fi
  \vbox to\@tempdima{\unvbox\tw@\vfil}}
\makeatother

\begin{document}
\maketitle
\begin{strip}
\def\TzEvent{92.7/84.6/88.2, 85.1/77.3/79.6, 81.8/69.6/73.2, 79.3/71.2/73.6, 75.0/69.0/70.0, 71.5/59.8/61.7, 66.5/53.4/56.6, 57.2/45.2/47.5, 51.4/44.7/47.0, 38.2/30.6/32.1}
\def\TzTiming{5.1/2.3, 6.8/1.0, 1.6/0.3, 5.0/-2.9, 5.6/1.0, 5.1/-1.0, 7.8/-5.0, 6.6/1.0, 4.4/-1.9, 5.5/-1.0}
\def\TzPoolPerc{5.4/4.3/6.4}
\def\TzPoolPhys{-0.6/-2.2/0.9}
\def\TzNext{94.8/88.1, 93.6/88.6, 92.0/83.2, 91.7/85.3, 90.0/83.0, 89.0/72.2, 85.3/74.5, 77.5/64.9, 76.2/69.1, 61.0/54.2}
\def\TzRewind{4.1/-1.5, 3.9/0.7, 6.2/0.0, 0.2/-2.8, 5.4/-1.2, 3.2/-0.8, 4.4/-4.8, 8.1/-12.4, 5.6/-2.6, 3.8/-2.7}
\def\TzRewindMean{-2.8}
\def\TzRewindMeanLab{$-$2.8}
\def\TzRewindPool{-2.8/-3.9/-1.8}
\def\TzRewindPoolLab{$-$2.8}
\def\TzRewindTimPool{4.5/3.4/5.5}
\def\TzRewindTimPoolLab{+4.5}

\centering
\vspace{-3mm}
\resizebox{\linewidth}{!}{%
\begin{tikzpicture}[
  font=\sffamily\footnotesize,
  >=Stealth,
  fam/.style={draw=#1!55, fill=white, rounded corners=2.5pt, line width=0.6pt, anchor=north west, inner sep=0pt},
  famhead/.style={fill=#1!13, draw=none, rounded corners=2.5pt, anchor=north west, inner sep=0pt, minimum height=5mm},
  famname/.style={anchor=west, font=\sffamily\footnotesize\bfseries, text=#1!70!black},
  famcount/.style={anchor=east, font=\sffamily\scriptsize, text=#1!70!black},
  lab/.style={font=\sffamily\fontsize{6}{7}\selectfont, anchor=north, inner sep=0pt, text=black!80},
  col/.style={draw=none, fill=white, rounded corners=2.5pt, line width=0.6pt, anchor=north west, inner sep=0pt},
  colhead/.style={anchor=north west, font=\sffamily\footnotesize\bfseries, inner sep=0pt},
  colsub/.style={anchor=north west, font=\sffamily\scriptsize\itshape, text=black!60, inner sep=0pt},
  ev/.style={star, star points=6, star point ratio=2.1, fill=evgold, draw=evgold!60!black, line width=0.3pt, inner sep=1.15pt},
]
\def\s{17.5mm}  %
\def\g{1.2mm}   %
\def\p{1.4mm}   %
\def\lh{3.3mm}  %
\newcommand{\thumb}[6]{%
  \node[anchor=north west, inner sep=0pt] (#1) at (#3,#4) {\includegraphics[width=\s]{figures/gallery/#2.jpg}};
  \draw[#6, line width=0.5pt] (#1.south west) rectangle (#1.north east);
  \node[lab] at ($(#1.south)+(0,-0.7mm)$) {#5};}
\newcommand{\famhead}[5]{%
  \node[famhead=#2, minimum width=#5-0.6mm] (#1h) at ($(#1.north west)+(0.3mm,-0.3mm)$) {};
  \node[famname=#2] at ($(#1h.west)+(1.4mm,0)$) {#3};
  \node[famcount=#2] at ($(#1h.east)+(-1.4mm,0)$) {#4};}

\pgfmathsetlengthmacro\wperc{3*\s+2*\g+2*\p}
\pgfmathsetlengthmacro\wphys{2*\s+\g+2*\p}
\pgfmathsetlengthmacro\hA{5.6mm+2*\s+2*\lh+\g+0.6mm}
\pgfmathsetlengthmacro\hB{5.6mm+\s+\lh+0.8mm}
\node[fam=perc, minimum width=\wperc, minimum height=\hA] (P) at (0,0) {};
\famhead{P}{perc}{Perceptual}{5 changes}{\wperc}
\pgfmathsetlengthmacro\xa{\p}\pgfmathsetlengthmacro\xb{\p+\s+\g}\pgfmathsetlengthmacro\xc{\p+2*\s+2*\g}
\pgfmathsetlengthmacro\ya{-5.6mm}\pgfmathsetlengthmacro\yb{-5.6mm-\s-\lh-\g}
\thumb{po}{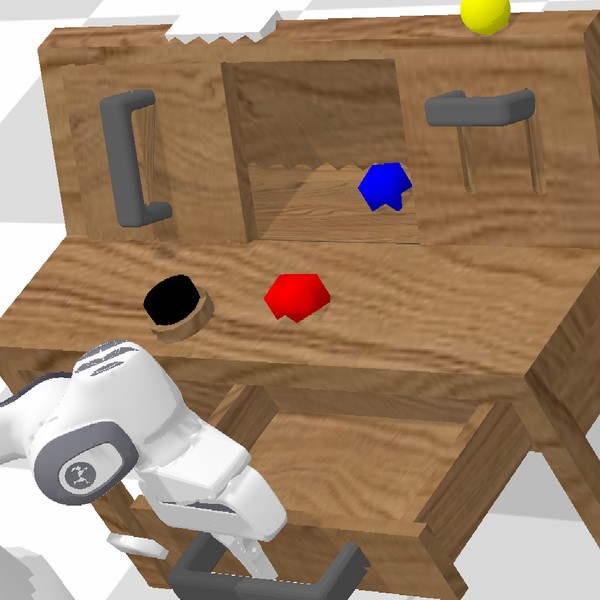}{\xa}{\ya}{\textbf{original}}{black!70}
\thumb{p1}{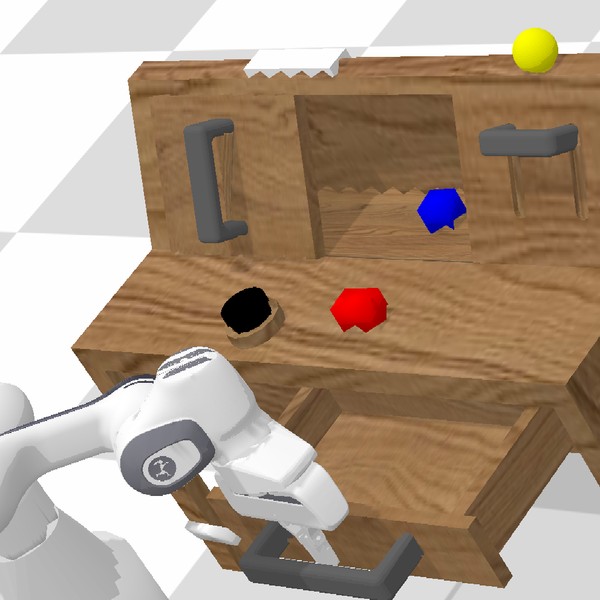}{\xb}{\ya}{camera shift}{black!15}
\thumb{p2}{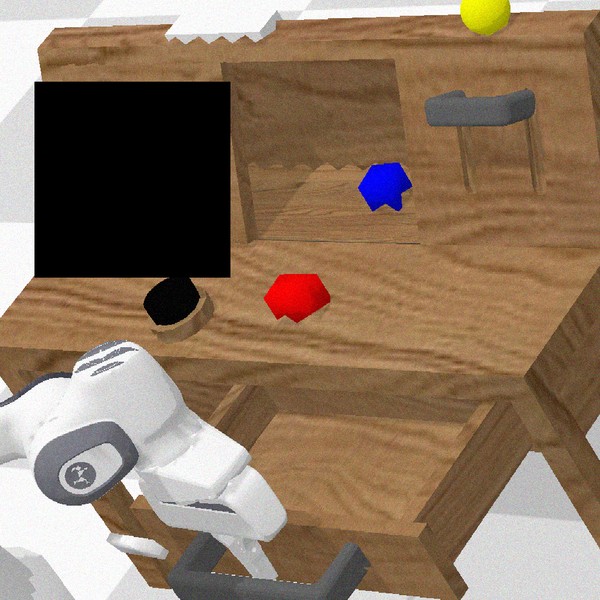}{\xc}{\ya}{sensor noise}{black!15}
\thumb{p3}{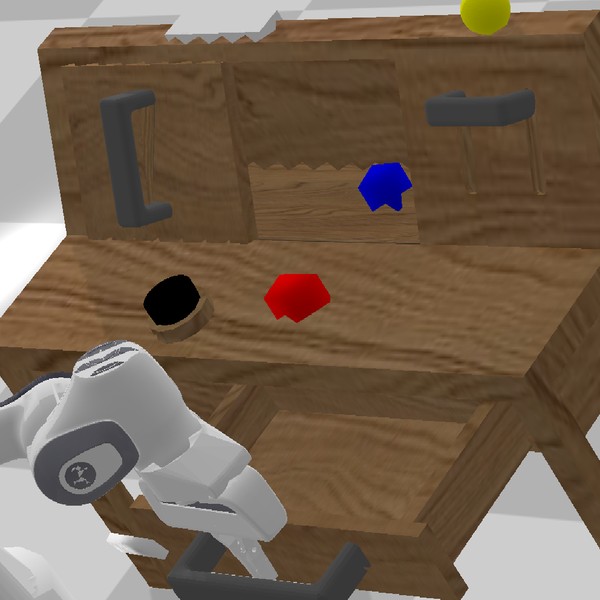}{\xa}{\yb}{illumination}{black!15}
\thumb{p4}{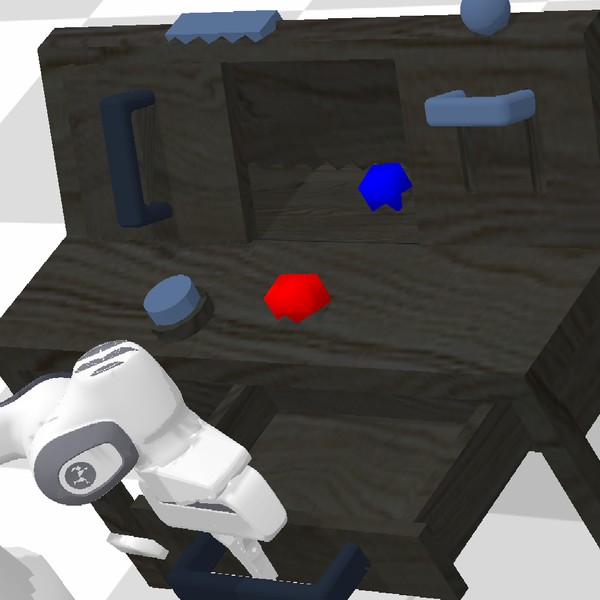}{\xb}{\yb}{visual theme}{black!15}
\thumb{p5}{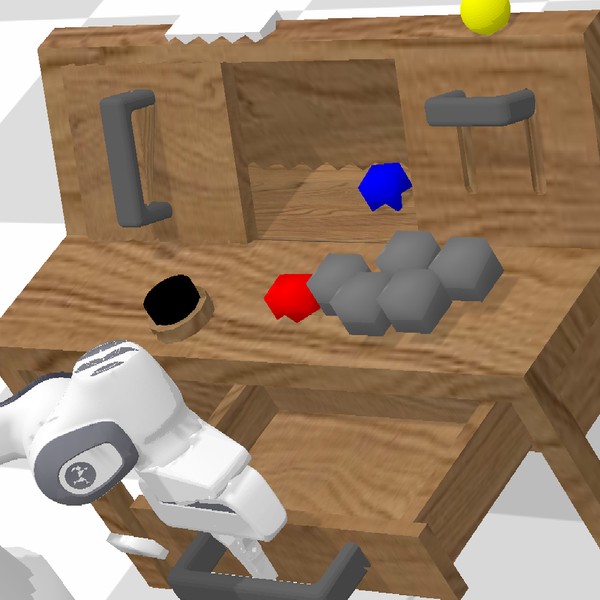}{\xc}{\yb}{distractors}{black!15}

\pgfmathsetlengthmacro\xP{\wperc+2mm}
\node[fam=phys, minimum width=\wphys, minimum height=\hA] (H) at (\xP,0) {};
\famhead{H}{phys}{Physical}{3 changes}{\wphys}
\pgfmathsetlengthmacro\xa{\xP+\p}\pgfmathsetlengthmacro\xb{\xP+\p+\s+\g}
\thumb{ho}{original}{\xa}{\ya}{\textbf{original}}{black!70}
\thumb{h1}{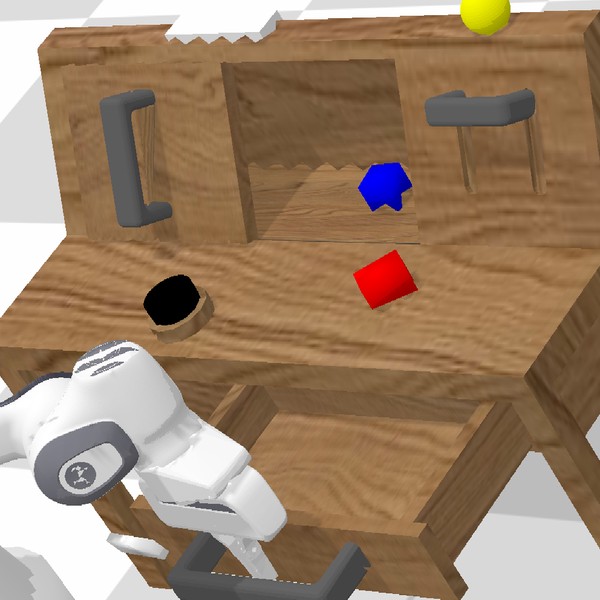}{\xb}{\ya}{target moved}{black!15}
\thumb{h2}{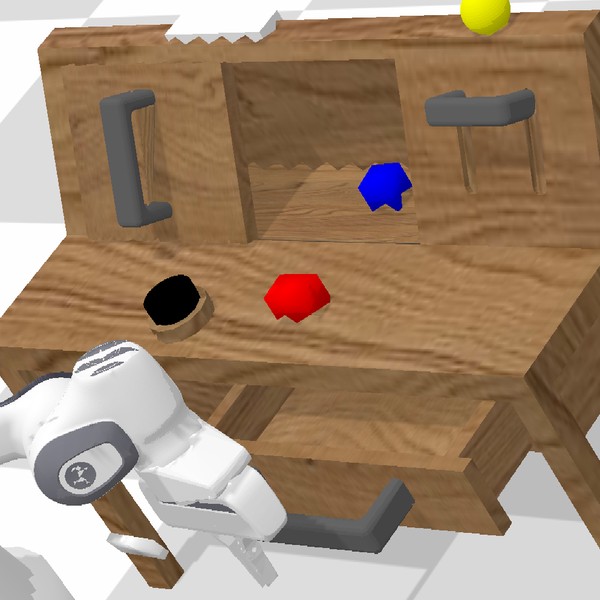}{\xa}{\yb}{receptacle moved}{black!15}
\thumb{h3}{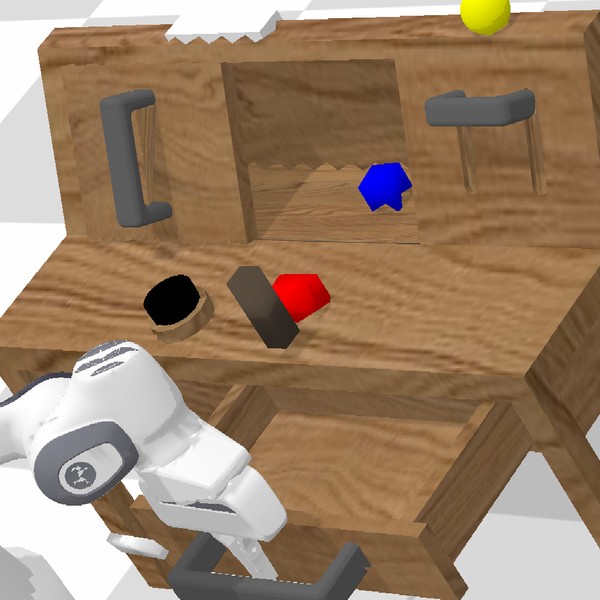}{\xb}{\yb}{obstacle inserted}{black!15}

\pgfmathsetlengthmacro\yB{-\hA-2mm}
\node[fam=emb, minimum width=\wperc, minimum height=\hB] (E) at (0,\yB) {};
\famhead{E}{emb}{Embodiment}{2 changes}{\wperc}
\pgfmathsetlengthmacro\xa{\p}\pgfmathsetlengthmacro\xb{\p+\s+\g}\pgfmathsetlengthmacro\xc{\p+2*\s+2*\g}
\pgfmathsetlengthmacro\yc{\yB-5.6mm}
\thumb{eo}{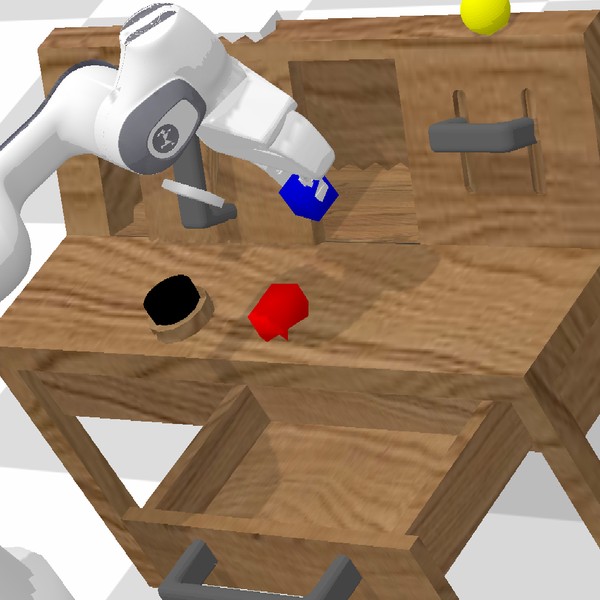}{\xa}{\yc}{\textbf{original}}{black!70}
\thumb{e1}{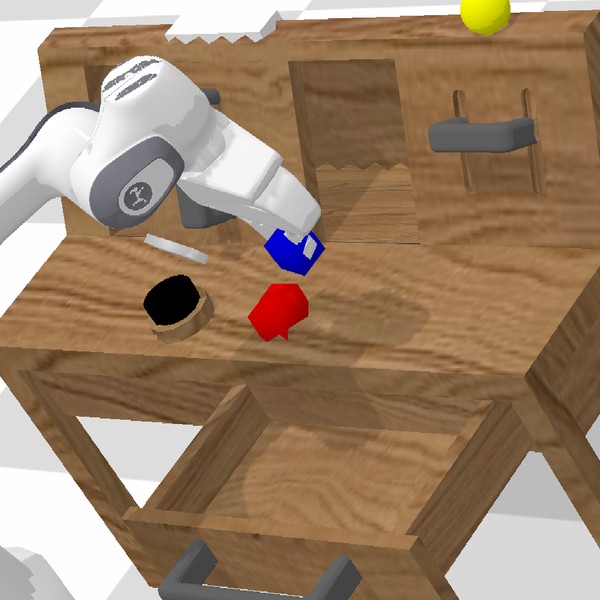}{\xb}{\yc}{arm push}{black!15}
\thumb{e2}{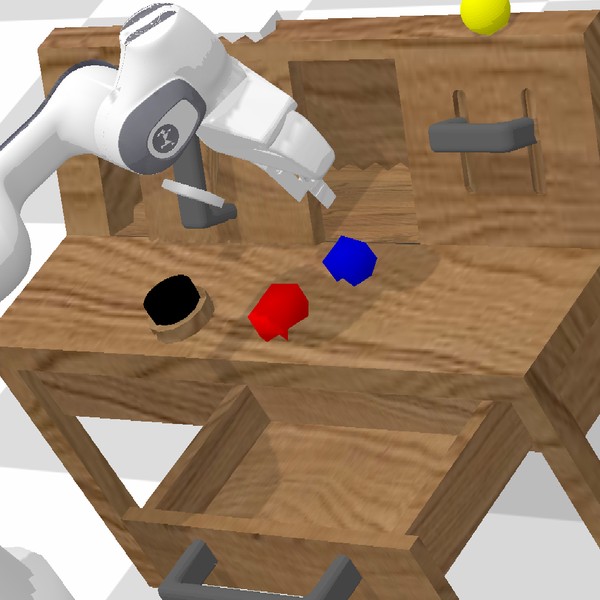}{\xc}{\yc}{grasp loss}{black!15}

\node[fam=intent, minimum width=\wphys, minimum height=\hB] (I) at (\xP,\yB) {};
\famhead{I}{intent}{Intent \& control}{3 types}{\wphys}
\pgfmathsetlengthmacro\wI{\wphys-2*\p}%
\def\isp{3.0mm}%
\def\irh{4.3mm}%
\newcommand{\irow}[7]{%
  \node[draw=#4, line width=0.5pt, fill=white, rounded corners=1pt, anchor=north west, inner sep=0pt,
        minimum width=#3, minimum height=\irh] (#1) at #2 {};
  \node[anchor=base west, inner sep=0pt, font=\sffamily\fontsize{5}{6}\selectfont] at ($(#1.north west)+(0.9mm,-1.6mm)$)
    {\textbf{#5}\hspace{0.9mm}#6};
  \node[anchor=base west, inner sep=0pt, font=\sffamily\fontsize{6}{7}\selectfont] at ($(#1.north west)+(0.9mm,-3.55mm)$)
    {``#7''};}
\irow{io}{($(I.north west)+(\p,-6.05mm)$)}{\wI}{black!70}{original}{}{grasp and lift the blue block}
\irow{i1}{($(io.south west)+(\isp,-1.2mm)$)}{\wI-\isp}{black!45}{null}{\textcolor{intent!80!black}{nothing changes}}{grasp and lift the blue block}
\irow{i2}{($(i1.south west)+(0,-0.7mm)$)}{\wI-\isp}{black!45}{transient}{\textcolor{intent!80!black}{block moved, then put back}}{grasp and lift the blue block}
\irow{i3}{($(i2.south west)+(0,-0.7mm)$)}{\wI-\isp}{black!45}{revised}{\textcolor{intent!80!black}{new goal}}{grasp and lift the red block}
\coordinate (isp) at ($(io.south west)+(1.2mm,0)$);
\foreach \t in {i1, i2, i3}
  \draw[intent!75, line width=0.6pt, rounded corners=1.2pt, -{Stealth[length=1.3mm, width=1.1mm]}] (isp) |- (\t.west);

\pgfmathsetlengthmacro\xM{\xP+\wphys+3mm}
\def\wM{37mm}
\pgfmathsetlengthmacro\hT{\hA+2mm+\hB}
\node[col, minimum width=\wM, minimum height=\hT] (M) at (\xM,0) {};
\node[colhead] at ($(M.north west)+(1.8mm,-1.6mm)$) {Three arms per case};
\node[colsub, font=\sffamily\fontsize{6}{7}\selectfont\itshape] at ($(M.north west)+(1.8mm,-5.6mm)$) {identical until the affected instruction};
\newcommand{\scene}{%
  \fill[black!6] (2,0) rectangle (35,1.4); \draw[black!45, line width=0.6pt] (2,1.4) -- (35,1.4);
  \fill[black!35, rounded corners=0.4pt] (4.6,1.4) rectangle (7.4,2.6);
  \draw[black!40, line width=2.2pt, line cap=round] (6,2.6) -- (8.6,9.6) -- (14.4,10.2);
  \fill[white] (8.6,9.6) circle (0.75); \draw[black!40, line width=0.5pt] (8.6,9.6) circle (0.75);
  \draw[black!45, line width=0.9pt] (13.4,9.4) -- (15.4,9.4) (13.6,9.4) -- (13.6,8.2) (15.2,9.4) -- (15.2,8.2);
  \draw[black!45, line width=0.9pt] (14.4,9.4) -- (14.4,10.2);}
\newcommand{\blockat}[2]{\fill[#2, rounded corners=0.3pt] (#1-1.1,1.4) rectangle (#1+1.1,3.6);}
\newcommand{\ghostat}[1]{\draw[black!45, dash pattern=on 0.8pt off 0.6pt, line width=0.5pt, rounded corners=0.3pt] (#1-1.1,1.4) rectangle (#1+1.1,3.6);}
\newcommand{\armlabel}[3]{%
  \node[anchor=north west, font=\sffamily\scriptsize\bfseries, text=#2, inner sep=0pt] at (2,14.4) {#1};
  \node[anchor=north east, font=\sffamily\fontsize{6}{7}\selectfont, text=black!65, inner sep=0pt, align=right] at (35,14.2) {#3};}
\begin{scope}[x=1mm, y=1mm]
  \begin{scope}[shift={($(M.north west)+(0,-26.5mm)$)}]
    \scene \blockat{22.5}{black!75}
    \draw[->, line width=1pt, black!80] (14.4,7.9) .. controls (16.5,7.6) and (21.5,7.2) .. (22.5,4.3);
    \armlabel{\Base}{black}{no change}
  \end{scope}
  \begin{scope}[shift={($(M.north west)+(0,-45mm)$)}]
    \scene \ghostat{22.5} \blockat{29.5}{black!75}
    \draw[->, line width=1pt, armpre] (14.4,7.9) .. controls (18,7.8) and (28.5,7.4) .. (29.5,4.3);
    \node[ev] at (14.4,7.9) {};
    \armlabel{\Pre}{armpre}{change in place\\from the start}
  \end{scope}
  \begin{scope}[shift={($(M.north west)+(0,-63.5mm)$)}]
    \scene \ghostat{22.5} \blockat{29.5}{black!75}
    \draw[line width=1pt, black!30] (14.4,7.9) .. controls (15.6,7.75) and (17.4,7.6) .. (18.6,7.1);
    \draw[->, line width=1pt, armmid] (18.6,7.1) .. controls (21,6.2) and (28.6,7.6) .. (29.5,4.3);
    \draw[black!30, dash pattern=on 1pt off 0.8pt, line width=0.6pt] (18.6,7.1) .. controls (20.6,6.3) and (22.2,5.8) .. (22.5,4.3);
    \node[ev] at (18.6,7.1) {};
    \armlabel{\Mid}{armmid}{change arrives\\while moving}
  \end{scope}
\end{scope}
\node[anchor=south west, font=\sffamily\fontsize{5.6}{7.2}\selectfont, align=left, inner sep=0pt, text width=\wM-2.6mm]
  at ($(M.south west)+(1.8mm,0.9mm)$)%
  {\tikz[baseline=-0.6ex]\node[ev]{}; change (here: target block moved)\\[1.5pt]
   \textcolor{armpre}{\textbf{\Pre}}${-}$\Base: cost of the changed world\\
   \textcolor{armmid}{\textbf{\Mid}}${-}$\textcolor{armpre}{\textbf{\Pre}}: effect of the timing};

\pgfmathsetlengthmacro\xR{\xM+\wM+3mm}
\def\wR{38mm}
\node[col, minimum width=\wR, minimum height=\hT] (R) at (\xR,0) {};
\foreach \xs in {\xM-1.5mm, \xR-1.5mm}
  \draw[black!70, line width=0.6pt, dash pattern=on 2.2pt off 1.6pt] (\xs,0) -- (\xs,-\hT);
\node[colhead] at ($(R.north west)+(1.8mm,-1.6mm)$) {Main results};
\tikzset{
  ptitle/.style={anchor=north west, font=\sffamily\fontsize{6}{7}\selectfont\bfseries, inner sep=0pt},
  tick/.style={anchor=north, font=\sffamily\fontsize{5}{6}\selectfont, text=black!55, inner sep=0pt},
  note/.style={font=\sffamily\fontsize{5.5}{6}\selectfont\bfseries, text=black!55, inner sep=0pt},
  dB/.style={draw=black!65, fill=white, line width=0.45pt},
  dP/.style={fill=armpre}, dM/.style={fill=armmid},
  rule/.style={black!15, line width=0.4pt},
  ylab/.style={rotate=90, anchor=south, font=\sffamily\fontsize{5.5}{6}\selectfont, text=black!60, inner sep=0pt}}
\def\rr{0.48mm}  %
\newcommand{\legdot}[1]{\tikz[baseline=-0.55ex]\fill[#1] (0,0) circle (0.5mm);}
\newcommand{\legopen}{\tikz[baseline=-0.55ex]\draw[dB] (0,0) circle (0.45mm);}
\newcommand{\pbar}[3]{\fill[#3] (0,#1-1.6) rectangle (#2,#1+1.6);}  %
\newcommand{\pci}[4]{\draw[#4, line width=0.6pt] (#2,#1) -- (#3,#1) (#2,#1-0.8) -- (#2,#1+0.8) (#3,#1-0.8) -- (#3,#1+0.8);}
\newcommand{\plab}[3]{\node[anchor=east, font=\sffamily\fontsize{6}{6.6}\selectfont, text=#3, inner sep=0pt, align=right] at (-6.0,#1) {#2};}
\newcommand{\pval}[4]{\node[font=\sffamily\fontsize{6}{7}\selectfont\bfseries, text=#4, inner sep=0pt, anchor=#3] at (#1) {#2};}
\newcommand{\paxes}{%
  \foreach \v in {-5,5} \draw[black!9, line width=0.3pt] (\v,-2.2) -- (\v,6.1);
  \foreach \v/\t in {-5/$-$5,0/0,5/5} \node[tick] at (\v,6.5) {\t};}
\newcommand{\pzero}{\draw[black!65, line width=0.5pt] (0,-2.2) -- (0,6.1);}
\node[ptitle] at ($(R.north west)+(1.8mm,-6.2mm)$) {Affected instruction (\%)};
\node[ylab] at ($(R.north west)+(3.0mm,-10.6mm-6.3mm)$) {10 policies};
\begin{scope}[shift={($(R.north west)+(4.6mm,-10.6mm)$)}, x=0.44mm, y=-1.4mm]
  \foreach \v in {40,60,80,100} {\draw[black!9, line width=0.3pt] (\v-30,-0.7) -- (\v-30,9.7);
                                 \node[tick] at (\v-30,10.1) {\v};}
  \foreach \b/\p/\m [count=\i from 0] in \TzEvent {
    \draw[black!30, line width=0.5pt] ({\p-30},\i) -- ({\b-30},\i);
    \fill[dP] ({\p-30},\i) circle (\rr); \fill[dM] ({\m-30},\i) circle (\rr); \draw[dB] ({\b-30},\i) circle (\rr);}
  \node[anchor=south east, font=\sffamily\fontsize{5.5}{6.6}\selectfont, align=left, inner sep=0pt] at (70,9.4)
    {\legopen\,\Base\\\legdot{armpre}\,\Pre\\\legdot{armmid}\,\Mid};
\end{scope}
\draw[rule] ($(R.north west)+(1.8mm,-27.6mm)$) -- ++(\wR-3.6mm,0);
\node[ptitle] at ($(R.north west)+(1.8mm,-28.8mm)$) {Timing effect, {\Mid}$-$\Pre (pp)};
\begin{scope}[shift={($(R.north west)+(22.0mm,-33.4mm)$)}, x=1.2mm, y=-1mm]
  \paxes
  \foreach \e/\lo/\hi in \TzPoolPerc {\pbar{0}{\e}{perc!85}\pci{0}{\lo}{\hi}{perc!40!black}\pval{\hi+0.5,0}{\PoolPerc}{west}{perc!60!black}}
  \foreach \e/\lo/\hi in \TzPoolPhys {\pbar{3.9}{\e}{phys!85}\pci{3.9}{\lo}{\hi}{phys!40!black}\pval{\hi+0.5,3.9}{\PoolPhys}{west}{phys!60!black}}
  \pzero
  \plab{0}{perceptual}{perc!70!black}\plab{3.9}{physical}{phys!70!black}
\end{scope}
\draw[rule] ($(R.north west)+(1.8mm,-42.4mm)$) -- ++(\wR-3.6mm,0);
\node[ptitle] at ($(R.north west)+(1.8mm,-43.6mm)$) {Rewind check, rewindable set (pp)};
\begin{scope}[shift={($(R.north west)+(22.0mm,-48.2mm)$)}, x=1.2mm, y=-1mm]
  \paxes
  \foreach \e/\lo/\hi in \TzRewindTimPool {\pbar{0}{\e}{perc!85}\pci{0}{\lo}{\hi}{perc!40!black}\pval{\hi+0.5,0}{\TzRewindTimPoolLab}{west}{perc!60!black}}
  \foreach \e/\lo/\hi in \TzRewindPool {\pbar{3.9}{\e}{perc!85}\pci{3.9}{\lo}{\hi}{perc!40!black}\pval{0.5,3.9}{\TzRewindPoolLab}{west}{perc!60!black}}
  \pzero
  \plab{0}{{\Mid}$-$\Pre}{black!75}\plab{3.9}{rewound$-$\Pre}{black!75}
\end{scope}
\draw[rule] ($(R.north west)+(1.8mm,-57.2mm)$) -- ++(\wR-3.6mm,0);
\node[ptitle] at ($(R.north west)+(1.8mm,-58.4mm)$) {Next instruction, after success (\%)};
\node[ylab] at ($(R.north west)+(3.0mm,-62.6mm-6.3mm)$) {10 policies};
\begin{scope}[shift={($(R.north west)+(4.6mm,-62.6mm)$)}, x=0.61mm, y=-1.4mm]
  \foreach \v in {60,70,80,90,100} {\draw[black!9, line width=0.3pt] (\v-50,-0.7) -- (\v-50,9.7);
                                    \node[tick] at (\v-50,10.1) {\v};}
  \foreach \b/\m [count=\i from 0] in \TzNext {
    \draw[black!30, line width=0.5pt] ({\m-50},\i) -- ({\b-50},\i);
    \fill[dM] ({\m-50},\i) circle (\rr); \draw[dB] ({\b-50},\i) circle (\rr);}
  \node[anchor=south east, font=\sffamily\fontsize{5.5}{6.6}\selectfont, align=left, inner sep=0pt] at (50,9.4)
    {\legopen\,\Base\\\legdot{armmid}\,\Mid};
\end{scope}
\end{tikzpicture}}
\vspace{-1mm}
\captionsetup{hypcap=false}

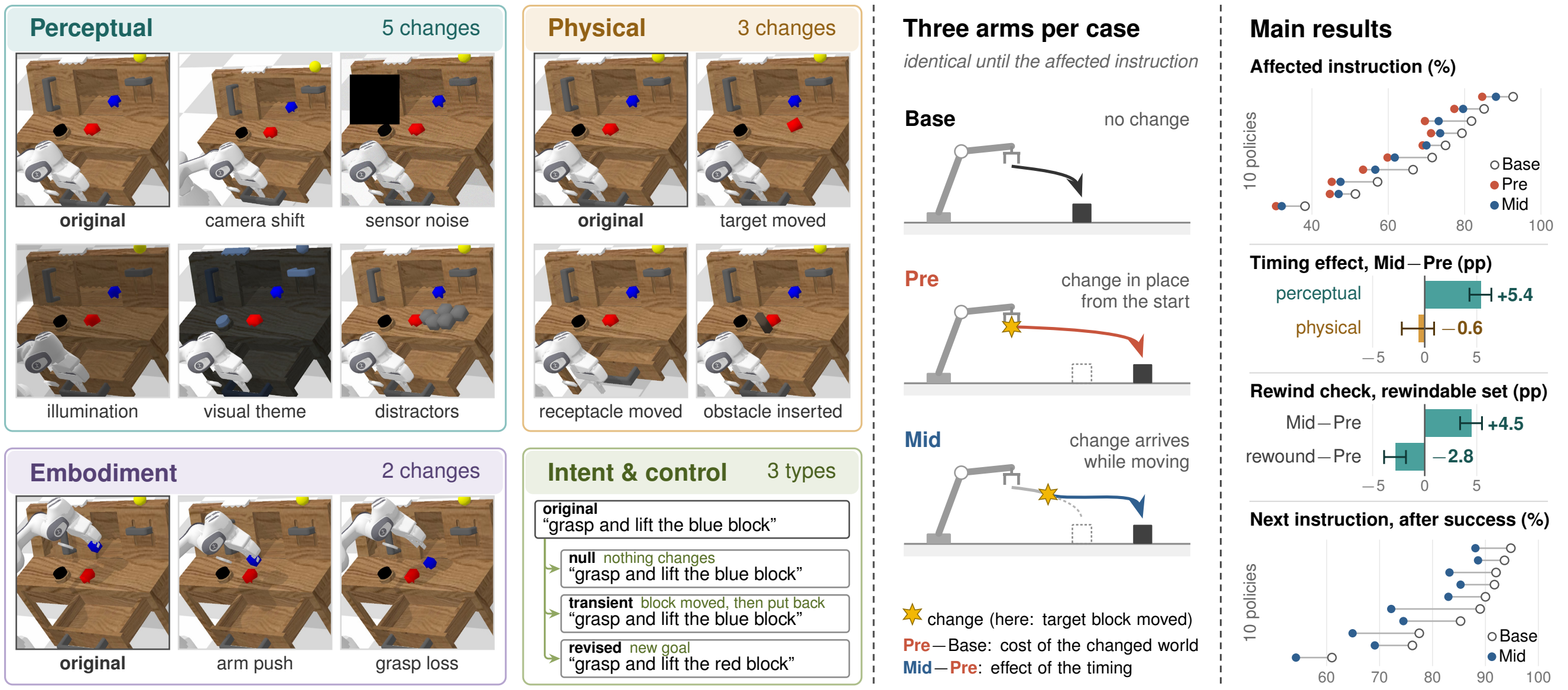
\captionof{figure}{\textbf{\maxbench asks when a change during execution hurts a policy, not only whether it does.}
On the left, each family of change is applied to the same moment of a \Base run, beside the unchanged original.
In the intent and control card only the revised instruction gives the robot a new goal, while the null and transient controls keep it, changing nothing or moving a block and putting it back.
In the middle, each case is run three times, identically until the affected instruction begins: without the change (\Base), with it in place from that moment (\Pre), and with it arriving while the robot is moving (\Mid). \Pre against \Base gives the cost of the changed world, and \Mid against \Pre the effect of its timing.
On the right, held-out results for ten policies (\cref{sec:results}) show, from top to bottom, success on the affected instruction in each run; the timing effect, positive for perceptual changes and not detectable for physical ones (which here include the changes to the robot); the perceptual timing effect on the cases where the robot can be rewound, where returning the arm and the policy's memory to the start of the instruction removes the benefit; and success on the next instruction among cases in which the affected one succeeded in both runs.}
\captionsetup{hypcap=true}
\label{fig:teaser}
\vspace{-1mm}
\end{strip}

\begin{abstract}
Robots must often keep working after the world changes, as when a block moves or the camera shifts while the robot still acts on what it saw before.
Recent work showed that one change during a task lowers success for every policy tested, but did not separate the effect of its timing from the cost of the changed world, or measure what it costs afterwards.
We introduce \maxbench, a benchmark of 7,828 cases on CALVIN's five-instruction chains, each with one change to the objects, the camera, the robot, or the instruction.
Each case compares three runs that are identical until the affected instruction begins: one without the change, one with it in place from that moment, and one in which it arrives while the robot is moving.
The second run against the first gives the cost of the changed world, and the third against the second the effect of meeting it mid-motion.
On 1,526 held-out cases per policy, with hypotheses fixed in advance, each of ten released policies loses \DTotLoMag{} to \DTotHiMag{} percentage points on the affected instruction over the eight core changes.
Whether timing matters depends on the change: one that alters what the robot sees costs \PoolPercMag{} points less mid-motion than from the start, whereas a change to the scene itself or to the robot shows no detectable effect.
Rewinding the robot to the start of the instruction removes the benefit for every policy; the stricter test we fixed in advance, that the rewound run then matches the second run, passes for three.
The cost also outlasts the affected instruction: the change stays in place, and a policy that still completes that instruction is \AftLoMag{} to \AftHiMag{} points less likely to complete the next one.
\par\vspace{-12pt}%
\end{abstract}

\section{Introduction}
\label{sec:intro}

A manipulation policy rarely acts in a world that holds still. A block may be nudged while the gripper approaches it, the camera may be bumped, the lights may change, or the arm itself may be pushed. In each case the policy has already acted on observations that no longer describe the scene. Simulation benchmarks give controlled and repeatable tests of robot policies~\cite{mees2022calvin,liu2023libero,james2020rlbench,yu2020meta,gu2023maniskill2}, and their robustness variants perturb the scene, the camera, or the instruction~\cite{fei2025liberoplus,zhou2025liberopro,pumacay2024colosseum,wang2026liberox}. Almost all of them set the perturbation before the first action. They test whether a policy can operate in a changed world, not whether it can keep going when the world changes.

\lmax~\cite{zhang2026liberomax} recently addressed this directly. It runs every task twice with identical actions up to a point just before the grasp, then changes the scene in the second run and lets the policy continue. All fourteen of its policies lose 11 to 26 points of success. This establishes \emph{that} a change during execution hurts, but not \emph{when} or \emph{why}. A change that arrives mid-motion differs from the same change present from the start in two opposing ways. The policy is committed to the old world: actions are queued, the arm is moving, and its memory holds a scene that no longer exists. Yet it has also made progress, which may carry it to the goal regardless. A comparison with an undisturbed run mixes both with the plain cost of acting in a changed scene. \lmax did compare a camera shift at reset and mid-task, and found it \emph{less} harmful mid-task, but a change at reset alters the whole trajectory, so the two runs no longer share a history. Its tasks are also single instructions, so nothing follows the change.

\begin{figure*}[t]
\centering
\includegraphics[width=\linewidth]{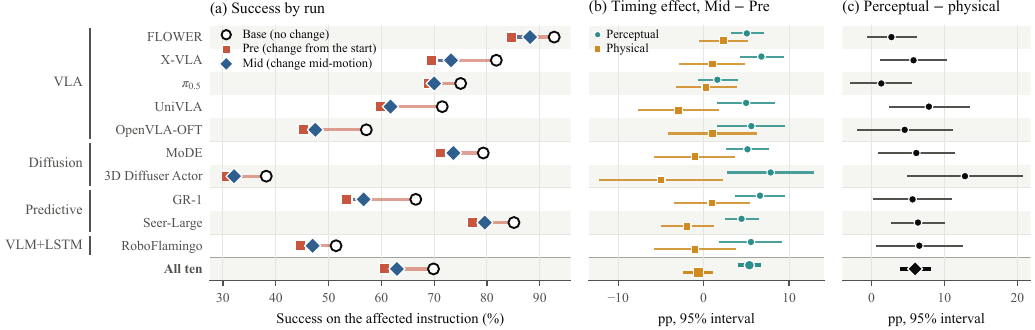}
\vspace{-5mm}
\caption{Results on the held-out cases for each policy and for all ten together. Panel (a) shows success on the affected instruction over the eight core changes and all assigned cases when nothing changes (\Base, open circles), when the change is already in place as the instruction begins (\Pre, \textcolor{armpre}{red squares}), and when it arrives while the robot is moving (\Mid, \textcolor{armmid}{blue diamonds}). Panel (b) shows the timing effect, \Mid$-$\Pre, on realized cases for \textcolor{perc}{perceptual} and \textcolor{phys}{physical} changes (the latter include the two changes to the robot), and panel (c) the gap between the two, with 95\% intervals. For a perceptual change the estimate is positive for every policy and its interval excludes zero for nine of the ten; for a physical change no policy's interval excludes zero. The gap in (c) is significant after correction for four policies.}
\label{fig:overview}
\vspace{-2mm}
\end{figure*}

We introduce \maxbench, which separates these effects with a third run. Each case is run three times, and the three runs, which we call \emph{arms}, share every action until the affected instruction begins (\cref{fig:teaser}). \Base continues undisturbed. \Mid follows \Base until the robot is moving and then receives the change, as in \lmax. \Pre receives the identical change at the moment the affected instruction begins, after the same history and from the same robot state. \Pre and \Mid thus receive the same change and differ in when it arrives: \Pre against \Base gives the cost of the changed world, and \Mid against \Pre the net effect of its timing. We build on CALVIN~\cite{mees2022calvin}, whose tasks are chains of five instructions and whose simulator reports the full state of the scene, so we can follow a change beyond the instruction it interrupts and observe how the robot responds, not only whether it succeeds. We keep \lmax's eight changes, the moment at which they arrive and its metrics, adapted to CALVIN.

\maxbench specifies 7,828 cases on CALVIN's official evaluation chains, each fixed before any policy is run. To \lmax's changes to the objects and the camera's view it adds changes to the robot (a push on the arm, a block slipping from the gripper) and to the instruction (a revised goal), and it varies when in the motion a change arrives. Smaller exploratory sets outside these cases vary how strong a change is and undo a change a moment later, and two diagnostic runs discard the policy's queued actions or rewind the robot to the start of the instruction. We evaluate \NPolWord released policies spanning vision-language-action (VLA) models, diffusion policies and models that predict from a history of frames. Our hypotheses were formed on 672 cases and frozen before we ran the 1,526 held-out cases per policy behind our headline numbers.

Every policy completes the affected instruction less often once the world changes, by \DTotLoMag{} to \DTotHiMag{} points (\cref{fig:overview}a), but whether timing matters depends on what changes. A \emph{perceptual} change, which alters what the robot sees (the camera, sensor noise, lighting, textures or distractors), costs \PoolPercMag{} points less when it arrives mid-motion than when it is in place from the start, and the estimate is positive for all \NPolWord policies. A \emph{physical} change, which moves the target or the receptacle, blocks the path or, in the grouping we fixed in advance, disturbs the robot, shows no detectable effect of timing (\PoolPhys{} \PoolPhysCI{}). The difference between the two, our primary test, is \PoolInter{} points \PoolInterCI{} (\cref{fig:overview}b,c), and it is significant for four of the policies individually. Rewinding removes the benefit: when we return the arm and the policy's memory to the start of the instruction and leave the world changed, the rewound run does worse than \Mid for every policy. A stricter test, that the rewound run then matches \Pre, passes for three policies, and an exploratory check attributes most of the remaining gap to the steps lost in rewinding. Finally, the cost outlasts the affected instruction. The change stays in place, and among cases in which a policy completes that instruction with and without the change, it is \AftLoMag{} to \AftHiMag{} points less likely to complete the next one. An exploratory comparison with \Pre suggests that this cost is about the same whether the change arrived mid-motion or was there from the start.

\noindent In summary, our contributions are:
\begin{enumerate}[itemsep=0.3ex, parsep=0pt, topsep=1pt, leftmargin=1.4em, label=\textbf{(\arabic*)}]
\item \textbf{A three-arm paired protocol on chains of instructions.} \maxbench adds a third arm, \Pre, to the paired protocol of \lmax. \Pre shares with \Mid the history before the affected instruction and the change itself, so the loss on that instruction is the sum of a cost of the changed world and an effect of its timing (\cref{eq:decomp}). The 7,828 cases also vary the moment of a change and include changes to the robot and the instruction; we evaluate a held-out third of the core and embodiment cases on \NPolWord policies and treat the other tracks as exploratory (\cref{sec:benchmark,sec:protocol}).
\item \textbf{A timing effect that depends on the kind of change.} On held-out cases, a perceptual change costs \PoolPercMag{} points less when it arrives mid-motion and a physical change shows no detectable effect; the difference, fixed in advance as our primary test, is \PoolInter{} points \PoolInterCI{} and is significant for four policies individually. Rewinding the robot removes the benefit for every policy, and the stricter check that the rewound run matches \Pre holds for three (\cref{sec:timing,sec:mechanism}).
\item \textbf{A cost that outlasts the affected instruction.} Among cases in which the affected instruction is completed with and without the change, every policy is \AftLoMag{} to \AftHiMag{} points less likely to complete the next one. An exploratory comparison with \Pre finds about the same cost when the change was there from the start (\cref{sec:chain}).
\end{enumerate}

\section{Related Work}
\label{sec:related}

\paragraph{Manipulation policies and action chunking.}
Vision-language-action models map camera images and a language instruction directly to robot actions~\cite{Brohan2023RT2VM,kim2024openvla,black2025pi0,pi05,zheng2026xvla,univla,reuss2025flower}. Other designs generate actions by diffusion~\cite{reuss2024mdt,reuss2025mode}, reason over a 3D representation of the scene~\cite{ke20243dda}, or draw on a history of frames, through a recurrent head~\cite{li2024roboflamingo} or by predicting the frames that come next~\cite{wu2024gr1,tian2025seer}. Most of these policies predict a short sequence of actions, called a chunk, and execute several of them before looking at the scene again~\cite{zhao2023aloha,chi2023diffusionpolicy}, which delays their reaction to anything that happens in between. Later methods therefore adapt the chunk length during execution~\cite{liang2026adaptive} or replan more often while keeping consecutive chunks consistent~\cite{black2025rtc,liu2025bid}. Queued actions are only one of the ways in which a policy is tied to its past, however, because the arm is already in motion and the policy may remember earlier frames. Our \NPolWord policies span chunks of 1 to 30 actions and include models with such a memory, which lets us ask, for each policy, where the benefit is held.

\paragraph{Benchmarks and robustness.}
Manipulation policies are mostly compared on simulation benchmarks, which have grown from collections of tabletop skills to language-conditioned tasks in household scenes~\cite{yu2020meta,james2020rlbench,mu2021maniskill,gu2023maniskill2,liu2023libero,nasiriany2024robocasa}. CALVIN~\cite{mees2022calvin} stands out among them because it asks for five instructions in a row, so that each one starts from wherever the previous one left the robot and the scene. A second line of work asks whether performance survives conditions unseen in training. LIBERO-Plus, LIBERO-PRO and LIBERO-X perturb the camera, the lighting, the layout of the objects and the wording of the instruction~\cite{fei2025liberoplus,zhou2025liberopro,wang2026liberox}, and other stress tests vary the scene, the distractors and the robot~\cite{pumacay2024colosseum,wang2025vlatest,zhang2024vlabench,li24simpler,rasouli2026distracted,zhang2026wam}. Corruption benchmarks for image classifiers further showed the value of grading each perturbation by severity~\cite{hendrycks2019benchmarking}, a practice we adopt when we vary the strength of a change. In all of these benchmarks the perturbation is fixed before the episode begins, so a policy is never asked to revise a motion it has already started.

\paragraph{Changes during execution.}
Fewer works change the world while the robot is acting. DynamicVLA and DOMINO~\cite{xie2026dynamicvla,fang2026dynamicmanipulation} study objects that are themselves in motion and reward fast, reactive control, so that motion is the task itself and not an interruption of it, and SwitchVLA and ReSteer~\cite{li2025switchvla,chen2026resteer} switch the instruction during a task. \lmax~\cite{zhang2026liberomax}, the closest work to ours, instead interrupts an ordinary task with one abrupt change, and introduced the paired comparison with an undisturbed run that we build on. It found that policies of very different design are vulnerable to the same changes, and that observing the scene more often does not remove the loss. Its paired comparison has only the undisturbed run as a reference, however, so the cost of acting in a changed world and the cost of meeting the change mid-motion remain mixed, its one test of timing covers a single change and three policies, and its single-instruction tasks leave open what a change costs later. \maxbench addresses these points with a third matched run and chains of instructions (\cref{tab:compare}).

\section{The \maxbench Benchmark}
\label{sec:benchmark}

CALVIN~\cite{mees2022calvin} evaluates a policy on \emph{chains} of five language instructions carried out in sequence in one scene. Each instruction, which CALVIN calls a subtask, has a limit of 360 steps. The policy's internal state is reset when an instruction begins, and the chain ends at the first instruction that fails. \maxbench introduces one \emph{change} into one instruction of a chain, the \emph{affected instruction}, and measures its effect by comparing runs that are identical up to that point.

\subsection{Three-arm paired evaluation}
\label{sec:arms}

A \emph{case} specifies a chain, the position $k\in\{1,\dots,5\}$ of the affected instruction, a change with fixed parameters, and an \emph{onset} rule that selects the step $t_\text{on}$ of instruction $k$ at which the change occurs. For a given policy and seed, every case is evaluated in three runs, or \emph{arms} (\cref{fig:teaser}). \Base runs the chain with no change, and one \Base rollout per chain serves all cases on that chain. \Mid replays the actions of \Base from the initial state up to $t_\text{on}$, applies the change, and returns control to the policy with its internal state intact, including any queued actions and remembered observations, as in the Dynamic run of \lmax~\cite{zhang2026liberomax}. \Pre replays \Base only to the start of instruction $k$, where the policy is reset as usual, and applies there exactly the change that \Mid receives, for example the same final pose of a relocated block.

The three arms therefore share the instructions, the initial state, the policy seed and every action through instruction $k{-}1$, and \Pre and \Mid also receive the same change. Unlike a change applied at the initial reset, as in the camera control of \lmax, \Pre leaves the trajectory before the affected instruction untouched, so these two arms differ only in whether the change was present when the policy began instruction $k$ or arrived after it had started to act. Let $Y^B_i$, $Y^P_i$ and $Y^M_i\in\{0,1\}$ denote success on the affected instruction of case $i$ in each arm, and let a bar denote the mean over a set of cases. The total effect of a change then splits exactly as
\begin{equation}
\underbrace{\overline{Y^M-Y^B}}_{\dtot} \;=\; \underbrace{\overline{Y^P-Y^B}}_{\dconf} \;+\; \underbrace{\overline{Y^M-Y^P}}_{\dtim}.
\label{eq:decomp}
\end{equation}
The \emph{configuration effect} $\dconf$ is the cost of the changed world when it is in place from the start of the instruction, and the \emph{timing effect} $\dtim$ is the additional cost, or benefit, of meeting it mid-motion. A negative $\dtim$ means that commitment to the old world hurts, a positive one that earlier progress helps. The split is exact on any common set of cases, but $\dtim$ is a net contrast: it includes both the progress made before the change and the shorter time that remains after it. We report all three in percentage points (pp). \Cref{sec:mechanism} divides $\dtim$ with two diagnostic arms.

\subsection{Changes and onsets}
\label{sec:events}

\paragraph{Changes.} The eight \emph{core} changes keep the names and purpose of those in \lmax (\cref{fig:teaser}, left), and we group them by the \emph{channel} through which they affect the policy. Five are \emph{perceptual} and change what the policy sees but not the geometry of the task: a camera shift, sensor noise, an illumination switch, a visual-theme switch and a burst of distractors. Three are \emph{physical} and change where the robot must go: relocation of the target or of the receptacle, and insertion of an obstacle. Each core change has two fixed variants that differ in strength or appearance, such as a relocation by 6 or by 12\,cm. \maxbench adds \emph{embodiment} changes, which act on the robot through a push on its arm or a grasp loss, in which a held block slips from the gripper, and an \emph{intent} change, which revises the instruction so that success is judged on a new goal. Two controls complete the set. A \emph{null control} replays the run to the onset and changes nothing, which measures the noise floor of each policy. A \emph{transient} relocation moves a block and returns it a few steps later, whereas the core changes persist to the end of the chain and the embodiment changes act once. \Cref{tab:events} lists all parameters and \cref{sec:supp_qual} shows examples.

\paragraph{Onsets.} An onset is computed from the \Base run using the simulator's privileged state, which the policy never observes, and never from an outcome, so the moment of a change is not chosen by whether the policy succeeds. Our default is the \emph{pre-grasp} onset of \lmax, the first step at which the gripper comes within a radius $r$ of the object that the instruction refers to. We calibrate $r=6$\,cm on development chains, which places the median onset near the middle of the instruction, and insert obstacles earlier, during the approach, so that they lie on the robot's path. To vary this moment, we add onsets at the \emph{start} of the instruction and at its \emph{approach}, \emph{contact}, \emph{transport} and \emph{place} phases (\cref{tab:supp_onsets}).

\subsection{Benchmark construction}
\label{sec:tracks}

\paragraph{Tracks.} We build on CALVIN's 1,000 evaluation chains, in the standard setting where policies are trained on scenes A, B and C and tested on the unseen scene D (ABC$\rightarrow$D). All cases are fixed before any policy is run and chosen without reference to policy outcomes, and a disjoint set of 200 chains is used only for calibration. The full benchmark, \emph{Max}, contains 7,828 cases on 999 chains in five \emph{tracks} (\cref{fig:lineage,tab:supp_alloc}). The \emph{core} track crosses the eight core changes and their two variants with the five positions on 60 chains each, which gives 4,680 cases because receptacle relocation has none at the first position, where nothing is held yet. The \emph{phase} track places four changes at up to six onsets (2,100 cases), the \emph{embodiment} track holds the arm push and grasp loss (450), the \emph{intent} track the instruction revisions (298) and the \emph{null} track the null controls (300). \Cref{tab:compare} compares this scope with related benchmarks.

\begin{table}[t]
\caption{Scope of \maxbench and related benchmarks. A check marks a property that a benchmark's cases cover, a circle one covered only in part, and a cross one that is not covered. \emph{Chain} means tasks of several instructions, \emph{Online} a change during execution, \emph{Prefix} identical actions up to the change, \emph{Pre/Mid} the same change before and during the motion, \emph{Onset} a varied moment of change, \emph{Emb./int.} changes to the robot and the instruction, \emph{Behav.} measures of how the robot responds, and \emph{Units} the scored variants, episodes or cases. This paper evaluates held-out core and embodiment cases of \maxbench on all ten policies (\cref{sec:protocol}).}
\label{tab:compare}
\centering
\fontsize{6.5}{7.8}\selectfont
\setlength{\tabcolsep}{1.4pt}
\renewcommand{\arraystretch}{1.12}
\begin{tabularx}{\linewidth}{X *{7}{c} r}
\toprule
\tableheader Benchmark & \tableheader Chain & \tableheader Online & \tableheader Prefix & \tableheader Pre/Mid & \tableheader Onset & \tableheader Emb./int. & \tableheader Behav. & \tableheader Units \\
\midrule
CALVIN~\cite{mees2022calvin} & \cmark & \xmark & \xmark & \xmark & \xmark & \xmark & \xmark & 1,000 \\
LIBERO-Plus~\cite{fei2025liberoplus} & \xmark & \xmark & \xmark & \xmark & \xmark & \xmark & \xmark & 10,030 \\
LIBERO-PRO~\cite{zhou2025liberopro} & \xmark & \xmark & \xmark & \xmark & \xmark & \xmark & \xmark & 10,000 \\
\lmax~\cite{zhang2026liberomax} & \xmark & \cmark & \cmark & \pmark & \xmark & \xmark & \pmark & 8,000 \\
\rowcolor{takeaway!7}
\textbf{\maxbench} & \cmark & \cmark & \cmark & \cmark & \cmark & \cmark & \cmark & 7,828 \\
\bottomrule
\end{tabularx}
\vspace{-3mm}
\end{table}

\paragraph{Per-policy realization.} The scene at instruction $k$ depends on what the policy did in the earlier instructions, so the exact result of a change, such as the pose of a relocated block, cannot be fixed in advance. A case therefore fixes only what is independent of the policy: the chain, the position, the change and its variant parameters, the onset rule and a random seed. The concrete change is computed from each policy's \Base run at the onset and applied in \Pre and \Mid. Cases are specified identically across policies, and \Pre and \Mid of a policy receive the same concrete change.

\paragraph{Validity.} Two properties of CALVIN require care. First, its simulator holds state outside the observations, and restoring a saved state drifts by up to 1\,cm. Every arm therefore replays its actions from the initial reset in a fresh simulator process, and each policy must pass a null-control check before evaluation: of 293 null runs, 283 reproduce \Base bit for bit and 290 keep the outcome of every instruction (\cref{sec:supp_seeds}). Second, CALVIN judges an instruction against the scene at its start, so moving an object can by itself complete or break a task. For changes that move an object or the robot, we update the moved object's pose in this reference scene and require that the verdict stay unchanged while the robot holds still for 30 steps. Each case is also replayed to its onset to confirm that the change can be applied and leaves the task feasible. In a human audit of 400 realized cases of one policy, an author who saw the change but not the outcome judged none infeasible and two unclear (\cref{sec:supp_humanaudit}). A case that fails these checks is \emph{unrealizable}, and a case whose \Base run never meets the onset, for example because the chain has already failed, is \emph{unreached}. Neither is discarded: as in \lmax, such a case keeps its \Base outcome in \Pre and \Mid and counts in every success rate. We call all other cases \emph{realized}. Runs that fail for technical reasons are rerun, never dropped (\cref{sec:supp_validity,sec:supp_audit}).

\paragraph{Lite and Max-A.} For rapid iteration, \emph{Lite} is a fixed subset of 672 Max cases, six for every combination of change, variant, and position or onset, drawn from the case descriptions alone. An exploratory set repeats selected Lite cases at five strengths. Because Lite shaped our hypotheses, we fixed them and the analysis code before running \emph{Max-A}, the first third of the cases, in chain order, for every change, variant and position of the core and embodiment tracks. Without its Lite cases, Max-A has 1,526 cases per policy on 385 chains. It is our confirmatory set, and each test is also reported on the union of Lite and Max-A (\cref{sec:supp_lite,sec:supp_prereg}).

\section{Results}
\label{sec:results}

\subsection{Experimental setup}
\label{sec:protocol}

\paragraph{Policies.} We evaluate ten released checkpoints trained for CALVIN ABC$\rightarrow$D. Five are VLAs (FLOWER~\cite{reuss2025flower}, X-VLA~\cite{zheng2026xvla}, $\pi_{0.5}$~\cite{pi05}, UniVLA~\cite{univla} and OpenVLA-OFT~\cite{kim2025oft}), two are diffusion policies (MoDE~\cite{reuss2025mode} and 3D Diffuser Actor~\cite{ke20243dda}, the only policy with depth input), two are predictive models that plan again at every step from ten past frames (GR-1~\cite{wu2024gr1} and Seer-Large~\cite{tian2025seer}), and RoboFlamingo~\cite{li2024roboflamingo} is a vision-language model with a recurrent memory. We run each policy with its reference inference code and compare its no-change score with the published one (\cref{sec:supp_policies}). Every policy uses CALVIN's limit of 360 steps per instruction, which lowers the scores of three policies published with longer limits, 3D Diffuser Actor most of all. \lmax also evaluated X-VLA and $\pi_{0.5}$.

\paragraph{Protocol.} We formed six hypotheses on Lite (H1 to H6, \cref{tab:prereg}) and fixed them, together with the analysis code, before running Max-A (\cref{sec:supp_prereg}). All headline numbers come from the 1,526 held-out \Track cases per policy, of which 1,377 are core cases, and the diagnostic arms of \cref{sec:mechanism} run on its 887 perceptual core cases. We report success on the affected instruction over all assigned cases, together with the differences of \cref{eq:decomp}. The change is realized in \maxa{45} to \maxa{81\%} of a policy's core cases and the others keep their \Base outcome, so these differences average over cases in which nothing changed. Comparisons of the timing effect between channels use only realized cases, which are the same for \Pre and \Mid, and count the two embodiment changes as physical, as the frozen analysis code does (\cref{sec:supp_prereg}). Unless noted, intervals are 95\% bootstrap intervals that resample whole chains, because the cases on a chain share one \Base rollout, and estimates pooled over policies weight each policy equally. Tests on single policies are Holm-corrected for the ten comparisons, and we claim no effect that lies within the noise a policy shows under the null control, in which \Mid reproduces \Base on all 30 Lite cases for every policy (\cref{sec:supp_metrics,sec:supp_seeds}). The analyses of strength, onset phase and instruction revision lie outside Max-A and are exploratory.

\begin{table*}[t]
\caption{Results for each policy on the held-out cases. $H/Q$ gives the actions a policy predicts and executes per query, and Chain the instructions it completes on average with no change. Success is how often the affected instruction is completed in each run, over all 1,377 assigned core cases, and the paired differences split the total loss into the cost of the changed world and the effect of timing. The last three columns give the timing effect by channel on realized cases, where physical also includes the two embodiment changes, and the difference between channels, with 95\% intervals; the difference is significant after correction for X-VLA, UniVLA, 3D Diffuser Actor and Seer-Large. Each policy runs with its released settings, so rows compare released policies, not designs.}
\label{tab:main}
\centering
\scriptsize
\setlength{\tabcolsep}{3.1pt}
\renewcommand{\arraystretch}{1.12}
\begin{tabularx}{\linewidth}{@{}ll c c ccc c>{\hspace{2.8pt}}c<{\hspace{2.8pt}}c ZZZ@{}}
\toprule
\multirow[c]{2}{*}[-0.5ex]{\tableheader Family} & \multirow[c]{2}{*}[-0.5ex]{\tableheader Policy} & \multirow[c]{2}{*}[-0.5ex]{\tableheader $H/Q$} & \multirow[c]{2}{*}[-0.5ex]{\tableheader Chain\rlap{$\mkern1mu\uparrow$}} &
\multicolumn{3}{c}{\tableheader Success (\%)\,$\uparrow$} & \multicolumn{3}{c}{\tableheader Paired difference (pp)} & \multicolumn{3}{c}{\tableheader $\dtim$ by channel (pp)} \\
\cmidrule(lr){5-7}\cmidrule(lr){8-10}\cmidrule(l){11-13}
& & & & \tableheader \Base & \tableheader \Pre & \tableheader \Mid & \tableheader $\dtot$\rlap{$\mkern1mu\uparrow$} & \tableheader $\dconf$\rlap{$\mkern1mu\uparrow$} & \tableheader $\dtim$ & \tableheader \textcolor{perc}{Perceptual} & \tableheader \textcolor{phys}{Physical} & \tableheader Difference \\
\midrule
\rowgray VLA & FLOWER & 10/10 & 4.53 & 92.7 & 84.6 & 88.2 & $-$4.6 & $-$8.1 & +3.6 & \tci{+5.1}{+3.3}{+6.9} & \tci{+2.3}{$-$0.5}{+5.1} & \tci{+2.7}{$-$0.4}{+6.1} \\
 & X-VLA & 30/20 & 4.10 & 81.8 & 69.6 & 73.2 & $-$8.6 & $-$12.2 & +3.6 & \tci{+6.8}{+4.4}{+9.3} & \tci{+1.0}{$-$2.7}{+4.7} & \tci{+5.8}{+1.3}{+10.3} \\
\rowgray  & $\pi_{0.5}$ & 5/5 & 3.73 & 75.0 & 69.0 & 70.0 & $-$5.0 & $-$6.0 & +1.0 & \tci{+1.6}{$-$0.5}{+3.9} & \tci{+0.3}{$-$3.1}{+3.8} & \tci{+1.3}{$-$2.8}{+5.4} \\
 & UniVLA & 12/12 & 3.73 & 71.5 & 59.8 & 61.7 & $-$9.8 & $-$11.7 & +1.9 & \tci{+5.0}{+1.7}{+8.3} & \tci{$-$2.9}{$-$7.5}{+1.6} & \tci{+7.9}{+2.5}{+13.4} \\
\rowgray  & OpenVLA-OFT & 1/1 & 2.85 & 57.2 & 45.2 & 47.5 & $-$9.7 & $-$12.0 & +2.3 & \tci{+5.6}{+1.7}{+9.4} & \tci{+1.0}{$-$4.0}{+6.1} & \tci{+4.6}{$-$1.9}{+11.0} \\
\midrule
Diffusion & MoDE & 10/10 & 4.05 & 79.3 & 71.2 & 73.6 & $-$5.7 & $-$8.1 & +2.5 & \tci{+5.1}{+2.8}{+7.6} & \tci{$-$1.0}{$-$5.7}{+3.5} & \tci{+6.1}{+1.1}{+11.2} \\
\rowgray  & 3D Diffuser Actor & 19/19 & 1.93 & 38.2 & 30.6 & 32.1 & $-$6.1 & $-$7.6 & +1.5 & \tci{+7.8}{+2.9}{+12.8} & \tci{$-$5.0}{$-$12.2}{+2.1} & \tci{+12.8}{+5.0}{+20.6} \\
\midrule
Predictive & GR-1 & 1/1 & 3.20 & 66.5 & 53.4 & 56.6 & $-$9.9 & $-$13.1 & +3.2 & \tci{+6.6}{+3.8}{+9.4} & \tci{+1.0}{$-$3.3}{+5.4} & \tci{+5.6}{+0.4}{+10.8} \\
\rowgray  & Seer-Large & 1/1 & 4.27 & 85.1 & 77.3 & 79.6 & $-$5.5 & $-$7.8 & +2.3 & \tci{+4.4}{+2.6}{+6.3} & \tci{$-$1.9}{$-$4.9}{+1.1} & \tci{+6.4}{+2.8}{+10.0} \\
\midrule
VLM+LSTM & RoboFlamingo & 1/1 & 2.48 & 51.4 & 44.7 & 47.0 & $-$4.4 & $-$6.7 & +2.3 & \tci{+5.5}{+1.9}{+9.1} & \tci{$-$1.0}{$-$5.7}{+3.7} & \tci{+6.6}{+0.7}{+12.4} \\
\midrule
\multicolumn{3}{@{}l}{\textbf{All ten policies}, equal weight} & 3.49 & 69.9 & 60.5 & 63.0 & $-$6.9 & $-$9.3 & +2.4 & \tci{\textbf{+5.4}}{+4.3}{+6.4} & \tci{$-$0.6}{$-$2.2}{+0.9} & \tci{\textbf{+6.0}}{+4.1}{+7.9} \\
 
\bottomrule
\end{tabularx}
\vspace{-2mm}
\end{table*}

\subsection{Every policy loses success after a change}
\label{sec:loss}

Every policy completes the affected instruction less often once the world changes (\cref{fig:overview}a, \cref{tab:main}). The total effect $\dtot$ ranges from \DTotMax{} to \DTotMin{} points, and between \maxa{5.4} and \maxa{25.7\%} of the affected instructions that a policy solves in \Base fail in \Mid (\cref{tab:supp_transitions}). These losses are smaller than in \lmax, but the two are not directly comparable, since CALVIN scores each short instruction separately and unrealized cases keep their \Base outcome.

For every policy the configuration effect $\dconf$ (\DConfMax{} to \DConfMin{} points) is larger in magnitude than the total effect, so the changed world costs more when it is in place from the start of the instruction than when it arrives mid-motion. \lmax observed this under a camera shift with a control applied at reset; for X-VLA it recurs here with a control that shares the history of the run (\Base \maxa{82.8}, \Pre \maxa{49.4}, \Mid \maxa{63.8\%}). Which change costs most differs between policies (\cref{fig:events}, \cref{sec:supp_events}).

\subsection{The timing effect depends on the channel}
\label{sec:timing}

\begin{figure}[t]
\centering
\includegraphics[width=\linewidth]{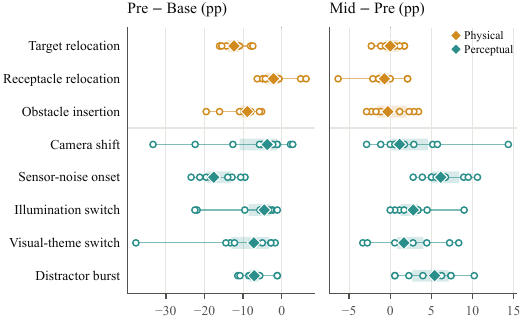}
\vspace{-5mm}
\caption{The cost of the changed world (\Pre$-$\Base, left) and the effect of timing (\Mid$-$\Pre, right) for each core change on the held-out cases. Each open circle is one policy, and the diamond, shading and thin line give the median, middle half and full range. For physical changes, at the top, the median timing effect is near or below zero, with policies on both sides, while perceptual changes mostly gain mid-motion.}
\label{fig:events}
\vspace{-3mm}
\end{figure}

Whether timing matters depends on the channel of the change (\cref{fig:overview}b). When a perceptual change arrives mid-motion, policies do better than when the same change is present from the start: $\dtim$ is positive for every policy, from \PercMin{} to \PercMax{} points, and \PoolPerc{} \PoolPercCI{} when pooled. For physical changes, which here include the two changes to the robot, we detect no such benefit. Their pooled $\dtim$ is \PoolPhys{} \PoolPhysCI{}: the estimate lies inside the $\pm$1-point band that we fixed in advance and its interval includes zero, so we report no detectable timing effect for physical changes in either direction (H2 in \cref{tab:prereg}), although the interval does not rule out a cost of up to 2.2 points.

Our primary test (H1) is the difference between the two channels. Pooled over the policies it is \PoolInter{} \PoolInterCI{} points ($p<0.001$; \cref{fig:overview}c), and it remains when 3D Diffuser Actor is left out and when the Lite cases are added (\cref{tab:prereg}). The difference is positive for every policy, from \InterMin{} to \InterMax{} points, but after correction it is significant for only four of them (X-VLA, Seer-Large, UniVLA and 3D Diffuser Actor). The design anticipated this, with a power of about 0.25 to detect a 5-point effect in one policy, and three training seeds of MoDE differ on Lite by about 4 points (\cref{sec:supp_seeds}). We therefore read the result as a property of these policies as a group and not as a ranking among them.

\begin{table}[t]
\caption{The hypotheses we fixed before running the held-out cases and how each test came out. Estimates are in pp with 95\% intervals, or ranges over the ten policies for per-policy tests, and significance is Holm-corrected over the policies (\cref{tab:supp_prereg_res}). H4 and H5 need sweeps that the held-out set does not contain; they were set aside as exploratory before it was run (\cref{sec:dose}).}
\label{tab:prereg}
\centering
\scriptsize
\setlength{\tabcolsep}{3pt}
\renewcommand{\arraystretch}{1.1}
\begin{tabularx}{\linewidth}{@{}Xll@{}}
\toprule
\tableheader Test & \tableheader Estimate (pp) & \tableheader Result \\
\midrule
\multicolumn{3}{@{}l}{\textbf{H1}\enspace timing helps perceptual more than physical changes} \\
\quad all ten policies & +6.0 [+4.1, +7.9] & holds \\
\quad without 3D Diffuser Actor & +5.2 [+3.4, +7.0] & holds \\
\quad with the Lite cases added & +6.8 [+5.2, +8.4] & holds \\
\quad each policy & +1.3 to +12.8 & 4/10 significant \\
\addlinespace
\multicolumn{3}{@{}l}{\textbf{H2}\enspace physical changes show no timing effect} \\
\quad all ten policies & $-$0.6 [$-$2.2, +0.9] & no effect seen \\
\addlinespace
\multicolumn{3}{@{}l}{\textbf{H3}\enspace rewinding removes the perceptual benefit} \\
\quad \Mid minus rewound run & +3.0 to +20.4 & 10/10 significant \\
\quad rewound run minus \Pre & $-$12.4 to +0.7 & 3/10 within $\pm$3 \\
\addlinespace
\textbf{H4}\enspace benefit grows with the damage & not run on held-out cases & exploratory \\
\textbf{H5}\enspace loss comes before contact & not run on held-out cases & exploratory \\
\addlinespace
\multicolumn{3}{@{}l}{\textbf{H6}\enspace a change still hurts the next instruction} \\
\quad \Mid minus \Base & $-$16.8 to $-$5.0 & 10/10 significant \\
 
\bottomrule
\end{tabularx}
\vspace{-3mm}
\end{table}

\Cref{fig:events} shows which changes are responsible. Sensor noise and distractors carry most of the perceptual benefit, with a positive timing effect for all ten policies, and the illumination and visual-theme switches add a smaller one. Target relocation, receptacle relocation and obstacle insertion have timing effects near or below zero for most policies.

\subsection{Rewinding the robot removes the benefit}
\label{sec:mechanism}

Why should a policy cope better with a corrupted view that arrives mid-motion? At the moment of the change, \Mid differs from \Pre in two ways: the policy may hold a queue of actions planned from the clean view, and its arm and memory are already part of the way through the motion. Two diagnostic arms remove these one at a time. \textbf{\Mid-flush} applies the change as \Mid does but discards the queued actions, so that the policy must plan again from the changed view. \textbf{\Mid-rewind} also returns the arm and the policy's memory to their state at the start of the instruction, while the world stays changed. If nothing else separates \Mid from \Pre, a rewound run should then succeed as often as \Pre. The rewind is not applied when the robot touches an object at the onset, so cases with a held block drop out (\cref{sec:supp_arms}). We run the rewind on every policy and the flush on the six that execute more than one action per query (H3).

\begin{figure}[t]
\centering
\includegraphics[width=\linewidth]{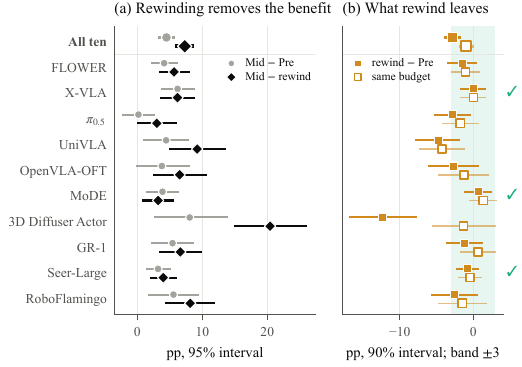}
\vspace{-5mm}
\caption{Panel (a) compares \Mid with \Pre and with a rewound run, whose arm and memory return to the start of the instruction while the world stays changed, on the realized perceptual cases in which the robot can be rewound, with 95\% intervals. \Mid beats the rewound run for every policy, so rewinding removes the benefit. Panel (b) shows what remains, the rewound run minus \Pre, as run and with \Pre given the same remaining steps, with 90\% intervals; a check marks each policy inside the shaded $\pm$3 band. The top row pools the ten policies (descriptive).}
\label{fig:h3}
\vspace{-3mm}
\end{figure}

Rewinding removes the benefit for every policy. \Mid succeeds more often than the rewound run for every policy, by \GainMin{} to \GainMax{} points, and each difference is significant after correction (\cref{fig:h3}a). This is what we would expect if the benefit lay in where the robot is when the change arrives: a policy that is part of the way through the motion has less left to do from the corrupted view. The stricter question, which decides H3, is whether the rewound run then matches \Pre. It does, to within the $\pm$3 points that we fixed in advance, for three policies (MoDE, X-VLA and Seer-Large), the only ones for which our rule confirms that the benefit comes from motion already under way (\cref{sec:supp_prereg}). FLOWER and GR-1 miss this margin by less than a point, and for the other five the rewound run falls below \Pre (\cref{fig:h3}b). Most of that shortfall arises because a rewound run restarts the instruction with only the steps that \Mid had left, a median of \maxa{28} to \maxa{66} fewer than the 360 that \Pre receives. When \Pre is held to the same number of steps (an exploratory check), the gap for 3D Diffuser Actor, which the 360-step limit constrains most, shrinks from \maxa{$-$12.4} to \maxa{$-$1.3} points, and every gap except that of UniVLA (\maxa{$-$4.3} \maxa{[$-$7.3, $-$1.3]}) is within \maxa{1.8} points of zero.

The flush arm indicates where a policy holds the benefit, and this differs between the released checkpoints. X-VLA, which executes 20 of its 30 predicted actions before it looks again, holds most of it in its queue (\maxa{+4.7} of the \maxa{+6.2} points by which \Mid beats its rewound run). FLOWER, MoDE and UniVLA hold it in the state of the robot and the policy's memory (\maxa{+4.1}, \maxa{+3.0} and \maxa{+7.1} points). These splits are a few points wide, about the spread between training seeds (\cref{sec:supp_seeds}), so we read them as descriptive. The four policies that plan again at every step have no queue and still beat the rewound run by \maxa{+4.0} to \maxa{+8.1} points, so a queue is not required.

The same progress does not help when an object moves (\Track, exploratory; \cref{tab:supp_behaviour}). After a target relocation, the tool centre turns toward the block's new pose within a few steps in the runs that fail as in those that succeed, and only \maxa{27} of the \maxa{150} failures across the ten policies pass through the block's old pose, so most failures do not reach for the old pose.

\subsection{The cost extends to the next instruction}
\label{sec:chain}

\begin{figure}[t]
\centering
\includegraphics[width=\linewidth]{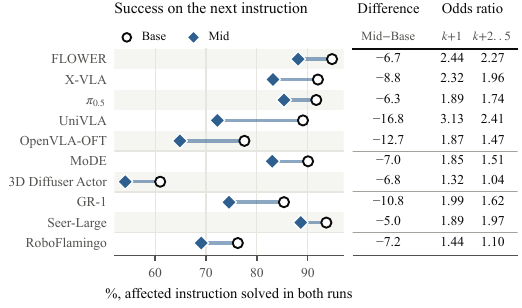}
\vspace{-5mm}
\caption{Success on the next instruction with no change (\Base) and after a change (\Mid), among held-out cases in which both runs completed the affected instruction. The comparison is conditional on that success and is not a causal estimate over all cases. The columns give the difference in points and how much higher the odds of failing are after the change, on the next instruction ($k{+}1$) and on each later one.}
\label{fig:aftershock}
\vspace{-3mm}
\end{figure}

Because CALVIN's tasks are chains, we can ask what a change costs after the policy has dealt with it. Where the affected instruction succeeded in both \Base and \Mid, we compare success on the next instruction, with the change still in place. It is lower after the change for all \NPolMaxWord policies, by \AftLoMag{} to \AftHiMag{} points, and significantly so for each of them in a one-sided test after correction (H6; \cref{fig:aftershock}, \cref{tab:supp_aftershock}). This comparison is conditional on success in both arms.

Every core change stays in place to the end of the chain, so this cost could come from the changed world or from having met the change mid-motion. The \Pre arm helps to tell them apart (exploratory; \cref{sec:supp_chain}). Where all three arms complete the affected instruction, success on the next one falls from \maxa{85.5\%} in \Base to \maxa{78.0\%} in \Pre and \maxa{77.5\%} in \Mid, and \Mid differs from \Pre by only \maxa{$-$0.5} \maxa{[$-$1.6, +0.5]} points (\maxa{$-$2.2} \maxa{[$-$4.0, $-$0.3]} for physical changes alone). The cost is thus about the same whenever the change arrived, which is consistent with a cost of the changed world that a policy pays again on the next instruction. The three arms also end the affected instruction in different states, which this comparison does not separate. In an exploratory hazard model the odds of failing stay \maxa{1.67} \maxa{[1.52, 1.85]} times higher on each later instruction.

\subsection{Changes to the robot, and exploratory analyses}
\label{sec:newch}\label{sec:dose}

\Track also covers the two embodiment changes for all \NPolMaxWord policies (\cref{tab:supp_embodiment}). A push on the arm changes success on the affected instruction by \PushTot{} \PushTotCI{} points on average over all assigned cases (\maxa{$-$5.6} \maxa{[$-$8.8, $-$2.7]} where it is realized), and it shows no timing effect (\PushTim{} \PushTimCI{}). A block slipping from the gripper applies only when a block is held at the onset, in \maxa{11} to \maxa{26} of the \maxa{49} cases per policy; where it occurs it costs \maxa{4.9} points on average \maxa{[$-$8.5, $-$1.7]} (\GraspTot{} \GraspTotCI{} over all assigned cases).

Three further analyses lie outside \Track, use small samples and are exploratory. Under sensor noise at four times the default strength (60 Lite cases per policy), \Mid recovers \lite{53} to \lite{67\%} of the loss in \Pre for five policies and \lite{8} to \lite{13\%} for four (H4; \cref{sec:supp_ladders}). When the onset of sensor noise is moved through the instruction for three policies, the loss falls mostly before contact (H5; \cref{sec:supp_phase}). A revised instruction is the only change whose pooled timing effect is negative, \lite{$-$9.4} \lite{[$-$16.4, $-$3.6]} points over 230 Lite cases, with no single policy significant (\cref{sec:supp_intent}).

\paragraph{Limitations.} All results come from one simulator and from released checkpoints with one training seed each on the held-out cases, run under a common limit of 360 steps, so we report effects for the ten policies as a group and do not rank them. No policy is run on all 7,828 cases: the confirmatory results use a held-out third of the core and embodiment tracks, whose cases are disjoint from Lite's but share chains with them, and the analyses of strength, onset phase and revised instructions use small sets outside it. Comparisons between channels use realized cases, so they are conditional on each policy's own \Base trajectory. The rewound run has fewer steps than \Pre and leaves out cases with a held block. Our hypotheses were fixed in a dated internal record before the held-out runs, not in a public registry.

\section{Conclusion}
\label{sec:conclusion}

We introduced \maxbench, a benchmark that measures not only whether a change during execution hurts a robot policy, but also when it hurts and what it costs afterwards. Its three arms share every action before the affected instruction, so each loss divides into the cost of acting in a changed world and the effect of meeting the change mid-motion. On held-out cases, ten released policies as a group lose less to a perceptual change when it arrives mid-motion, a benefit that rewinding the robot removes, and show no detectable effect of timing for a physical change. A policy that copes with a change still fails the next instruction more often while the change stays in place. The benchmark makes it possible to test whether a response to change can be selective, keeping progress under a corrupted view and dropping it once an object has moved. We will release the cases, the runner, the per-case results and the analysis code.

\section*{LLM Usage Statement}
The authors conceived the study, designed the benchmark and the evaluation protocol, decided what to run, how to analyse it and what to conclude, and carried out the human feasibility audit. An AI coding agent (Claude, Anthropic) worked under their supervision on four parts of the paper. It wrote and tested the code to the authors' specifications, including the scripts that build the benchmark cases. It launched and monitored the experiments and the analysis scripts that the authors designed. It produced the figures from the results. It drafted and edited the text to the authors' outline. Three AI tools (Claude Fable 5.2, Anthropic, and GPT 6 Astra, OpenAI, both at high reasoning effort; Deep Review, Google) also reviewed a draft, and the authors decided which of their suggestions to adopt. The authors reviewed and approved the code and the text and are responsible for all content.

{
    \small
    \bibliographystyle{ieeenat_fullname}
    \bibliography{main}
}

\clearpage
\setcounter{figure}{0}\setcounter{table}{0}
\renewcommand{\thefigure}{S\arabic{figure}}\renewcommand{\theHfigure}{S\arabic{figure}}
\renewcommand{\thetable}{S\arabic{table}}\renewcommand{\theHtable}{S\arabic{table}}
\clearpage
\maketitlesupplementary
\appendix
\renewcommand{\thesection}{\Alph{section}}
\renewcommand{\topfraction}{0.95}\renewcommand{\bottomfraction}{0.7}\renewcommand{\textfraction}{0.05}
\renewcommand{\floatpagefraction}{0.85}\renewcommand{\dbltopfraction}{0.95}\renewcommand{\dblfloatpagefraction}{0.85}
\setcounter{topnumber}{4}\setcounter{bottomnumber}{3}\setcounter{totalnumber}{6}\setcounter{dbltopnumber}{4}

\newcommand{\tocrow}[4]{\noindent\hbox to \linewidth{\strut\hspace*{#1}\makebox[2.5em][l]{#2}#3\hfil #4}\par\vskip 0pt plus 1fil\relax}
\newcommand{\tocsec}[1]{\tocrow{0pt}{\textbf{\hyperref[#1]{\ref*{#1}}}}{\textbf{\hyperref[#1]{\nameref*{#1}}}}{\hyperref[#1]{\pageref*{#1}}}}
\newcommand{\tocsub}[1]{\tocrow{0.8em}{\hyperref[#1]{\ref*{#1}}}{\hspace*{0.8em}\hyperref[#1]{\nameref*{#1}}}{\hyperref[#1]{\pageref*{#1}}}}
\newcommand{\tocgap}{\vskip 3pt plus 1fil\relax}
\begin{center}
{\large\bfseries\color{takeaway} Contents}
\end{center}
\vspace{-1mm}
{\small
\makeatletter\let\texorpdfstring\@secondoftwo\makeatother %
\setlength{\parskip}{0pt}\setlength{\lineskip}{0pt}\setlength{\baselineskip}{1.08\baselineskip}
\ifdefined\nolinenumbers\nolinenumbers\fi %
\tocsec{sec:supp_related} \tocsub{sec:supp_rel_bench} \tocsub{sec:supp_rel_calvin} \tocsub{sec:supp_rel_commit} \tocgap
\tocsec{sec:supp_construction} \tocsub{sec:supp_design} \tocsub{sec:supp_alloc} \tocsub{sec:supp_eventdefs} \tocsub{sec:supp_validity} \tocsub{sec:supp_humanaudit} \tocsub{sec:supp_qual} \tocgap
\tocsec{sec:supp_lite} \tocsub{sec:supp_lite_sel} \tocsub{sec:supp_lite_agree} \tocgap
\tocsec{sec:supp_protocol} \tocsub{sec:supp_metrics} \tocsub{sec:supp_policies} \tocsub{sec:supp_seeds} \tocgap
\tocsec{sec:supp_breakdowns} \tocsub{sec:supp_outcomes} \tocsub{sec:supp_events} \tocsub{sec:supp_position} \tocgap
\tocsec{sec:supp_decomp} \tocsub{sec:supp_arms} \tocsub{sec:supp_strongnoise} \tocsub{sec:supp_transient} \tocgap
\tocsec{sec:supp_dose} \tocsub{sec:supp_ladders} \tocsub{sec:supp_phase} \tocgap
\tocsec{sec:supp_chain} \tocgap
\tocsec{sec:supp_behaviour} \tocsub{sec:supp_behav_metrics} \tocsub{sec:supp_gate} \tocgap
\tocsec{sec:supp_newch} \tocsub{sec:supp_embodiment} \tocsub{sec:supp_intent} \tocgap
\tocsec{sec:supp_auditprereg} \tocsub{sec:supp_audit} \tocsub{sec:supp_prereg} \tocgap
\tocsec{sec:supp_discussion} \tocsub{sec:supp_interp} \tocsub{sec:supp_limits} \tocsub{sec:supp_future} \tocsub{sec:supp_compute}
\vskip 0pt plus -1fil\relax %
\vskip 0pt plus -1fil\relax
\newpage}
\ifdefined\linenumbers\linenumbers\fi

\section{Additional Related Work}
\label{sec:supp_related}

\subsection{Benchmarks and mid-execution changes}
\label{sec:supp_rel_bench}
Beyond the benchmarks in \cref{sec:related}, household, bimanual and safety-oriented suites broaden scenes and failure types~\cite{nasiriany2024robocasa,cui2026liberosafety}, an analysis of long-horizon household tasks argues that success rates alone can hide unsafe behaviour and overstate performance~\cite{rasouli2026howvlas}, and causal and intervention-based environments vary physical factors in a controlled way~\cite{ahmed2020causalworld}. Most measure success under conditions fixed at the start. \lmax~\cite{zhang2026liberomax} is the first we know of to pair rollouts around a single change applied after an identical executed prefix; \maxbench adopts that unit and adds the arms, tracks and measurements described in \cref{sec:benchmark}.

\subsection{Language-conditioned policies on CALVIN}
\label{sec:supp_rel_calvin}
CALVIN was introduced with play-data baselines~\cite{mees2022calvin} and hierarchical latent-plan policies~\cite{mees2022hulc}. Diffusion transformers conditioned on multimodal goals~\cite{reuss2024mdt}, mixture-of-experts denoisers~\cite{reuss2025mode} and flow-matching VLAs~\cite{reuss2025flower} now approach saturation of the ABC$\rightarrow$D average chain length. Vision-language backbones with recurrent heads~\cite{li2024roboflamingo,li2024robovlms}, video-pretrained GPT policies~\cite{wu2024gr1}, predictive inverse-dynamics models~\cite{tian2025seer}, latent-action VLAs~\cite{univla} and 3D policies~\cite{ke20243dda} represent other design choices. High \Base success makes the paired loss on the affected instruction interpretable: a policy that already fails cannot lose.

\subsection{Action chunking and replanning}
\label{sec:supp_rel_commit}
Action chunking~\cite{zhao2023aloha} and diffusion policies~\cite{chi2023diffusionpolicy} execute several actions open loop. Bidirectional decoding~\cite{liu2025bid}, real-time chunking~\cite{black2025rtc} and adaptive chunk lengths~\cite{liang2026adaptive} trade off consistency and reactivity. Our flush, rewind and gate arms (\cref{sec:supp_decomp,sec:supp_gate}) measure how the timing effect changes when the queued actions are discarded or the robot and policy are returned to the start of the instruction.

\section{Benchmark Construction and Examples}
\label{sec:supp_construction}

\subsection{Design requirements}
\label{sec:supp_design}
Five requirements define the comparison. \emph{Temporal placement}: the change occurs after the policy has begun the affected instruction. \emph{Shared history}: all arms share the instructions, initial state, seed and executed prefix through instruction $k{-}1$. \emph{Same change}: \Pre and \Mid receive the same concrete change. \emph{Every case counts}: unreached and unrealizable cases stay in every success rate. \emph{Blind to outcomes}: the cases, the onsets and the concrete changes are all chosen without reference to policy outcomes.

\subsection{Chain selection and allocation}
\label{sec:supp_alloc}
\begin{figure*}[!tp]
\centering
\resizebox{\linewidth}{!}{%
\begin{tikzpicture}[
  font=\sffamily\footnotesize,
  panel/.style={draw=black!18, fill=black!2.5, rounded corners=3pt, inner sep=0pt, minimum height=46mm, anchor=north west},
  head/.style={font=\sffamily\small\bfseries, anchor=north},
  sub/.style={font=\sffamily\scriptsize\itshape, text=black!60, anchor=north},
  flow/.style={-{Stealth[length=2.2mm]}, line width=1.1pt, black!55},
  trk/.style={font=\sffamily\scriptsize\bfseries, anchor=west, inner sep=0pt},
  cnt/.style={font=\sffamily\scriptsize\bfseries, anchor=west, inner sep=0pt},
  dsc/.style={font=\sffamily\scriptsize, text=black!65, anchor=west, inner sep=0pt},
  chip/.style={fill=#1!12, draw=#1!60, text=#1!70!black, rounded corners=1.5pt, font=\sffamily\scriptsize,
               inner sep=1.4pt, minimum width=21.5mm, minimum height=4.4mm},
  >=Stealth]

\node[panel, minimum width=44mm] (p1) at (0,0) {};
\node[head] at ($(p1.north)+(0,-2mm)$) {CALVIN};
\node[anchor=north, font=\sffamily\footnotesize] at ($(p1.north)+(0,-7.0mm)$) {\textbf{1,000} chains of \textbf{5} instructions};
\node[anchor=north, inner sep=0pt] (img) at ($(p1.north)+(0,-13.6mm)$)
  {\includegraphics[width=30mm]{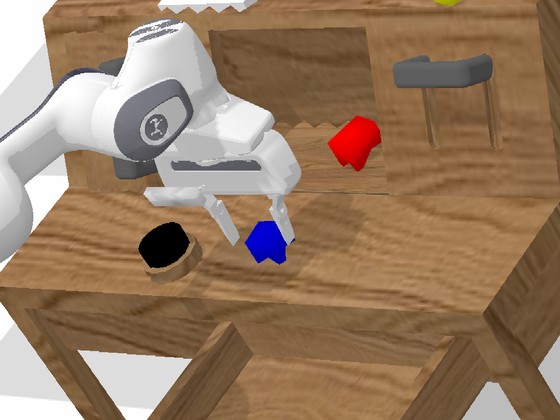}};
\node[anchor=south, font=\sffamily\scriptsize, text=black!65, align=center, inner sep=0pt] at ($(p1.south)+(0,1.6mm)$)
  {trained on scenes A, B, C\\[0.5pt]
   tested on the unseen scene D};

\node[panel, minimum width=100mm] (p2) at ($(p1.north east)+(6mm,0)$) {};
\node[head] at ($(p2.north)+(0,-2mm)$) {Five tracks of changes};
\node[sub] at ($(p2.north)+(0,-6.2mm)$) {each case adds one change to one chain, chosen before any policy is run};
\def\ucase{0.0041mm}  %
\def\dy{5.0mm}
\coordinate (r0) at ($(p2.north west)+(4mm,-14.4mm)$);   %
\coordinate (b0) at ($(p2.north west)+(72mm,-14.4mm)$);  %
\draw[black!15] ($(r0)+(0,3.1mm)$) -- ($(r0)+(92mm,3.1mm)$);
\newcommand{\trkbar}[4]{\fill[#4] ($(b0)+(#2*\ucase,-#1*\dy-1.4mm)$) rectangle ++(#3*\ucase,2.8mm);}
\foreach \r/\name/\d in {0/Core/{8 changes $\times$ 2 variants $\times$ 5 positions},
                         1/Phase/{4 changes at up to 6 onsets},
                         2/Embodiment/{arm pushed or object dropped},
                         3/Intent/{instruction revised during the motion},
                         4/Null/{replayed, nothing changes}} {
  \node[trk] at ($(r0)+(0,-\r*\dy)$) {\name};
  \node[dsc] at ($(r0)+(21mm,-\r*\dy)$) {\d};
}
\trkbar{0}{0}{3000}{perc}\trkbar{0}{3000}{1680}{phys}
\trkbar{1}{0}{600}{perc}\trkbar{1}{600}{900}{phys}\trkbar{1}{1500}{600}{emb}
\trkbar{2}{0}{450}{emb}
\trkbar{3}{0}{298}{intent}
\trkbar{4}{0}{300}{intent}
\foreach \r/\n/\x in {0/{4,680}/4680, 1/{2,100}/2100, 2/450/450, 3/298/298, 4/300/300}
  \node[cnt] at ($(b0)+(\x*\ucase+1.2mm,-\r*\dy)$) {\n};
\draw[black!15] ($(r0)+(0,-4*\dy-3.1mm)$) -- ($(r0)+(92mm,-4*\dy-3.1mm)$);
\node[anchor=north, inner sep=0pt, font=\sffamily\scriptsize] at ($(p2.north)+(0,-39.6mm)$) {%
  \textcolor{perc}{\rule{2.4mm}{2.4mm}}\,perceptual\hspace{3.2mm}%
  \textcolor{phys}{\rule{2.4mm}{2.4mm}}\,physical\hspace{3.2mm}%
  \textcolor{emb}{\rule{2.4mm}{2.4mm}}\,embodiment\hspace{3.2mm}%
  \textcolor{intent}{\rule{2.4mm}{2.4mm}}\,intent \& control};

\node[panel, minimum width=58mm, draw=black!30] (p3) at ($(p2.north east)+(6mm,0)$) {};
\node[anchor=north, font=\sffamily\bfseries\fontsize{15}{16}\selectfont] at ($(p3.north)+(0,-2.4mm)$) {\maxbench};
\node[anchor=north, font=\sffamily\small] at ($(p3.north)+(0,-9.8mm)$) {\textbf{7,828} cases on \textbf{999} chains};
\def\cw{14mm}   %
\coordinate (s0) at ($(p3.north west)+(5.0mm,-38.6mm)$);  %
\foreach \i/\f/\c in {3/original/intent, 2/emb_arm_push/emb, 1/target_relocation/phys, 0/distractor_burst/perc} {
  \begin{scope}[shift={($(s0)+(\i*3.0mm,\i*2.9mm)$)}]
    \fill[black, opacity=0.18] (0.5mm,-0.5mm) rectangle ++(\cw,\cw);
    \node[anchor=south west, inner sep=0pt] at (0,0) {\includegraphics[width=\cw]{figures/gallery/\f.jpg}};
    \draw[\c, line width=1.2pt] (0,0) rectangle (\cw,\cw);
  \end{scope}
}
\foreach \j/\c/\lab/\n in {0/intent/{Intent \& control}/598, 1/emb/Embodiment/{1,050},
                           2/phys/Physical/{2,580}, 3/perc/Perceptual/{3,600}}
  \node[anchor=west, inner sep=0pt, font=\sffamily\scriptsize] at ($(p3.north west)+(32.6mm,-19.6mm-\j*4.6mm)$)
    {\textcolor{\c}{\rule[-0.2mm]{2.4mm}{2.4mm}}\,\textcolor{\c!70!black}{\lab}~\textbf{\n}};
\node[anchor=south, font=\sffamily\scriptsize, text=black!65] at ($(p3.south)+(0,1.6mm)$)
  {each case is run as \Base, \Pre and \Mid (phase: \Base and \Mid)};

\draw[flow] (p1.east) -- (p2.west);
\draw[flow] (p2.east) -- (p3.west);
\end{tikzpicture}}
\vspace{-5mm}
\caption{From CALVIN's chains to the five tracks of \maxbench. CALVIN's 1,000 test chains each run five instructions in a scene the policies never saw during training. Every case adds one change to one chain at one moment, and the cases fall into five tracks, drawn as bars sized by their number of cases and coloured by family of change. The right panel counts cases by family across all tracks, so its embodiment count includes the 600 arm pushes of the phase track. Each case is run three ways as in \cref{fig:teaser}, except that phase cases run only \Base and \Mid, and the cases are specified identically for every policy.}
\label{fig:lineage}
\vspace{-3mm}
\end{figure*}

\Cref{fig:lineage} shows how the benchmark is built from CALVIN's evaluation chains, and \cref{tab:supp_alloc} gives the Max allocation. In the core track, every combination of a change, a variant and a position has 60 chains, except receptacle relocation at the first position, which has none because nothing is held at the start of a chain (4,680 cases rather than 4,800). Chains are drawn independently for each combination with seed \texttt{20261028} from the 1,000 frozen evaluation chains, whose list is fixed once because CALVIN's sampler depends on process scheduling; one chain is drawn by no combination, so Max covers 999. The phase track places target relocation at four onsets (not transport or place, when the target is already held), receptacle relocation at five (not transport), and camera shift and arm push at all six, with 100 cases per onset, and runs only \Mid. The embodiment track has 150 cases for each arm-push variant at the pre-grasp onset and 150 grasp losses at transport. The intent track has 298 cases because two of its 300 drawn chains admit no feasible revision.

\begin{table}[!tp]
\caption{Number of cases in each track of Max and of Lite. Every case shares its chain's \Base rollout and receives a \Pre and a \Mid rollout, except phase cases, which receive only \Mid. Lite takes six cases per combination, drawn with seed \texttt{20261029}.}
\label{tab:supp_alloc}
\centering
\footnotesize
\setlength{\tabcolsep}{2.6pt}
\begin{tabularx}{\linewidth}{@{}l >{\raggedright\arraybackslash}X >{\raggedright\arraybackslash}X r r@{}}
\toprule
\tableheader Track & \tableheader Changes & \tableheader Onsets & \tableheader Max & \tableheader Lite \\
\midrule
Core & 8 core changes, 2 variants, $k{=}1..5$ & pre-grasp (obstacle: approach) & 4,680 & 468 \\
Phase & target, receptacle, camera, arm push & up to six onsets & 2,100 & 126 \\
Embodiment & arm push, grasp loss & pre-grasp, transport & 450 & 18 \\
Intent & instruction revision & pre-grasp & 298 & 30 \\
Null & null control & pre-grasp & 300 & 30 \\
\midrule
Total & & & 7,828 & 672 \\
\bottomrule
\end{tabularx}
\end{table}

\subsection{Changes, onsets and parameters}
\label{sec:supp_eventdefs}
\Cref{tab:events} lists every change, its channel and its two fixed variants; \cref{tab:supp_onsets} gives the onset rules. All onsets are computed from the \Base run using the simulator's privileged state. The pre-grasp radius ($r{=}6$\,cm, measured from the tool centre point to the axis-aligned bounding box of the object) was calibrated on 944 development instructions so that the median onset falls in the middle of the instruction (46\%). An onset whose condition first holds only at the final action of an instruction would land at the start of the next instruction, so such a case counts as unreached (\cref{sec:supp_audit}).

\paragraph{Magnitude of each change.} Magnitudes are set by their effect, not raw units. The static camera is about 4\,m from the table with a 10$^\circ$ field of view, so camera shifts are orbits about the look-at point calibrated by image displacement (10--20 pixels for variant A). Block relocations are sampled on the block's own support surface, at least 6\,cm from other blocks, and must remain visible.

\paragraph{The same change in \Pre.} The concrete change is computed once, from the \Base state at the onset, and \Pre applies that same change at the start of the instruction. A relocated block or joint is set to the same final pose. A grasp loss places the block on the surface at the point below where it was held at the onset, since nothing is held yet when the instruction begins. An arm push displaces the arm by the same vector from its pose at the start. A case is therefore realized, or not, for both arms together. For the two changes to the robot, \Pre thus meets a related but not identical change: a block already lying at the drop point, a relocation rather than a lost grasp, and the same displacement applied to a different arm pose. The split $\dtot=\dconf+\dtim$ holds on any common set of cases regardless, but for these changes $\dtim$ also contains this difference. In both arms the order is the same: \Base's actions are replayed from the initial reset in a fresh process up to the branch point; the policy is reset (\Pre) or its state at the onset restored (\Mid); the change is applied, followed by a collision pass; the reference scene is recorded (\cref{sec:supp_validity}); and the run continues from a new observation, \Mid first executing any actions left in its queue.

\begin{table*}[!tp]
\caption{Every change and how it is realized in CALVIN. The core changes share names and purpose with \lmax, and each has two fixed variants, A and B. The channel is how we expect timing to matter. The track column says where a change is reported: the core changes enter the headline numbers, and the embodiment and intent changes and the null control are reported separately, with the embodiment changes also counted as physical in the comparison between channels. Each transient relocation is paired with a Lite relocation case and lies outside the five tracks.}
\label{tab:events}
\centering
\footnotesize
\setlength{\tabcolsep}{3.2pt}
\renewcommand{\arraystretch}{1.08}
\begin{tabularx}{\linewidth}{@{}l l Y l l@{}}
\toprule
\tableheader Change & \tableheader Channel & \tableheader CALVIN realization (variant A / B) & \tableheader Onset & \tableheader Track \\
\midrule
Target relocation & \textcolor{phys}{Physical} & Block moved 6 / 12\,cm on the table (B also rotated 45$^\circ$); drawer or slider jumped to 30 / 60\% of its travel & pre-grasp & core \\
Receptacle relocation & \textcolor{phys}{Physical} & Stack-base block moved 6 / 12\,cm; drawer or slider shifted 20 / 40\% while still receiving the block & pre-grasp & core \\
Obstacle insertion & \textcolor{phys}{Physical} & 4\,cm cube / 3$\times$3$\times$10\,cm post placed in the approach corridor & approach & core \\
Camera shift & \textcolor{perc}{Perceptual} & Static camera rotated by 0.7$^\circ$ yaw and 0.35$^\circ$ pitch / by 1.4$^\circ$ and 0.7$^\circ$ with a 1.4\,cm lift and a 1$^\circ$ wider field of view & pre-grasp & core \\
Sensor-noise onset & \textcolor{perc}{Perceptual} & Gaussian noise ($\sigma{=}0.05$) and 10\% occlusion on the static camera / on both cameras & pre-grasp & core \\
Illumination switch & \textcolor{perc}{Perceptual} & Light direction, colour, ambient and diffuse terms changed (neutral / warm) & pre-grasp & core \\
Visual-theme switch & \textcolor{perc}{Perceptual} & Table and furniture recoloured (beige / slate) & pre-grasp & core \\
Distractor burst & \textcolor{perc}{Perceptual} & Five neutral-coloured objects (3 / 5\,cm) dropped on free table space & pre-grasp & core \\
\midrule
Transient relocation & \textcolor{phys}{Transient} & Target or receptacle moved as above, returned after 2--20 steps & pre-grasp & exploratory \\
Arm push & \textcolor{emb}{Embodiment} & End effector displaced 5 / 10\,cm in a seeded horizontal direction, held block carried along & pre-grasp, phases & embodiment \\
Grasp loss & \textcolor{emb}{Embodiment} & Held block released onto the table below the gripper & transport & embodiment \\
Instruction revision & \textcolor{intent}{Intent} & Goal of the instruction swapped to a different feasible task in the same scene; success judged on the new goal & pre-grasp & intent \\
Null control & \textcolor{intent}{Control} & Replay to the onset, change nothing & pre-grasp & null \\
\bottomrule
\end{tabularx}
\end{table*}

\begin{table}[!htbp]
\caption{The six onset rules, each computed from the \Base run. The approach and pre-grasp rules fire no earlier than 10 steps into the instruction, and a pre-grasp onset whose tool centre starts inside $r$ fires after exactly 10 steps.}
\label{tab:supp_onsets}
\centering
\footnotesize
\setlength{\tabcolsep}{3pt}
\begin{tabularx}{\linewidth}{@{}lY@{}}
\toprule
\tableheader Onset & \tableheader Definition \\
\midrule
Instruction start & First step of instruction $k$, the time point of \Pre; the policy is reset there as in \Pre, so this \Mid starts from the same robot and policy state and step budget \\
Approach & Distance from the tool centre to the object's bounding box down by 25\% from its value at the start \\
Pre-grasp & Tool centre first within $r{=}6$\,cm of the object's bounding box (of the receptacle for place and stack tasks; the rule of \lmax) \\
Contact & First contact between the robot and the object \\
Transport & Held block more than 3\,cm above its height at the start \\
Place & After transport, held block within 6\,cm of its receptacle \\
\bottomrule
\end{tabularx}
\end{table}

\subsection{Validity and treatment of failures}
\label{sec:supp_validity}

\paragraph{Exact replay.} Hidden simulator state (joint and object velocities, button logic, the controller's target pose) is not in CALVIN's observations, and restoring a saved state within a process drifts by up to 1\,cm. Every arm replays actions from the initial reset in a fresh process, rendered or not. Each policy seeds its sampling noise per query from the case, so a replayed run draws the same actions. The policy's internal state (queued actions, histories, buffers of robot state) is saved at the onset. Each policy runs a null-control check before evaluation: 283 of 293 null runs reproduce \Base bit for bit and 290 keep the outcome of every instruction (\cref{sec:supp_seeds}). MoDE's 12 null runs were spread over three copies of its policy server and are all bit-identical, so copies of a server keep the pairing exact.

\paragraph{Updating the reference scene.} CALVIN's task checker reports every task that is satisfied between a reference state, recorded at the start of the instruction, and the current state. Moving an object can therefore satisfy a push task by itself (a 6\,cm move can register as a push), or break a lift. In \Mid the reference stays the state at the start of the instruction except for what the change moved, which takes its state just after the change and a collision pass: the pose and contacts of a relocated or dropped block, or of a block riding in a moved drawer or slider, and the joint state of that drawer or slider. A block carried by the gripper during an arm push keeps its reference contacts instead, and its reference position is shifted by the arm's achieved displacement, so that lift progress made before the push still counts (first row of \cref{tab:supp_audit}). Every other object is still judged against the start of the instruction. In \Pre the reference is the whole scene just after the change. A transient relocation restores the block's reference when the block returns, and a revised instruction swaps the task that is checked at the onset. We verified the update once, on a 12\,cm relocation, where the unchanged reference reports a false push and the updated one reports nothing while the robot holds still for 30 steps; this hold is not repeated for every case.

\paragraph{Checks before evaluation.} Cases of the changes that place or move an object (target and receptacle relocation, obstacle insertion, distractor burst and the transient relocations) are replayed to their onset in a fresh process and the change applied, followed by a collision pass. The change must alter at least one camera image, a target block must stay visible (at least 6 pixels in the static camera), the change must create no new contact with the robot, and every object not held must move less than 5\,mm while the robot holds still for 20 steps. The other changes are not replayed before evaluation. Feasibility is not tested in simulation: it rests on how the changes are sampled (\cref{sec:supp_eventdefs}) and is audited by hand on a sample of accepted cases (\cref{sec:supp_humanaudit}). A case that fails a check, or for which no valid change can be found, is \emph{unrealizable}; a case whose chain fails before instruction $k$, or whose onset rule never fires within it, is \emph{unreached}. Both keep their \Base outcome and stay in every success rate, and coverage reports how often (\cref{sec:supp_outcomes}). Technical failures are rerun, never dropped. The distractors are light boxes (50\,g) with collision shapes: the checks exclude contact with the robot at the onset, but a later contact with the robot or another object is neither prevented nor analysed, so the distractor burst is perceptual by design rather than isolated from the physics.

\subsection{Human feasibility audit}
\label{sec:supp_humanaudit}
We drew 400 realized held-out FLOWER cases from Max-A with a fixed seed: 80 each for target relocation, receptacle relocation and obstacle insertion, 54 arm pushes, all 26 realized grasp losses, and 16 for each perceptual change. For each case we re-rendered the static and gripper cameras just before and just after the change from the stored \Base actions, with no policy in the loop. One author judged every case on a labeling page that shows the instruction and the change but no outcome, answering whether a competent robot could still complete the instruction from the state after the change (feasible, infeasible or unclear). Of the 400 cases, 398 were judged feasible, none infeasible and two unclear (\cref{tab:supp_humanaudit}); counting the unclear ones as infeasible gives a feasible share of 99.5\% [98.2, 99.9]. In the two unclear cases, a block was moved deep onto the slider inside the cabinet, and the drawer was partly closed while a block was carried to it; FLOWER completed both in \Pre and \Mid. One case, an illumination switch, was judged without its instruction text because of a page error, and its answer was kept. The audit has one rater and one policy, covers only cases that the automated checks accepted, and judges physical feasibility from the state just after the change in both cameras, not what a policy that sees one camera can observe. The case selection, the rendered frames, the answers and the labeling page will be released with the benchmark for reproducibility.

\begin{table}[!htbp]
\caption{Human feasibility audit of 400 realized held-out FLOWER cases. The feasible share is taken over the cases judged feasible or infeasible, so the two unclear cases are left out of it, with a 95\% Wilson interval. No case was judged infeasible.}
\label{tab:supp_humanaudit}
\centering
\footnotesize
\setlength{\tabcolsep}{3pt}
\begin{adjustbox}{max width=\columnwidth}
\begin{tabular}{@{}l r r r r c@{}}
\toprule
\tableheader Change & $n$ & \tableheader Feasible & \tableheader Infeasible & \tableheader Unclear & \tableheader Feasible share (\%) \\
\midrule
Target relocation & 80 & 79 & 0 & 1 & 100 [95.4, 100] \\
Receptacle relocation & 80 & 79 & 0 & 1 & 100 [95.4, 100] \\
Obstacle insertion & 80 & 80 & 0 & 0 & 100 [95.4, 100] \\
Arm push & 54 & 54 & 0 & 0 & 100 [93.4, 100] \\
Grasp loss & 26 & 26 & 0 & 0 & 100 [87.1, 100] \\
Perceptual (five changes) & 80 & 80 & 0 & 0 & 100 [95.4, 100] \\
\midrule
All & 400 & 398 & 0 & 2 & 100 [99.0, 100] \\
\bottomrule
\end{tabular}\end{adjustbox}
\end{table}

\subsection{Qualitative intervention examples}
\label{sec:supp_qual}
\Cref{fig:gallery_perc,fig:gallery_phys} show both variants of every core change and the two changes to the robot in the static and gripper cameras, just before and just after the change. \Cref{fig:qual_perc,fig:qual_phys} show the three arms of one case per change from CALVIN's static and gripper cameras, and \cref{fig:qual_seq,fig:qual_seq_phys} follow two of them after the onset. \Base and \Mid are shown at the onset, immediately before and after the change; \Pre is shown at the start of instruction $k$, where the same changed world is present from the first observation.

\newlength{\qualw}\setlength{\qualw}{0.168\textwidth}

\newcommand{\qualcell}[1]{\IfFileExists{figures/qual/#1.jpg}{\includegraphics[width=\linewidth]{figures/qual/#1.jpg}}{\fcolorbox{black!25}{black!4}{\parbox[c][0.82\linewidth][c]{0.9\linewidth}{\centering\footnotesize\color{black!45} render pending\\\texttt{\detokenize{#1}}}}}}

\newcommand{\qualmark}[2]{\qualcell{#1}\llap{\raisebox{0.5mm}{\setlength{\fboxsep}{0.5pt}\colorbox{white}{\footnotesize #2}}\hspace{0.5mm}}\qualgrip{#1}}
\newcommand{\qualgrip}[1]{\IfFileExists{figures/qual/#1-g.jpg}{\par\vspace{1pt}\includegraphics[width=\linewidth]{figures/qual/#1-g.jpg}}{}}
\newcommand{\armsep}{\noalign{\vskip 4pt}\multicolumn{6}{@{}r@{}}{\makebox[\dimexpr 5\qualw+8\tabcolsep\relax]{\color{black}\leaders\hbox to 4pt{\hss\rule[0.6ex]{2.2pt}{0.5pt}\hss}\hfill}}\\[3pt]}
\newcommand{\armlabel}[1]{\IfFileExists{figures/qual/camera_shift-base-g.jpg}{\raisebox{-\qualw}{\rotatebox{90}{\makebox[2\qualw][c]{\footnotesize #1}}}\hspace{1pt}\raisebox{-\qualw}{\rotatebox{90}{\makebox[\qualw][c]{\scriptsize\color{black!60}gripper}\makebox[\qualw][c]{\scriptsize\color{black!60}static}}}}{\rotatebox{90}{\makebox[\qualw][c]{\footnotesize #1}}}}

\newlength{\galw}\setlength{\galw}{0.19\textwidth}
\newcommand{\galtitle}[2]{\centering\footnotesize #1\\[-1pt]{\scriptsize\emph{#2}}}
\newcommand{\galimg}[1]{\includegraphics[width=\linewidth]{figures/gallery/#1.jpg}}
\newcommand{\galcams}{\rotatebox{90}{\makebox[0.5\galw][c]{\scriptsize\color{black!60}gripper}\makebox[0.5\galw][c]{\scriptsize\color{black!60}static}}}

\begin{figure*}[p]
\centering
\setlength{\tabcolsep}{1.5pt}
\begin{tabular}{@{}c*{5}{p{\galw}}@{}}
 & \galtitle{Camera shift A}{rotate blue block left} & \galtitle{Sensor noise A}{unstack block} & \galtitle{Illumination A}{push pink block left} & \galtitle{Visual theme A}{lift red block (table)} & \galtitle{Distractors A}{move slider left} \tabularnewline
\galcams & \galimg{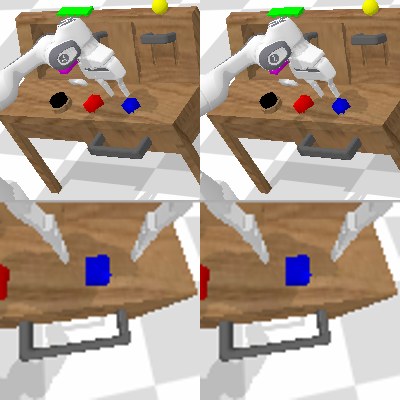} & \galimg{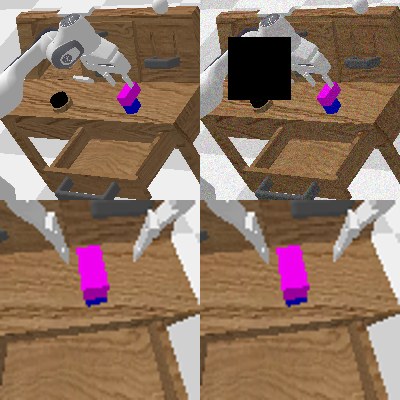} & \galimg{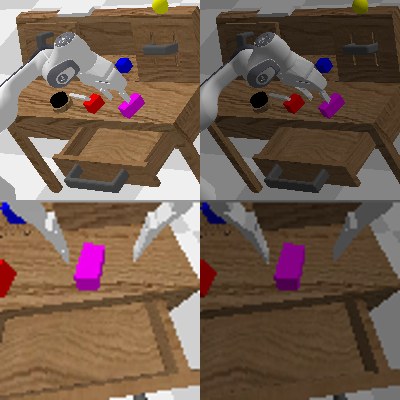} & \galimg{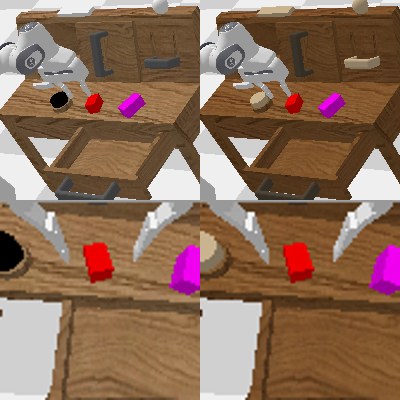} & \galimg{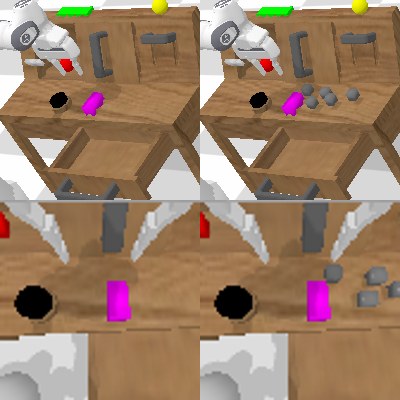} \tabularnewline[3pt]
 & \galtitle{Camera shift B}{place in drawer} & \galtitle{Sensor noise B}{turn off LED} & \galtitle{Illumination B}{lift pink block (drawer)} & \galtitle{Visual theme B}{turn on lightbulb} & \galtitle{Distractors B}{push red block left} \tabularnewline
\galcams & \galimg{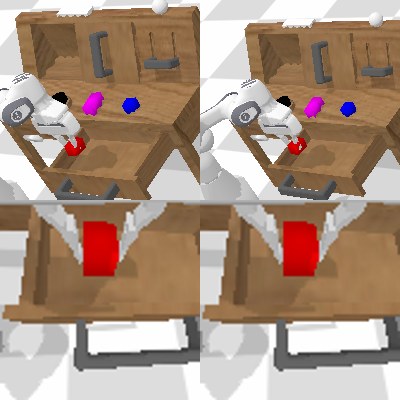} & \galimg{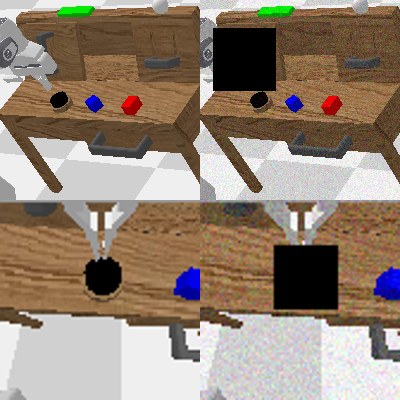} & \galimg{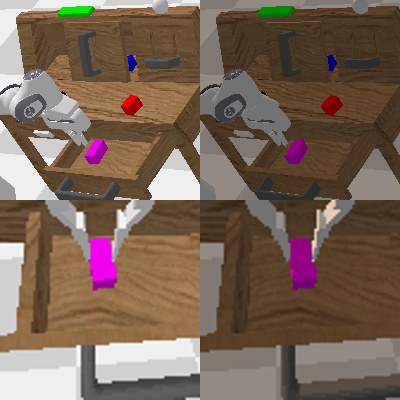} & \galimg{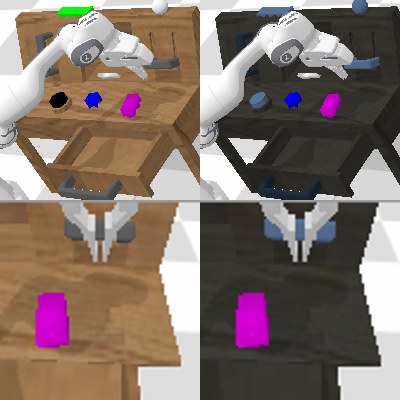} & \galimg{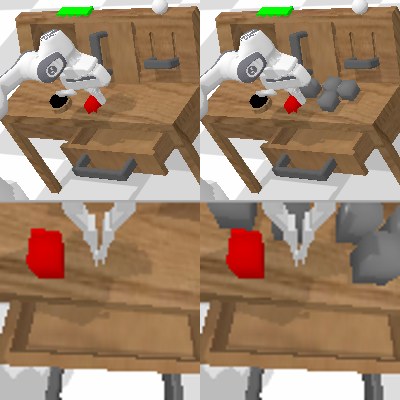} \tabularnewline
\end{tabular}
\caption{Every perceptual change, in both variants. Each panel shows the static camera (top) and the gripper camera (bottom) just before (left) and just after (right) the change, with the instruction under its title. Frames are re-rendered from the stored \Base actions of held-out FLOWER cases from the human audit (\cref{sec:supp_humanaudit}), with no policy in the loop, and were chosen to show each change clearly, without reference to any outcome. The camera shift moves only the static camera, and sensor-noise variant A corrupts only the static camera, B both. The gripper camera has 84$\times$84 pixels and is shown enlarged.}
\label{fig:gallery_perc}
\end{figure*}

\begin{figure*}[p]
\centering
\setlength{\tabcolsep}{1.5pt}
\begin{tabular}{@{}c*{5}{p{\galw}}@{}}
 & \galtitle{Target, block A}{push pink block right} & \galtitle{Target, block B}{rotate pink block right} & \galtitle{Target, drawer A}{open drawer} & \galtitle{Receptacle, stack base B}{stack block} & \galtitle{Receptacle, drawer A}{place in drawer} \tabularnewline
\galcams & \galimg{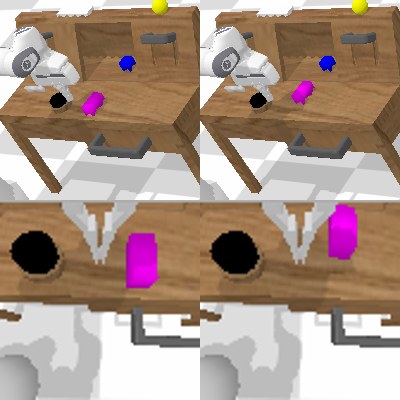} & \galimg{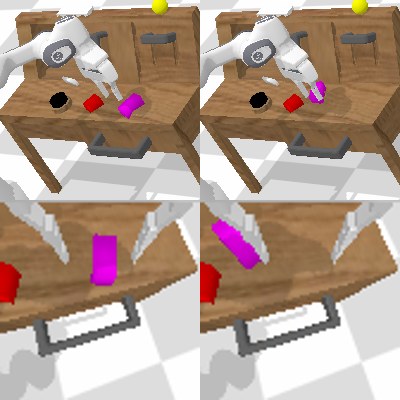} & \galimg{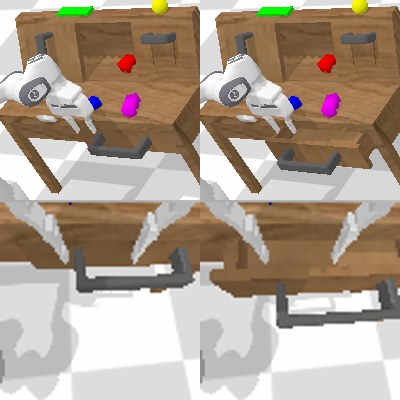} & \galimg{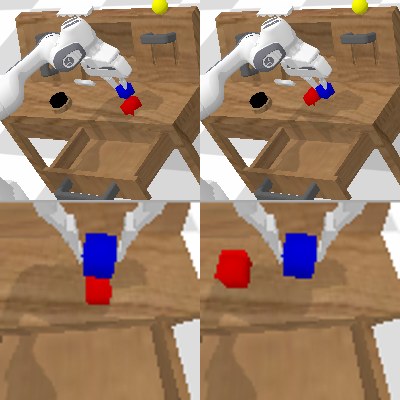} & \galimg{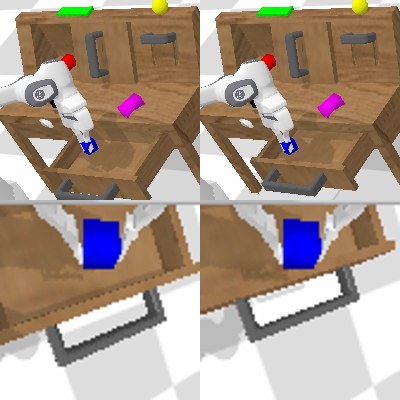} \tabularnewline[3pt]
 & \galtitle{Obstacle A (cube)}{lift blue block (table)} & \galtitle{Obstacle B (post)}{lift red block (table)} & \galtitle{Arm push A}{turn on LED} & \galtitle{Arm push B}{lift pink block (slider)} & \galtitle{Grasp loss}{lift pink block (table)} \tabularnewline
\galcams & \galimg{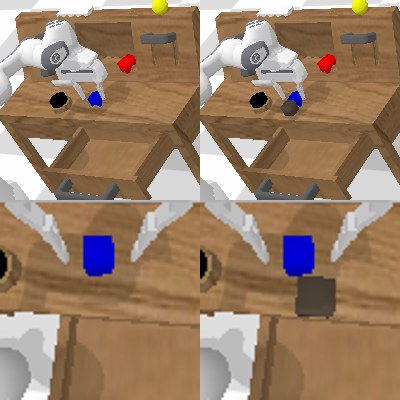} & \galimg{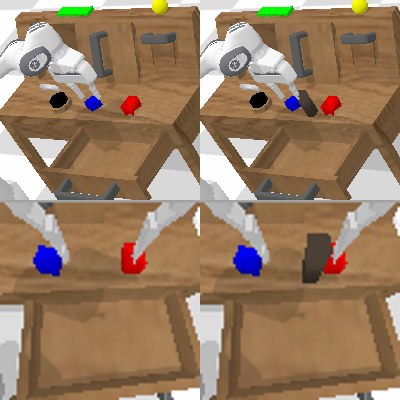} & \galimg{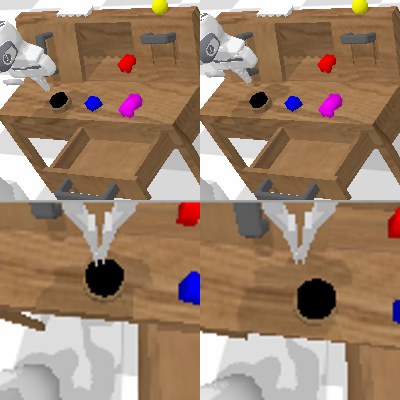} & \galimg{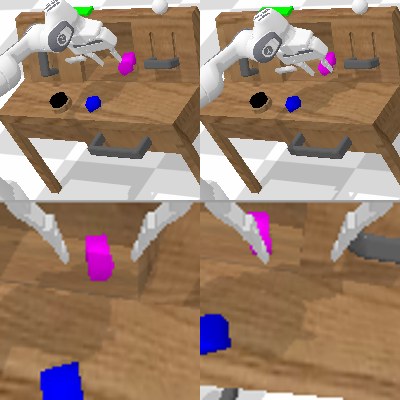} & \galimg{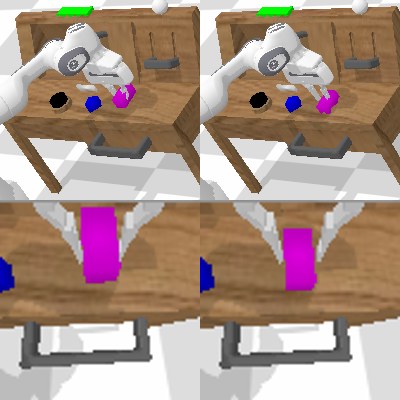} \tabularnewline
\end{tabular}
\caption{Every physical and embodiment change, with layout and selection as in \cref{fig:gallery_perc}. Target relocation moves a block by 6\,cm (A) or by 12\,cm with a rotation (B), or moves a drawer along its travel; receptacle relocation moves the stack base from under a held block (B) or shifts the drawer while a block is carried to it (A); obstacle insertion places a cube (A) or a post (B) in the approach corridor; an arm push displaces the end effector by 5 (A) or 10\,cm (B); a grasp loss releases the held block.}
\label{fig:gallery_phys}
\end{figure*}

\begin{figure*}[p]
\centering
\setlength{\tabcolsep}{2pt}
\renewcommand{\arraystretch}{0.6}
\begin{tabular}{@{}r *{5}{p{\qualw}}@{}}
& \centering\footnotesize Camera shift & \centering\footnotesize Sensor-noise onset & \centering\footnotesize Illumination switch & \centering\footnotesize Visual-theme switch & \centering\footnotesize\arraybackslash Distractor burst \\
\armlabel{\Base} & \qualmark{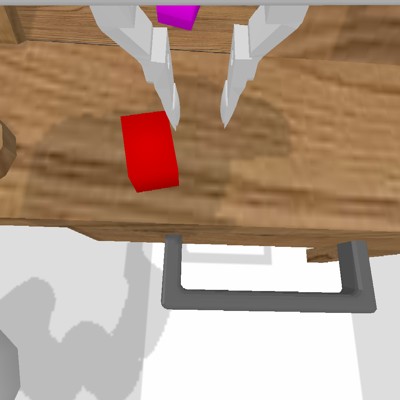}{\cmark} & \qualmark{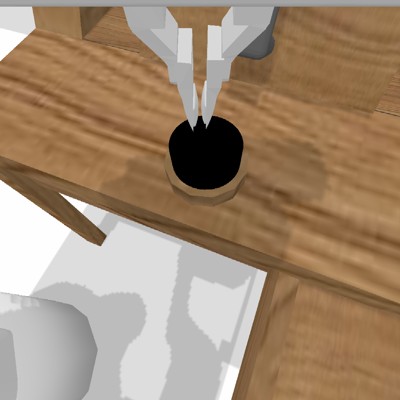}{\cmark} & \qualmark{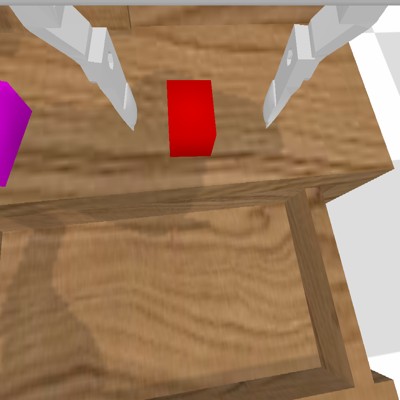}{\cmark} & \qualmark{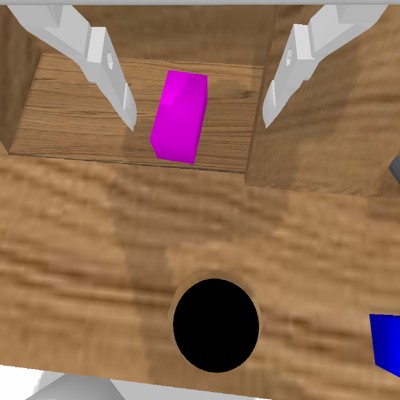}{\cmark} & \qualmark{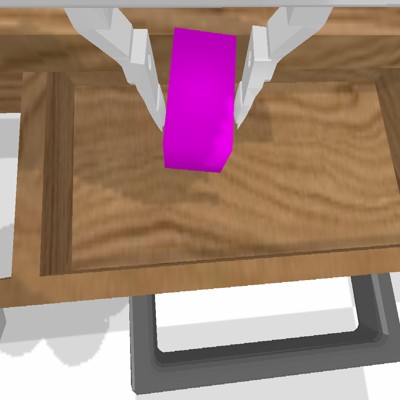}{\cmark} \\
\armsep
\armlabel{\Pre} & \qualmark{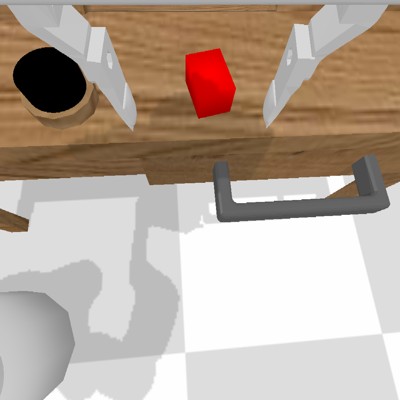}{\cmark} & \qualmark{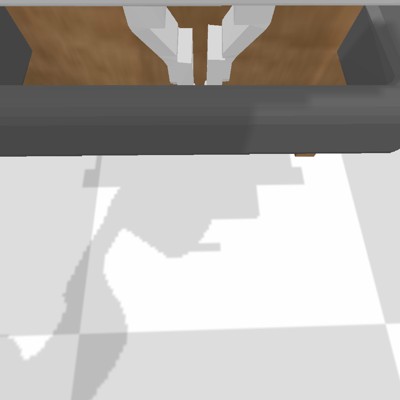}{\xmark} & \qualmark{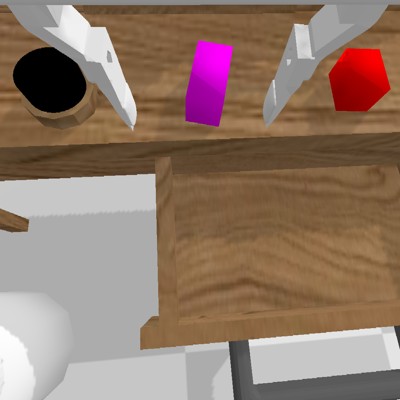}{\xmark} & \qualmark{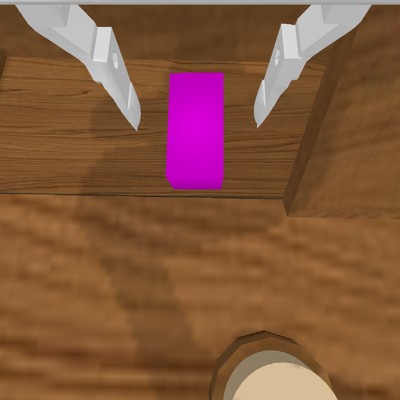}{\cmark} & \qualmark{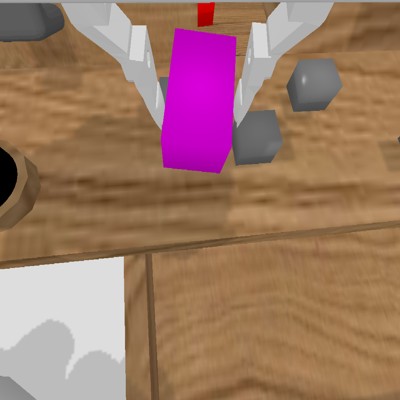}{\xmark} \\
\armsep
\armlabel{\Mid} & \qualmark{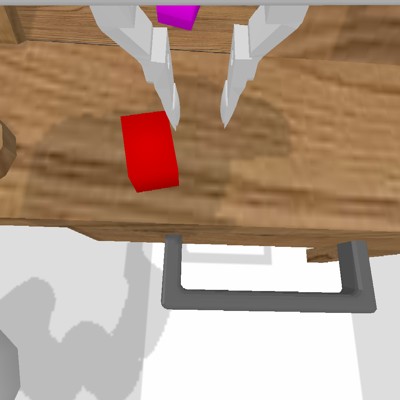}{\cmark} & \qualmark{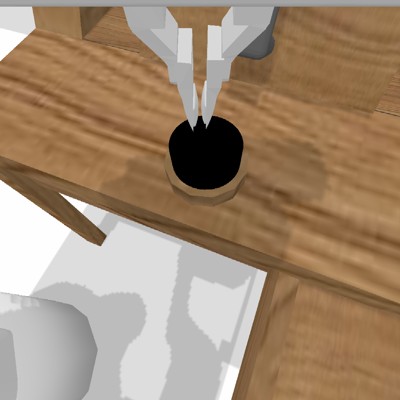}{\cmark} & \qualmark{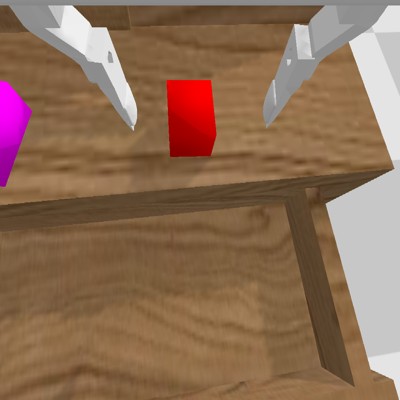}{\cmark} & \qualmark{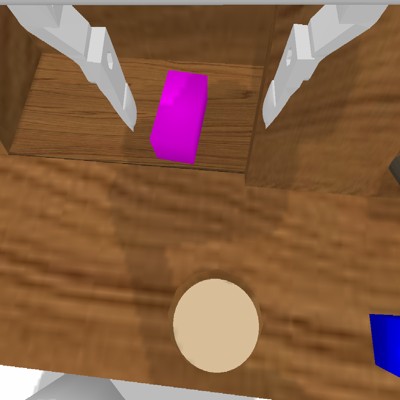}{\cmark} & \qualmark{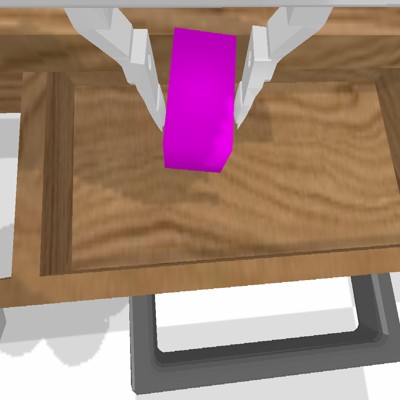}{\cmark} \\
\end{tabular}
\caption{The three arms of one case for each perceptual change. Each arm shows the static camera above the gripper camera; \Base and \Mid are shown at the onset, and \Pre at the start of instruction $k$ in the same changed world. The task and scene geometry are unchanged. Frames are re-rendered from the stored actions of held-out FLOWER cases with variant A in which \Base completed the affected instruction. Where one exists, the case shown has \Mid completing and \Pre failing the affected instruction, so the panels illustrate the timing benefit rather than typical outcomes. The mark in each frame gives that arm's outcome on the affected instruction.}
\label{fig:qual_perc}
\end{figure*}

\begin{figure*}[p]
\centering
\setlength{\tabcolsep}{2pt}
\renewcommand{\arraystretch}{0.6}
\begin{tabular}{@{}r *{5}{p{\qualw}}@{}}
& \centering\footnotesize Target relocation & \centering\footnotesize Receptacle relocation & \centering\footnotesize Obstacle insertion & \centering\footnotesize Arm push & \centering\footnotesize\arraybackslash Grasp loss \\
\armlabel{\Base} & \qualmark{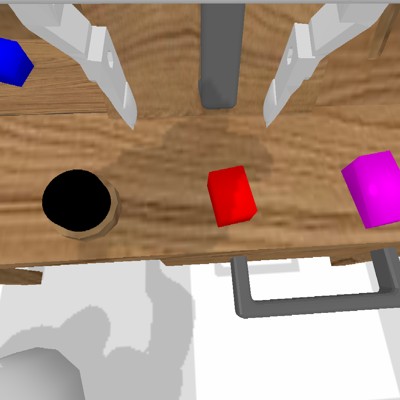}{\cmark} & \qualmark{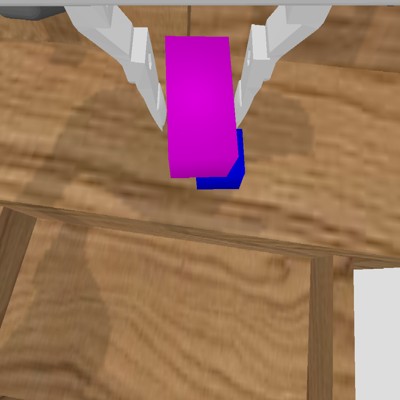}{\cmark} & \qualmark{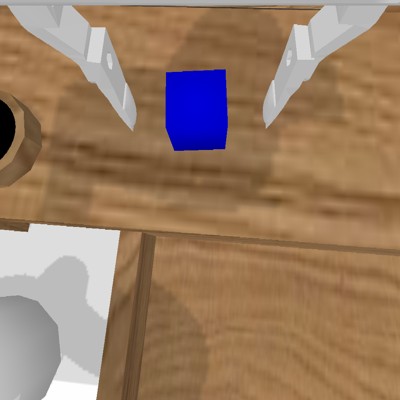}{\cmark} & \qualmark{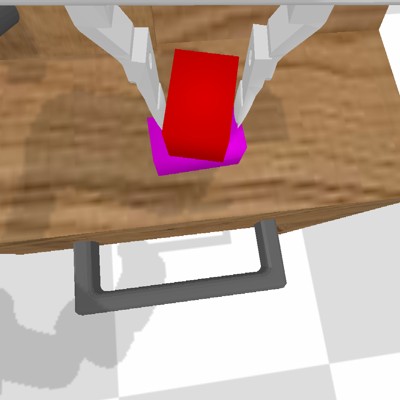}{\cmark} & \qualmark{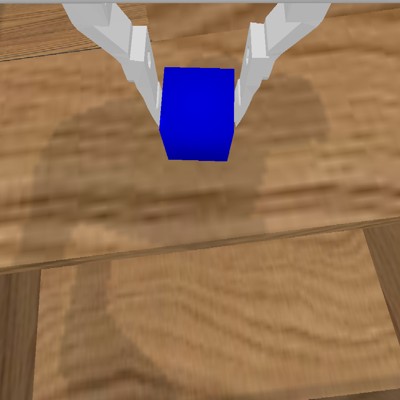}{\cmark} \\
\armsep
\armlabel{\Pre} & \qualmark{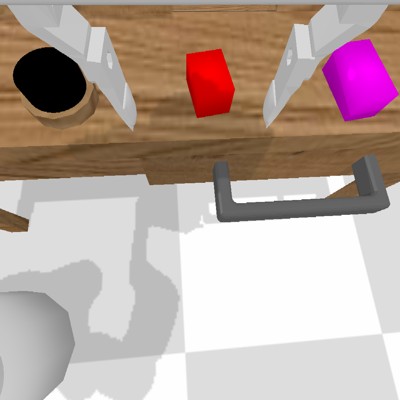}{\xmark} & \qualmark{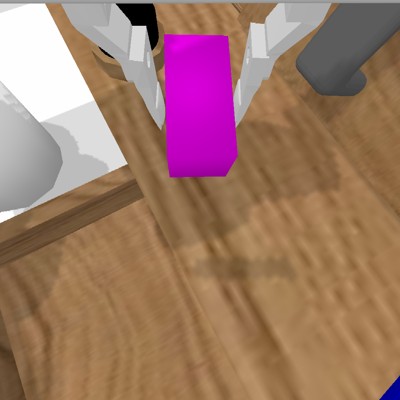}{\cmark} & \qualmark{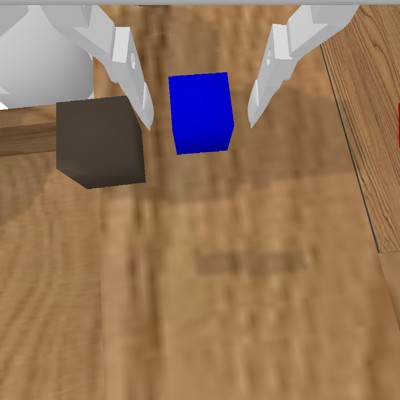}{\cmark} & \qualmark{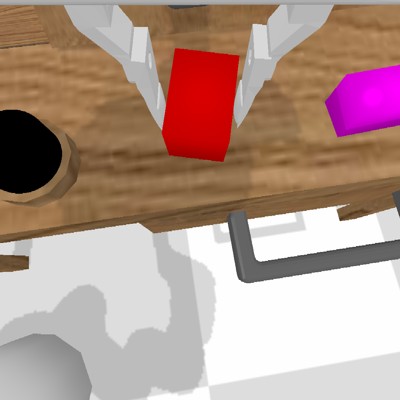}{\cmark} & \qualmark{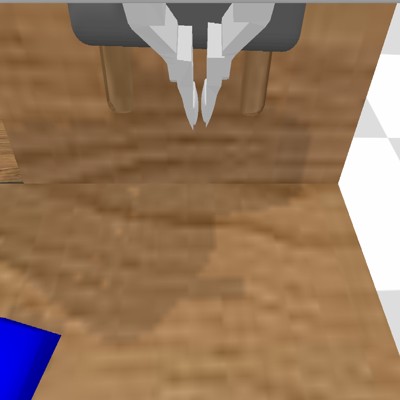}{\cmark} \\
\armsep
\armlabel{\Mid} & \qualmark{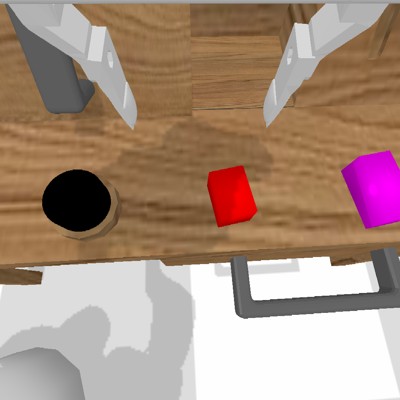}{\xmark} & \qualmark{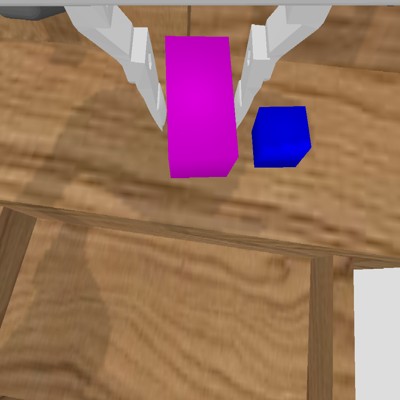}{\xmark} & \qualmark{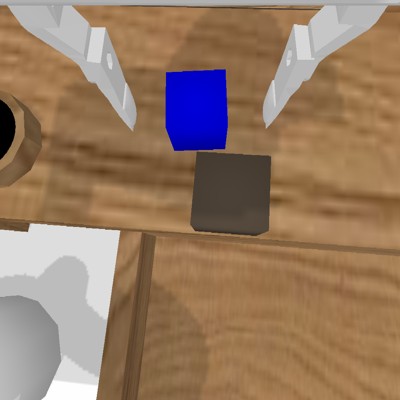}{\cmark} & \qualmark{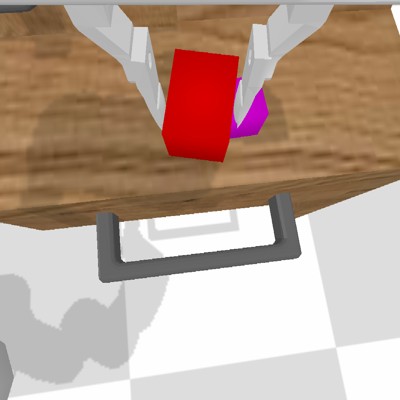}{\xmark} & \qualmark{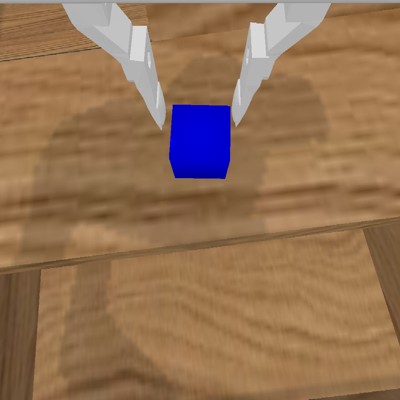}{\cmark} \\
\end{tabular}
\caption{The three arms of one case for each physical and embodiment change, with the layout of \cref{fig:qual_perc}. The instruction is fixed while the object, the path or the state of the robot changes. Cases are chosen and rendered as in \cref{fig:qual_perc}, except that, where one exists, the case shown has \Mid failing the affected instruction.}
\label{fig:qual_phys}
\end{figure*}

\newcommand{\sqc}[1]{\IfFileExists{figures/qual/#1.jpg}{\includegraphics[width=\linewidth]{figures/qual/#1.jpg}\qualgrip{#1}}{}}

\begin{figure*}[p]
\centering
\setlength{\tabcolsep}{2pt}
\renewcommand{\arraystretch}{0.6}
\begin{tabular}{@{}r *{5}{p{\qualw}}@{}}
& \multicolumn{5}{c}{\footnotesize\textcolor{perc}{\textbf{Sensor-noise onset}}, \emph{turn off the LED}} \\
& \centering\footnotesize $t_0$ & \centering\footnotesize $+15$ & \centering\footnotesize $+30$ & \centering\footnotesize $+60$ & \centering\footnotesize\arraybackslash $+100$ \\
\armlabel{\Base} & \sqc{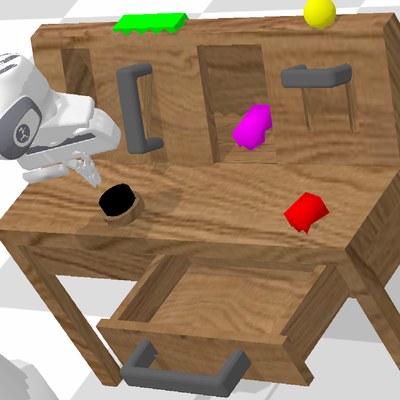} & \sqc{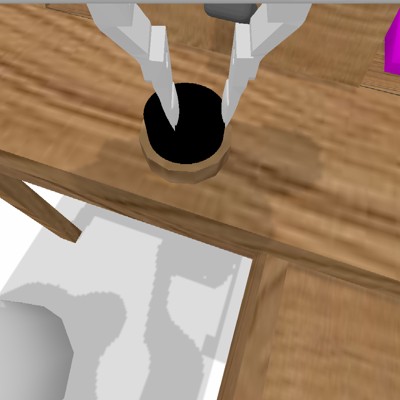} & \sqc{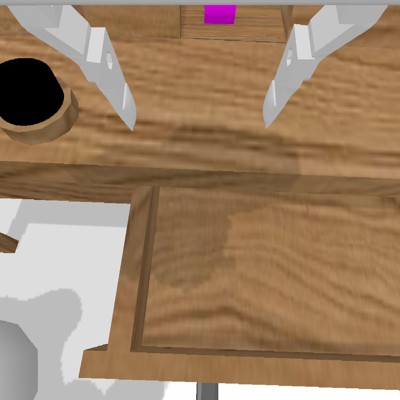} & \sqc{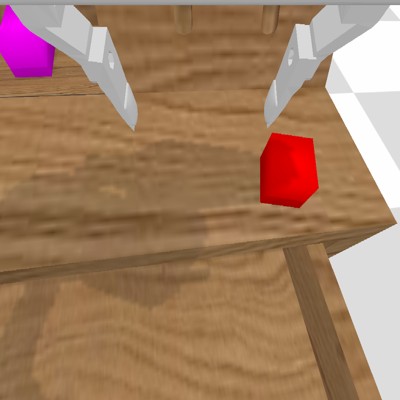} & \sqc{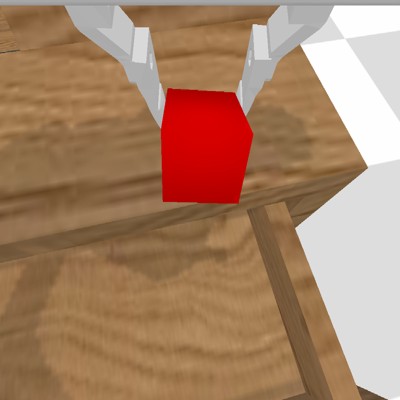} \\
\armsep
\armlabel{\Pre} & \sqc{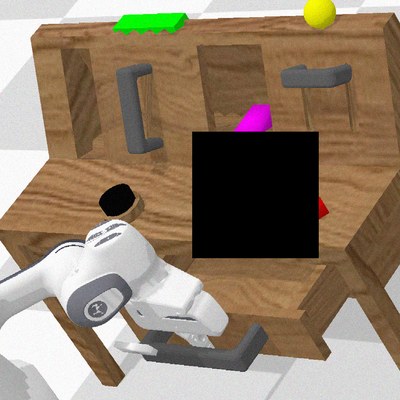} & \sqc{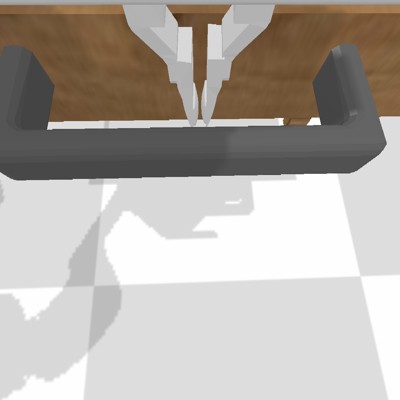} & \sqc{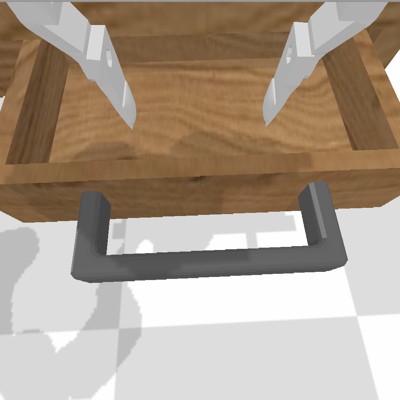} & \sqc{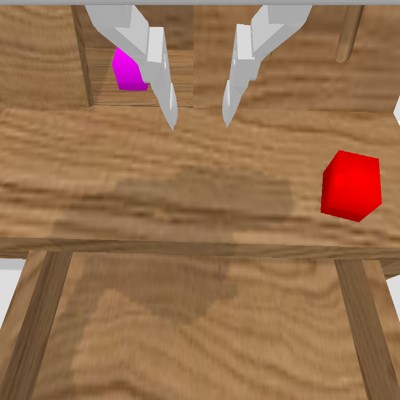} & \sqc{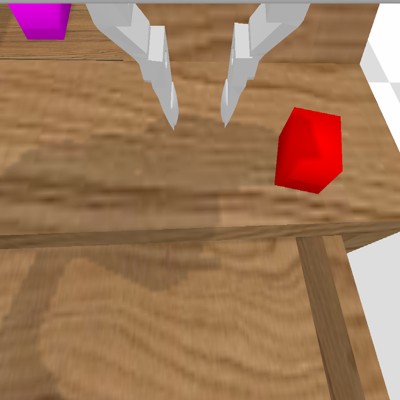} \\
\armsep
\armlabel{\Mid} & \sqc{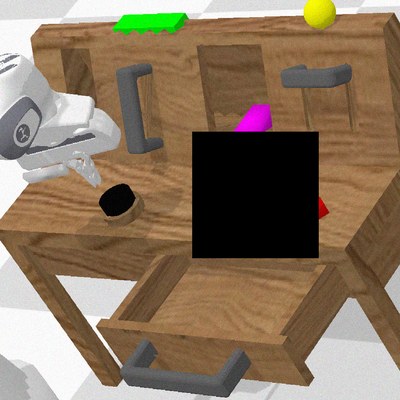} & \sqc{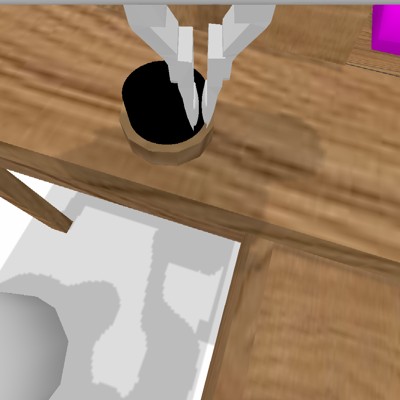} & \sqc{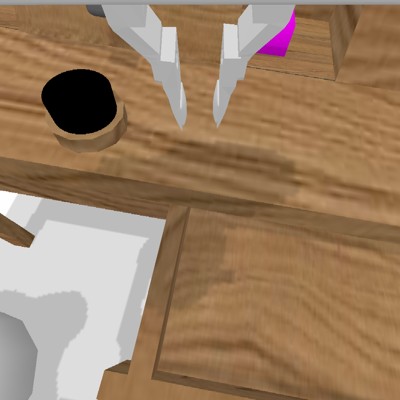} & \sqc{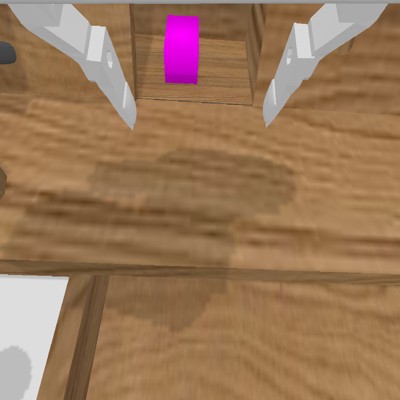} & \sqc{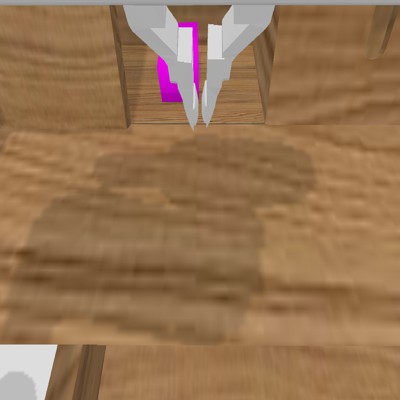} \\
\end{tabular}
\caption{What each arm does after a perceptual change, in one held-out FLOWER case. Each arm shows the static camera above the gripper camera, $t$ steps after $t_0$, which is the \Mid onset for \Base and \Mid and the start of the instruction for \Pre. \Mid, already at the switch when the view is corrupted, turns the LED off within 15 steps; \Pre, starting the instruction in the corrupted view, has not turned it off after 100 steps and fails the instruction.}
\label{fig:qual_seq}
\end{figure*}

\begin{figure*}[p]
\centering
\setlength{\tabcolsep}{2pt}
\renewcommand{\arraystretch}{0.6}
\begin{tabular}{@{}r *{5}{p{\qualw}}@{}}
& \multicolumn{5}{c}{\footnotesize\textcolor{phys}{\textbf{Receptacle relocation}}, \emph{stack the block}} \\
& \centering\footnotesize $t_0$ & \centering\footnotesize $+15$ & \centering\footnotesize $+30$ & \centering\footnotesize $+60$ & \centering\footnotesize\arraybackslash $+100$ \\
\armlabel{\Base} & \sqc{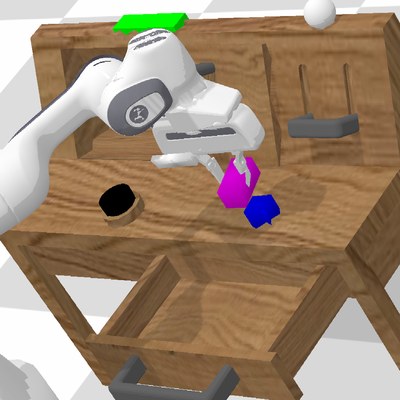} & \sqc{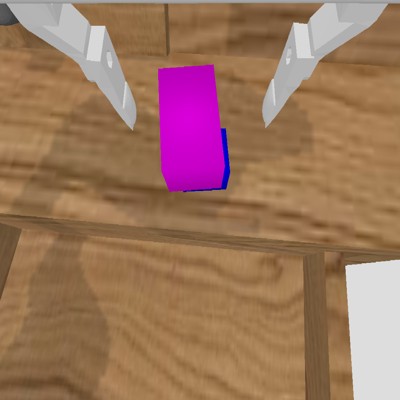} & \sqc{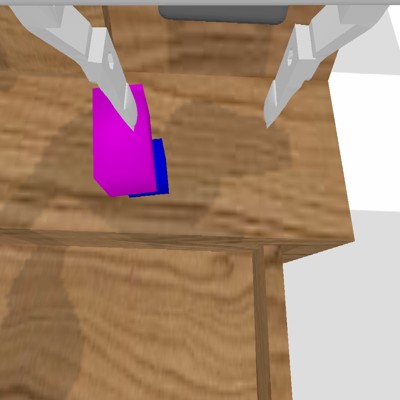} & \sqc{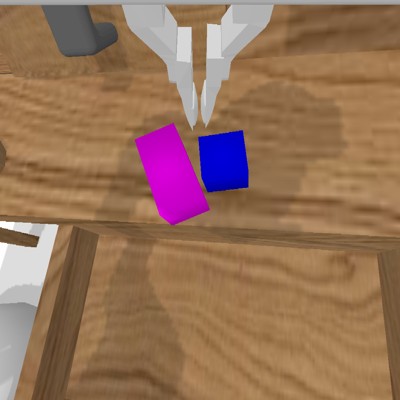} & \sqc{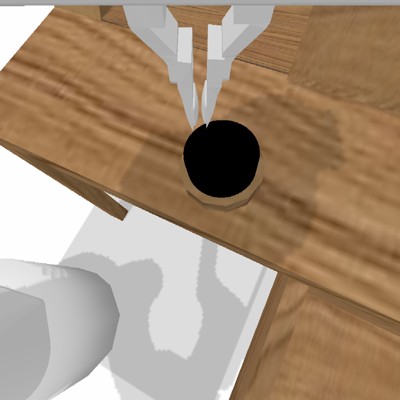} \\
\armsep
\armlabel{\Pre} & \sqc{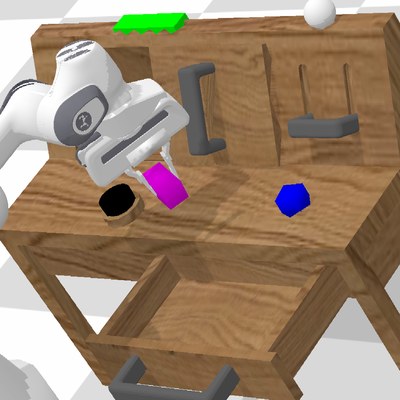} & \sqc{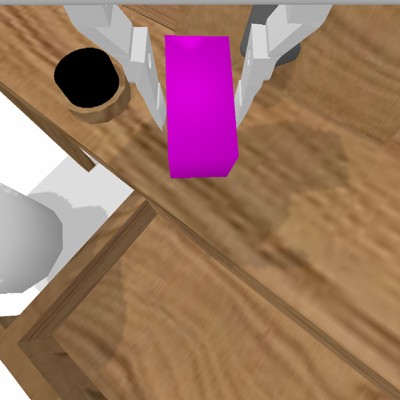} & \sqc{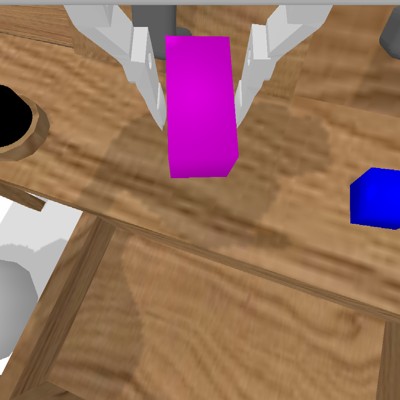} & \sqc{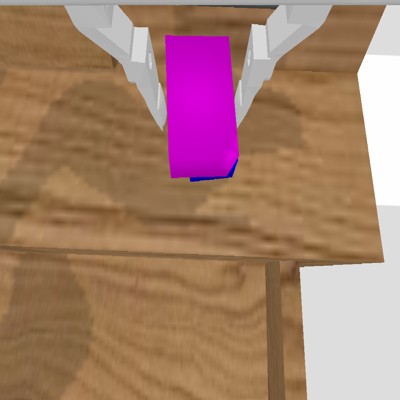} & \sqc{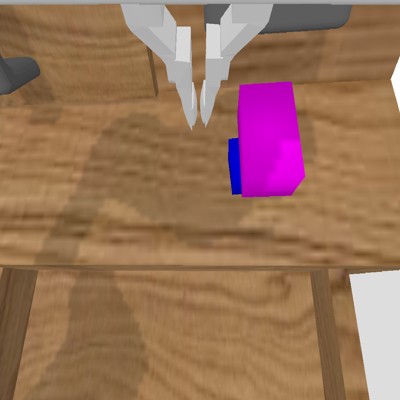} \\
\armsep
\armlabel{\Mid} & \sqc{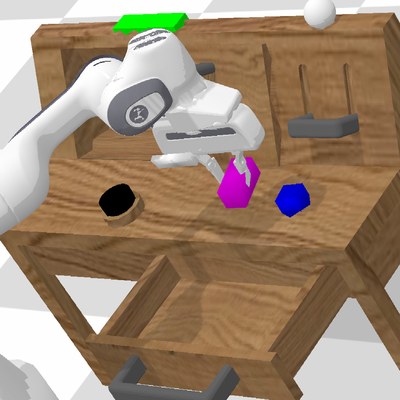} & \sqc{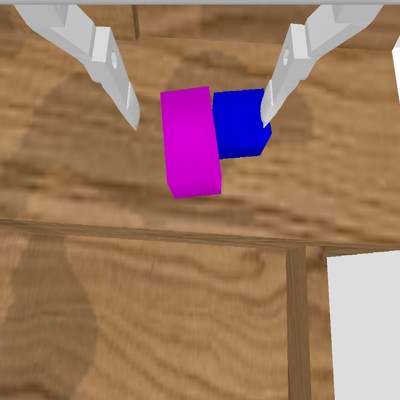} & \sqc{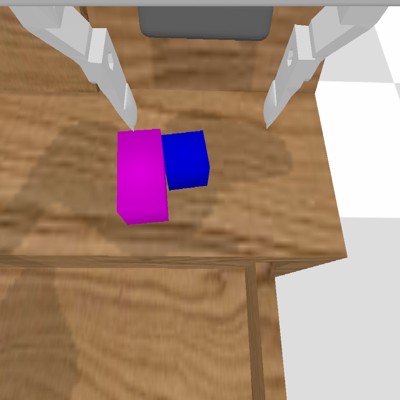} & \sqc{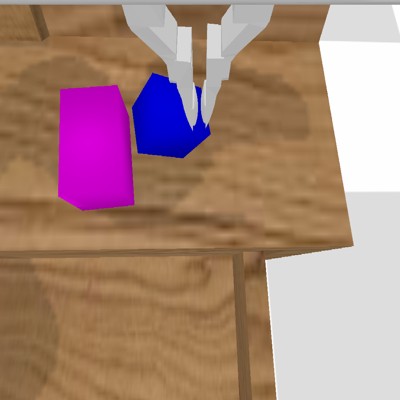} & \sqc{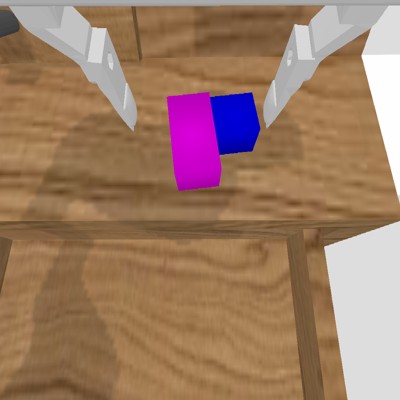} \\
\end{tabular}
\caption{What each arm does after a physical change, in one held-out FLOWER case, with the layout of \cref{fig:qual_seq}. \Mid sets the held block down beside the moved receptacle; \Pre, which sees the receptacle's new position from the start, carries the block to it and stacks it. The case was chosen, as in \cref{fig:qual_phys}, for a \Mid failure and is not typical: FLOWER completes 86\% of its 50 block-receptacle \Mid arms and reaches the old pose in 2 of them (\cref{tab:supp_behaviour}).}
\label{fig:qual_seq_phys}
\end{figure*}

\section{CALVIN-MAX Lite}
\label{sec:supp_lite}

\subsection{Outcome-independent subset}
\label{sec:supp_lite_sel}
Lite is a fixed set of 672 Max cases: six per combination, selected from the list of cases with seed \texttt{20261029} using only the change, variant, position and track (\cref{tab:supp_alloc}). The headline Lite analyses use the 486 core and embodiment cases; every policy runs the same cases, at about one tenth of Max's rollouts. Because Lite informed the hypotheses in \cref{sec:supp_prereg}, Max-A excludes its cases.

\subsection{Agreement with Max-A}
\label{sec:supp_lite_agree}
\begin{figure*}[!tp]
\centering
\includegraphics[width=\linewidth]{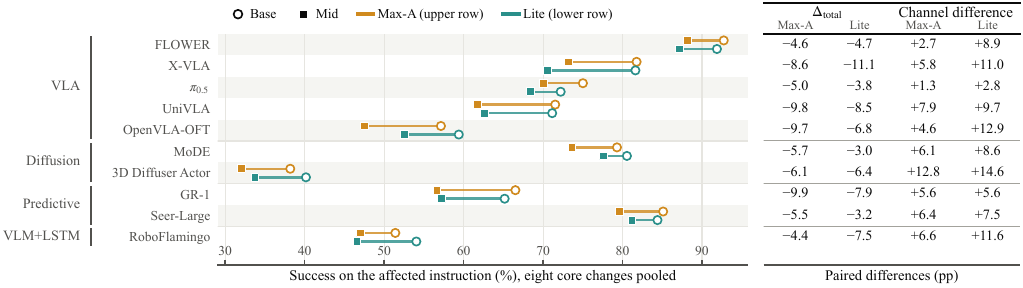}
\vspace{-5mm}
\caption{Success on Lite and on the held-out cases for each policy. Each policy has an upper row for the held-out Max-A cases and a lower row for Lite; open circles are \Base and squares \Mid, and the two sets share no case. The columns on the right give $\dtot$ and the difference in $\dtim$ between channels on each set, in pp. Lite reproduces the losses on the held-out cases but overstates the difference between channels.}
\label{fig:supp_lite_max}
\end{figure*}

\Cref{tab:supp_lite_max} compares Lite with the held-out Max-A cases on every policy, as \lmax does for its 800-pair Lite track. The two sets share no case. \Base and \Mid success agree within about five points. The difference in $\dtim$ between channels keeps its sign for every policy, but it is smaller on Max-A for nine of the ten (equal for GR-1): the Lite estimates that shaped the hypotheses were inflated, as expected when hypotheses are chosen on the same cases. Averaged over the ten policies, this difference is \lite{9.3} points on Lite (\lite{8.9} over the nine whose Lite runs had finished by the freeze, all but OpenVLA-OFT) and \maxa{6.0} on Max-A, and on both sets it is smallest for $\pi_{0.5}$ and largest for 3D Diffuser Actor (\cref{fig:supp_lite_max}). The two sets do share chains: 204 of the 385 Max-A chains also carry a Lite case. On the Max-A cases of the other 181 chains (exploratory), the pooled difference between channels is \maxa{+4.8} \maxa{[+2.0, +7.6]} points, against \maxa{+6.0} \maxa{[+4.1, +7.9]} on all Max-A cases, and the physical $\dtim$ is \maxa{+0.4} \maxa{[$-$2.1, +2.7]}.

\begin{table}[!htbp]
\caption{Lite and the held-out cases, policy by policy. For the held-out Max-A cases and for Lite, the table gives success (\%) on the affected instruction over the core changes, $\dtot$, and the difference in $\dtim$ between channels (Diff., pp).}
\label{tab:supp_lite_max}
\centering
\footnotesize
\setlength{\tabcolsep}{2.2pt}
\begin{tabular}{@{}l rrrr rrrr@{}}
\toprule
& \multicolumn{4}{c}{\tableheader Max-A} & \multicolumn{4}{c}{\tableheader Lite} \\
\cmidrule(lr){2-5}\cmidrule(l){6-9}
\tableheader Policy & \Base & \Mid & $\dtot$ & Diff. & \Base & \Mid & $\dtot$ & Diff. \\
\midrule
FLOWER & 92.7 & 88.2 & $-$4.6 & +2.7 & 91.9 & 87.2 & $-$4.7 & +8.9 \\
X-VLA & 81.8 & 73.2 & $-$8.6 & +5.8 & 81.6 & 70.5 & $-$11.1 & +11.0 \\
$\pi_{0.5}$ & 75.0 & 70.0 & $-$5.0 & +1.3 & 72.2 & 68.4 & $-$3.8 & +2.8 \\
UniVLA & 71.5 & 61.7 & $-$9.8 & +7.9 & 71.2 & 62.6 & $-$8.5 & +9.7 \\
OpenVLA-OFT & 57.2 & 47.5 & $-$9.7 & +4.6 & 59.4 & 52.6 & $-$6.8 & +12.9 \\
MoDE & 79.3 & 73.6 & $-$5.7 & +6.1 & 80.6 & 77.6 & $-$3.0 & +8.6 \\
3D Diffuser Actor & 38.2 & 32.1 & $-$6.1 & +12.8 & 40.2 & 33.8 & $-$6.4 & +14.6 \\
GR-1 & 66.5 & 56.6 & $-$9.9 & +5.6 & 65.2 & 57.3 & $-$7.9 & +5.6 \\
Seer-Large & 85.1 & 79.6 & $-$5.5 & +6.4 & 84.4 & 81.2 & $-$3.2 & +7.5 \\
RoboFlamingo & 51.4 & 47.0 & $-$4.4 & +6.6 & 54.1 & 46.6 & $-$7.5 & +11.6 \\
 
\bottomrule
\end{tabular}
\end{table}

\section{Evaluation Protocol and Policy Details}
\label{sec:supp_protocol}

\subsection{Metrics and statistical comparisons}
\label{sec:supp_metrics}
The primary outcome is success on the affected instruction. We report \Base, \Pre and \Mid success over all assigned cases and the three differences of \cref{eq:decomp}; \lmax's metrics (paired transitions, conditional regression, trigger and response coverage) are computed for \Mid against \Base. Coverage is the share of assigned cases in which the change was realized. Comparisons of $\dtim$ between channels use realized cases. Down the chain we report the position profile, paired chain length and the \emph{cost to the next instruction}: success on instruction $k{+}1$ among cases in which the affected instruction was completed in both \Base and \Mid. Intervals are 95\% intervals from a bootstrap that resamples whole chains (a chain-cluster bootstrap), because cases on one chain share a \Base rollout; pooled estimates weight policies equally and resample chains jointly, since policies share chains. Intervals are percentiles of this bootstrap with seed 0: 5,000 draws for pooled estimates and the confirmatory tests, 2,000 for each policy's per-change summaries and 1,000 for the hazard model. The confirmatory tests (\cref{sec:supp_prereg}) use the same draws: a one-sided $p$ is the share of draws on the wrong side of zero, so the smallest value it can show is below 1 in 5,000, and per-policy tests are Holm-corrected over the ten policies within each hypothesis, with no correction across hypotheses. Two tables use exact two-sided McNemar tests instead, which treat cases as independent: the bold cells of \cref{tab:supp_events}, Holm-corrected over the eight changes of a policy, and the revised instructions of \cref{tab:supp_intent}, uncorrected.

\subsection{Released checkpoints and reproduction}
\label{sec:supp_policies}
Each policy is served by an adapter that reproduces its reference evaluation code: preprocessing, action space (relative or absolute end-effector actions), chunk length and sampling steps. A policy is admitted when its \Base average chain length on the 1,000 chains is close to its published ABC$\rightarrow$D value; we treat a gap under 0.1 as a match and report larger gaps in \cref{tab:supp_policies}. GR-1 exceeds its published value by 0.14 under its reference protocol. All policies use CALVIN's limit of 360 steps per instruction. Three published values use longer limits: X-VLA (720), $\pi_{0.5}$ (480; its RLinf~\cite{yu2025rlinf} checkpoint) and 3D Diffuser Actor (1,200). At its native 1,200 steps our 3D Diffuser Actor adapter reaches \lite{3.62} on the first 100 chains (standard error about 0.15), against the published 3.27, so most of its lower score here comes from the step limit; this check covers 100 chains and is not the published evaluation, so it cannot attribute the whole gap. MoDE's published value is from its checkpoint's own evaluation log. OpenVLA-OFT, RLinf's CALVIN checkpoint trained on ABC, has no published ABC$\rightarrow$D value, so it is admitted on its own \Base. $H$ is the number of actions returned by one query and $Q$ the number executed before the next observation; every policy keeps its released settings.

\begin{table}[!tp]
\caption{The evaluated checkpoints. In the input column, s is the static camera and g the gripper camera. $H$ is the number of actions returned by one query and $Q$ the number executed before the next observation. A $\ddagger$ marks a published value obtained with a longer step limit per instruction, given in parentheses, and -- a policy with no published ABC$\rightarrow$D value. Ours is the \Base average chain length on the 1,000 chains at 360 steps.}
\label{tab:supp_policies}
\centering
\footnotesize
\setlength{\tabcolsep}{3pt}
\renewcommand{\arraystretch}{1.08}
\begin{adjustbox}{max width=\columnwidth}
\begin{tabular}{@{}ll rr rr@{}}
\toprule
\tableheader Policy & \tableheader Input & \tableheader $H$ & \tableheader $Q$ & \tableheader Published & \tableheader Ours \\
\midrule
FLOWER & RGB s+g & 10 & 10 & 4.53 & 4.531 \\
X-VLA & RGB s+g & 30 & 20 & 4.22$^{\ddagger}$\,(720) & 4.099 \\
$\pi_{0.5}$ & RGB s+g, state & 5 & 5 & 3.84$^{\ddagger}$\,(480) & 3.734 \\
UniVLA & RGB s & 12 & 12 & 3.80 & 3.729 \\
OpenVLA-OFT & RGB s & 1 & 1 & -- & 2.845 \\
MoDE & RGB s+g & 10 & 10 & 4.05 & 4.048 \\
3D Diffuser Actor & RGB-D s+g & 19 & 19 & 3.27$^{\ddagger}$\,(1200) & 1.928 \\
GR-1 & RGB s+g, 10-frame hist. & 1 & 1 & 3.06 & 3.204 \\
Seer-Large & RGB s+g, 10-frame hist. & 1 & 1 & 4.30 & 4.274 \\
RoboFlamingo & RGB s+g, 12-step hist. & 1 & 1 & 2.48 & 2.481 \\
 
\bottomrule
\end{tabular}\end{adjustbox}
\end{table}

\subsection{Noise floor and training seeds}
\label{sec:supp_seeds}

\paragraph{Null control.} Replaying to the onset and changing nothing must reproduce \Base. \Cref{tab:supp_null} gives two checks. All ten policies ran Lite's 30 null-control cases. \Base, \Pre and \Mid agree on every case for nine of them, and \Mid agrees with \Base for all ten ($\Delta{=}0$). For UniVLA one \Pre run completes the affected instruction that \Base and \Mid fail, so on its 30 cases \Pre$-$\Base is $+$3.3 points and \Mid$-$\Pre $-$3.3. Of the 478 null \Pre and \Mid runs that reached the onset across the ten policies (the other cases count with their \Base outcome), this is the only one whose affected-instruction outcome differs from \Base. It describes replay discrepancies on these cases, not a bound on the error of runs in which something changes. Every policy also passed a null-control check before admission: null \Mid runs that resume in the middle of a chunk and null \Pre runs that resume at the start of an instruction, each compared with \Base on the simulator state at every step and on the outcome of every instruction. These checks were run at onboarding and differ in size, from 8 to 40 runs per policy. Of 293 such runs, 283 are bit-identical and 290 keep the outcome of every instruction. The other ten diverge after the collision-detection pass that follows every change in \Pre and \Mid; repeated without the pass, all ten are bit-identical. The pass belongs to the single code path that applies every change, the null control included, because contacts are recomputed only by such a pass. It runs in both arms that receive the change, though at different moments, so it need not cancel exactly in \Mid$-$\Pre, and it adds a small perturbation to \Pre$-$\Base and \Mid$-$\Base.

\begin{table}[!htbp]
\caption{The null control, which replays the run to the onset and changes nothing. The left columns give \Base, \Pre and \Mid success (\%) on the 30 Lite null-control cases and $\dtot$ (pp). The right columns give the check each policy passed before admission: the null runs that are bit-identical to \Base over the whole chain (Identical) and those that keep the outcome of every instruction (Outcomes), out of the runs in each policy's check. \Mid reproduces \Base on the Lite cases for all ten policies.}
\label{tab:supp_null}
\centering
\footnotesize
\setlength{\tabcolsep}{2.4pt}
\begin{tabular}{@{}l r c r r r@{}}
\toprule
& \multicolumn{3}{c}{\tableheader Lite null track} & \multicolumn{2}{c}{\tableheader Null check} \\
\cmidrule(lr){2-4}\cmidrule(l){5-6}
\tableheader Policy & $n$ & \Base / \Pre / \Mid & $\dtot$ & \tableheader Identical & \tableheader Outcomes \\
\midrule
FLOWER & 30 & 96.7 / 96.7 / 96.7 & +0.0 & 8/8 & 8/8 \\
X-VLA & 30 & 86.7 / 86.7 / 86.7 & +0.0 & 8/8 & 8/8 \\
$\pi_{0.5}$ & 30 & 83.3 / 83.3 / 83.3 & +0.0 & 37/38 & 38/38 \\
UniVLA & 30 & 83.3 / 86.7 / 83.3 & +0.0 & 37/40 & 39/40 \\
OpenVLA-OFT & 30 & 66.7 / 66.7 / 66.7 & +0.0 & 37/38 & 38/38 \\
MoDE & 30 & 90.0 / 90.0 / 90.0 & +0.0 & 12/12 & 12/12 \\
3D Diffuser Actor & 30 & 46.7 / 46.7 / 46.7 & +0.0 & 32/34 & 34/34 \\
GR-1 & 30 & 70.0 / 70.0 / 70.0 & +0.0 & 35/38 & 36/38 \\
Seer-Large & 30 & 96.7 / 96.7 / 96.7 & +0.0 & 39/39 & 39/39 \\
RoboFlamingo & 30 & 50.0 / 50.0 / 50.0 & +0.0 & 38/38 & 38/38 \\
 
\bottomrule
\end{tabular}
\end{table}

\paragraph{Training seeds.} MoDE's three released training seeds run the same 486 Lite cases (\cref{tab:supp_seeds}). The perceptual timing effect and the difference between channels are positive for every seed and vary by 3 to 4 points between seeds, about half the effect; the physical timing effect ranges from +0.9 to $-$5.7 points, and every seed's interval includes zero. \lmax reports one seed per policy.

\begin{table}[!tp]
\caption{The timing effect $\dtim$ by channel for each of MoDE's three released training seeds on Lite, in pp with 95\% intervals. The perceptual effect and the difference between channels are positive for every seed.}
\label{tab:supp_seeds}
\centering
\footnotesize
\setlength{\tabcolsep}{3pt}
\begin{adjustbox}{max width=\columnwidth}
\begin{tabular}{@{}lccc@{}}
\toprule
\tableheader Seed & \tableheader Perceptual & \tableheader Physical & \tableheader Difference \\
\midrule
1 & \tci{+9.4}{+5.1}{+13.9} & \tci{+0.9}{$-$7.2}{+9.0} & \tci{+8.6}{$-$0.6}{+18.0} \\
2 & \tci{+6.8}{+3.0}{+10.8} & \tci{$-$0.9}{$-$7.7}{+6.2} & \tci{+7.7}{$-$0.0}{+15.3} \\
3 & \tci{+6.2}{+2.5}{+10.0} & \tci{$-$5.7}{$-$13.4}{+2.0} & \tci{+11.8}{+3.1}{+20.5} \\
\midrule
Pooled & \tci{+7.5}{+4.7}{+10.4} & \tci{$-$1.9}{$-$5.8}{+2.0} & \tci{+9.4}{+4.4}{+14.5} \\
 
\bottomrule
\end{tabular}\end{adjustbox}
\end{table}

\section{Paired Outcomes and Breakdowns}
\label{sec:supp_breakdowns}

\subsection{Paired outcomes and coverage}
\label{sec:supp_outcomes}
\begin{figure}[!tp]
\centering
\includegraphics[width=\linewidth]{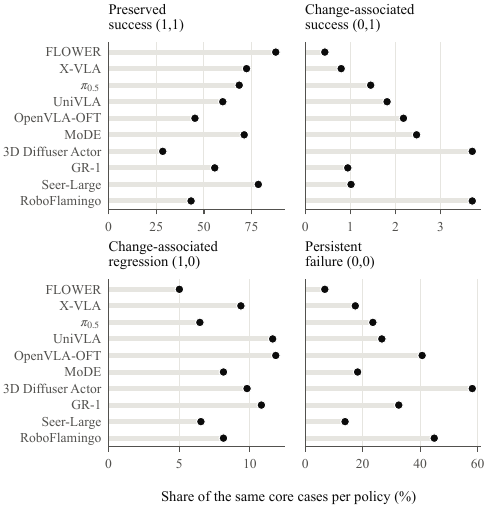}
\caption{How the core cases of each policy split into the four paired outcomes of \Mid against \Base, on the held-out cases (\maxa{1,377} per policy). The split separates change-associated regression (event-associated regression in \lmax) from low \Base success. The panels use separate horizontal scales so that rare change-associated successes $(0,1)$ remain visible.}
\label{fig:outcomes}
\end{figure}

\begin{figure*}[!tp]
\centering
\includegraphics[width=\linewidth]{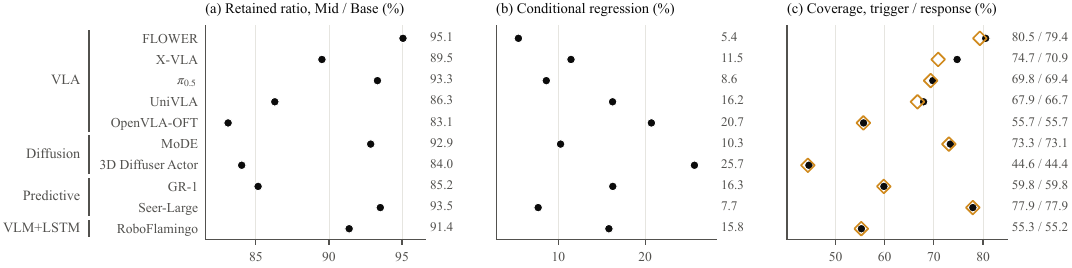}
\caption{Three diagnostics for each policy on the held-out core cases. Panel (a) gives the retained ratio, \Mid success over \Base success, and panel (b) the conditional regression, the share of \Base successes that fail in \Mid. Panel (c) gives the trigger coverage, the share of assigned cases in which the change was realized (filled), and the response coverage, the share in which the policy was also queried at least once after the onset within the affected instruction (hollow). The panels use separate horizontal scales. Most of each policy's \Base successes survive the change, but the share lost varies almost fivefold.}
\label{fig:diagnostics}
\end{figure*}

\Cref{fig:outcomes} partitions the core cases of each policy into the four paired outcomes of \Mid against \Base, and \cref{tab:supp_transitions} gives the counts. \Cref{fig:diagnostics} adds three per-policy diagnostics: the retained ratio (\Mid success divided by \Base success, including change-associated successes), conditional regression (the share of \Base successes that fail in \Mid), and coverage (the share of assigned cases in which the change was realized). Unreached and unrealizable cases count toward every success rate. 3D Diffuser Actor's low coverage reflects its low \Base success at 360 steps: chains that fail before instruction $k$ never reach the onset. Pooled over the ten policies' held-out core cases, \maxa{73.6}\% of the \maxa{8,870} assigned perceptual cases and \maxa{52.1}\% of the \maxa{4,900} physical ones are realized. In both channels the chain fails before instruction $k$ in about a fifth of the cases (\maxa{19.8} and \maxa{20.1}\%; 4 to 47\% by policy), and the onset rule never fires in \maxa{6.0} and \maxa{7.6}\%. The channels differ in unrealizable cases: \maxa{0.5}\% of perceptual cases against \maxa{20.3}\% of physical ones, where \maxa{15.8}\% of all physical cases have no valid change, mostly target relocations and obstacle insertions, and \maxa{4.5}\% fail a check. Comparisons between channels therefore rest on a physical subset selected by these rules as well as by each policy's \Base run. Trigger and response coverage are shown separately in \cref{fig:diagnostics}c; they differ by at most \maxa{4} points (X-VLA, whose 20-action chunks can finish the instruction before the next query).

\begin{table}[!htbp]
\caption{Paired outcomes of \Base and \Mid on the eight core changes, on the held-out cases. Cov.\ is the share of cases in which the change was realized, the next four columns count the cases with each pair of outcomes (\Base, \Mid), and Cond.\ regr.\ is the share of \Base successes that fail in \Mid.}
\label{tab:supp_transitions}
\centering
\footnotesize
\setlength{\tabcolsep}{2.4pt}
\begin{adjustbox}{max width=\columnwidth}
\begin{tabular}{@{}lrrrrrrc@{}}
\toprule
\tableheader Policy & $n$ & \tableheader Cov.\,\% & (1,1) & (1,0) & (0,1) & (0,0) & \tableheader Cond.\ regr.\,\% \\
\midrule
FLOWER & 1377 & 80.5 & 1208 & 69 & 6 & 94 & \makebox[1.8em][r]{5.4} \\
X-VLA & 1377 & 74.7 & 997 & 129 & 11 & 240 & \makebox[1.8em][r]{11.5} \\
$\pi_{0.5}$ & 1377 & 69.8 & 944 & 89 & 20 & 324 & \makebox[1.8em][r]{8.6} \\
UniVLA & 1377 & 67.9 & 825 & 160 & 25 & 367 & \makebox[1.8em][r]{16.2} \\
OpenVLA-OFT & 1377 & 55.7 & 624 & 163 & 30 & 560 & \makebox[1.8em][r]{20.7} \\
MoDE & 1377 & 73.3 & 980 & 112 & 34 & 251 & \makebox[1.8em][r]{10.3} \\
3D Diffuser Actor & 1377 & 44.6 & 391 & 135 & 51 & 800 & \makebox[1.8em][r]{25.7} \\
GR-1 & 1377 & 59.8 & 767 & 149 & 13 & 448 & \makebox[1.8em][r]{16.3} \\
Seer-Large & 1377 & 77.9 & 1082 & 90 & 14 & 191 & \makebox[1.8em][r]{7.7} \\
RoboFlamingo & 1377 & 55.3 & 596 & 112 & 51 & 618 & \makebox[1.8em][r]{15.8} \\
 
\bottomrule
\end{tabular}\end{adjustbox}
\end{table}

\subsection{Results for each change}
\label{sec:supp_events}
\Cref{tab:supp_events} reports every change for every policy on the held-out core cases (\maxa{142} to \maxa{180} per change and policy, all assigned cases). Bold $\dtot$ cells have an exact McNemar test of \Mid against \Base below 0.05 after Holm correction over the eight changes of the policy. Conditional regression is the share of \Base successes lost in \Mid. Profiles differ by policy. X-VLA loses most under camera shift (\Base\,\maxa{82.8}, \Pre\,\maxa{49.4}, \Mid\,\maxa{63.8}), as it did in \lmax; OpenVLA-OFT, which sees only the static camera, loses most under an illumination switch (\maxa{59.0}\,$\rightarrow$\,\maxa{36.5} under \Pre), as does UniVLA; GR-1 under a visual-theme switch (\maxa{72.8}\,$\rightarrow$\,\maxa{35.0}); FLOWER and Seer-Large are nearly unaffected by camera and lighting but lose under sensor noise and theme changes, and sensor noise is the largest configuration cost for MoDE, 3D Diffuser Actor and RoboFlamingo. \Cref{tab:supp_variants} splits $\dtot$ by parameter variant. The two variants of a change have nearly the same number of assigned cases (grasp loss has one variant) and are realized about equally often, except target relocation (\maxa{66}\% of variant-A cases against \maxa{36}\% of variant B, pooled over the policies) and arm push (\maxa{78} against \maxa{70}\%); since $\dtot$ counts unrealized cases at their \Base outcome, a variant realized less often has more of its cases fixed at zero difference. The variants shift the loss by up to tens of points for one policy, most for the visual-theme switch, sensor noise and camera shift, where variant B is stronger on average; as in \lmax, a benchmark's choice of parameters matters, and every estimate keeps both variants.

\begin{table}[!htbp]
\caption{The total effect $\dtot$ for each parameter variant on the held-out core and embodiment cases, in pp and averaged over the ten policies, with the range of the per-policy difference between A and B. The variants shift the loss most for the visual-theme switch, sensor noise and camera shift, where variant B is stronger on average.}
\label{tab:supp_variants}
\centering
\footnotesize
\setlength{\tabcolsep}{3pt}
\begin{tabular}{@{}lrrc@{}}
\toprule
\tableheader Change & \tableheader A & \tableheader B & \tableheader A$-$B across policies \\
\midrule
Target relocation & $-$13.4 & $-$10.6 & \trange{$-$7.6}{+6.5} \\
Receptacle relocation & $-$3.9 & $-$0.7 & \trange{$-$12.5}{+2.9} \\
Obstacle insertion & $-$6.8 & $-$13.0 & \trange{$-$4.0}{+15.1} \\
Camera shift & $-$1.7 & $-$8.5 & \trange{$-$3.4}{+28.4} \\
Sensor-noise onset & $-$4.9 & $-$15.2 & \trange{$-$2.0}{+28.1} \\
Illumination switch & $-$4.3 & $-$5.5 & \trange{$-$5.6}{+6.7} \\
Visual-theme switch & $-$1.3 & $-$14.9 & \trange{+1.0}{+60.5} \\
Distractor burst & $-$1.6 & $-$2.0 & \trange{$-$3.4}{+8.0} \\
Arm push & $-$3.2 & $-$5.2 & \trange{$-$6.0}{+10.0} \\
 
\bottomrule
\end{tabular}
\end{table}

\begin{table*}[!tp]
\caption{Every core change for every policy on the held-out cases. Success and conditional regression are in \% and differences in pp, over all assigned cases, each with $\pm$ one standard error from the bootstrap over chains (a 95\% interval is about $\pm$2 standard errors). A bold total effect has an exact McNemar test of \Mid against \Base below 0.05 after Holm correction over the eight changes; this test treats cases as independent. Differences are computed before rounding, so a total can differ from the sum of its parts by 0.1. The sensitivity to each change varies across policies.}
\label{tab:supp_events}
\centering
\footnotesize
\setlength{\tabcolsep}{1pt}
\renewcommand{\arraystretch}{1.0}
\begin{tabular*}{\textwidth}{@{\extracolsep{\fill}}l *{10}{c}@{}}
\toprule
\tableheader Change & \tableheader FLOWER & \tableheader X-VLA & \tableheader $\pi_{0.5}$ & \tableheader UniVLA & \evhead{OpenVLA-}{OFT} & \tableheader MoDE & \evhead{3D Diffuser}{Actor} & \tableheader GR-1 & \evhead{Seer-}{Large} & \evhead{Robo-}{Flamingo} \\
\midrule
\multicolumn{11}{@{}l}{\tableheader Base success on the affected instruction (\%)} \\
\cmidrule{1-11}
Target relocation & \evse{94.3}{2.0} & \evse{79.9}{3.3} & \evse{77.6}{3.9} & \evse{70.1}{3.9} & \evse{63.8}{4.1} & \evse{86.8}{2.8} & \evse{40.2}{4.2} & \evse{70.1}{3.9} & \evse{83.3}{3.2} & \evse{46.6}{4.4} \\
Receptacle relocation & \evse{93.7}{2.4} & \evse{82.4}{4.1} & \evse{76.1}{4.4} & \evse{47.9}{5.1} & \evse{42.3}{5.0} & \evse{64.1}{5.1} & \evse{31.7}{4.8} & \evse{62.0}{5.1} & \evse{83.1}{3.9} & \evse{38.7}{4.9} \\
Obstacle insertion & \evse{94.8}{1.6} & \evse{86.2}{2.8} & \evse{75.3}{3.5} & \evse{71.8}{4.0} & \evse{55.7}{4.2} & \evse{82.2}{3.1} & \evse{40.8}{4.2} & \evse{67.2}{3.9} & \evse{83.9}{3.0} & \evse{64.9}{4.2} \\
Camera shift & \evse{94.8}{1.8} & \evse{82.8}{3.1} & \evse{73.6}{3.6} & \evse{79.9}{3.3} & \evse{62.1}{3.7} & \evse{84.5}{2.8} & \evse{36.2}{3.8} & \evse{67.8}{3.9} & \evse{89.7}{2.4} & \evse{56.9}{3.9} \\
Sensor-noise onset & \evse{90.5}{2.2} & \evse{76.5}{3.6} & \evse{76.0}{3.6} & \evse{78.8}{3.2} & \evse{55.3}{4.2} & \evse{79.9}{3.4} & \evse{39.7}{4.1} & \evse{68.7}{3.9} & \evse{91.1}{2.3} & \evse{53.6}{4.1} \\
Illumination switch & \evse{91.0}{2.3} & \evse{83.7}{3.1} & \evse{74.2}{3.4} & \evse{74.7}{3.6} & \evse{59.0}{4.1} & \evse{79.2}{3.3} & \evse{39.3}{4.1} & \evse{61.8}{4.1} & \evse{85.4}{3.0} & \evse{52.2}{4.1} \\
Visual-theme switch & \evse{91.7}{2.0} & \evse{81.7}{2.9} & \evse{72.2}{3.9} & \evse{68.3}{3.8} & \evse{61.1}{3.9} & \evse{77.8}{3.2} & \evse{38.3}{3.6} & \evse{72.8}{3.6} & \evse{85.0}{2.9} & \evse{51.7}{4.0} \\
Distractor burst & \evse{91.5}{2.2} & \evse{81.2}{3.3} & \evse{75.6}{3.7} & \evse{76.1}{3.6} & \evse{55.1}{4.1} & \evse{77.3}{3.6} & \evse{38.1}{3.9} & \evse{60.8}{4.4} & \evse{79.0}{3.6} & \evse{44.3}{4.3} \\
\midrule
\multicolumn{11}{@{}l}{\tableheader Configuration effect, Pre $-$ Base (pp)} \\
\cmidrule{1-11}
Target relocation & \evse{$-$10.9}{2.6} & \evse{$-$16.1}{3.0} & \evse{$-$14.4}{3.2} & \evse{$-$12.6}{2.8} & \evse{$-$11.5}{3.3} & \evse{$-$15.5}{3.1} & \evse{$-$8.0}{2.6} & \evse{$-$12.1}{3.1} & \evse{$-$12.6}{2.6} & \evse{$-$7.5}{2.8} \\
Receptacle relocation & \evse{$-$4.2}{1.9} & \evse{$-$6.3}{3.5} & \evse{$-$0.7}{1.2} & \evse{+4.9}{3.8} & \evse{$-$4.9}{2.8} & \evse{+6.3}{3.3} & \evse{$-$2.8}{1.9} & \evse{$-$0.7}{1.8} & \evse{$-$1.4}{3.0} & \evse{$-$4.2}{3.3} \\
Obstacle insertion & \evse{$-$8.0}{2.1} & \evse{$-$19.5}{3.4} & \evse{$-$5.2}{2.4} & \evse{$-$9.8}{3.1} & \evse{$-$5.7}{2.6} & \evse{$-$8.0}{2.6} & \evse{$-$8.0}{2.6} & \evse{$-$10.9}{2.8} & \evse{$-$9.8}{2.5} & \evse{$-$16.1}{2.9} \\
Camera shift & \evse{$-$1.1}{0.8} & \evse{$-$33.3}{3.7} & \evse{+2.3}{1.6} & \evse{$-$12.6}{2.6} & \evse{$-$22.4}{3.2} & \evse{$-$1.1}{1.1} & \evse{$-$5.2}{3.2} & \evse{$-$5.7}{2.0} & \evse{$-$2.3}{1.1} & \evse{+2.9}{2.5} \\
Sensor-noise onset & \evse{$-$12.8}{2.7} & \evse{$-$9.5}{2.6} & \evse{$-$14.0}{3.1} & \evse{$-$19.6}{3.4} & \evse{$-$21.2}{3.5} & \evse{$-$17.9}{3.0} & \evse{$-$23.5}{3.4} & \evse{$-$17.3}{3.0} & \evse{$-$10.6}{2.4} & \evse{$-$18.4}{3.8} \\
Illumination switch & \evse{$-$2.2}{1.1} & \evse{$-$1.1}{1.1} & \evse{$-$2.8}{2.0} & \evse{$-$21.9}{3.3} & \evse{$-$22.5}{3.5} & \evse{$-$4.5}{2.1} & \evse{$-$3.4}{2.3} & \evse{$-$9.6}{2.4} & \evse{$-$4.5}{1.7} & \evse{$-$5.6}{2.9} \\
Visual-theme switch & \evse{$-$13.3}{2.7} & \evse{$-$2.8}{1.2} & \evse{$-$6.7}{2.2} & \evse{$-$7.8}{2.4} & \evse{$-$5.0}{2.3} & \evse{$-$14.4}{3.1} & \evse{$-$1.7}{2.8} & \evse{$-$37.8}{3.8} & \evse{$-$12.2}{2.9} & \evse{$-$2.8}{3.3} \\
Distractor burst & \evse{$-$11.4}{2.5} & \evse{$-$8.5}{2.1} & \evse{$-$5.7}{2.0} & \evse{$-$10.8}{2.7} & \evse{$-$1.1}{2.5} & \evse{$-$6.8}{2.6} & \evse{$-$6.8}{3.0} & \evse{$-$7.4}{2.3} & \evse{$-$8.0}{2.3} & \evse{$-$1.1}{2.8} \\
\midrule
\multicolumn{11}{@{}l}{\tableheader Timing effect, Mid $-$ Pre (pp)} \\
\cmidrule{1-11}
Target relocation & \evse{+1.7}{1.3} & \evse{$-$0.6}{2.6} & \evse{+0.6}{1.9} & \evse{$-$1.1}{2.3} & \evse{+1.7}{1.9} & \evse{$-$0.6}{2.2} & \evse{$-$2.3}{2.6} & \evse{+1.1}{2.0} & \evse{+0.0}{1.4} & \evse{+0.0}{1.4} \\
Receptacle relocation & \evse{+0.0}{2.7} & \evse{+2.1}{2.3} & \evse{$-$2.1}{1.8} & \evse{$-$1.4}{2.7} & \evse{+2.1}{2.3} & \evse{$-$1.4}{2.9} & \evse{$-$6.3}{2.6} & \evse{$-$0.7}{2.1} & \evse{$-$0.7}{2.1} & \evse{$-$0.7}{2.2} \\
Obstacle insertion & \evse{+3.4}{1.8} & \evse{$-$1.1}{2.5} & \evse{+2.9}{2.4} & \evse{$-$2.3}{2.9} & \evse{$-$2.9}{2.5} & \evse{+0.0}{2.8} & \evse{$-$1.7}{2.3} & \evse{+2.3}{2.4} & \evse{$-$0.6}{2.0} & \evse{+1.1}{2.7} \\
Camera shift & \evse{+0.6}{1.0} & \evse{+14.4}{3.4} & \evse{$-$2.9}{1.5} & \evse{+5.2}{2.6} & \evse{+5.7}{3.2} & \evse{+1.7}{1.7} & \evse{+2.9}{3.0} & \evse{+0.0}{1.8} & \evse{+0.6}{1.0} & \evse{$-$1.1}{2.1} \\
Sensor-noise onset & \evse{+8.9}{2.4} & \evse{+5.0}{2.4} & \evse{+5.0}{2.7} & \evse{+2.8}{3.2} & \evse{+5.6}{2.4} & \evse{+10.6}{2.9} & \evse{+9.5}{3.0} & \evse{+6.7}{2.4} & \evse{+3.9}{2.1} & \evse{+6.7}{2.8} \\
Illumination switch & \evse{+1.1}{0.8} & \evse{+0.6}{1.0} & \evse{+0.0}{2.0} & \evse{+4.5}{3.5} & \evse{+9.0}{3.3} & \evse{+1.7}{1.8} & \evse{+3.4}{2.7} & \evse{+2.8}{2.5} & \evse{+2.8}{1.2} & \evse{+3.4}{2.2} \\
Visual-theme switch & \evse{+1.7}{1.9} & \evse{+2.8}{1.2} & \evse{+1.7}{1.5} & \evse{+0.6}{2.1} & \evse{$-$3.3}{1.9} & \evse{+1.7}{2.3} & \evse{$-$2.8}{2.8} & \evse{+8.3}{2.2} & \evse{+4.4}{2.2} & \evse{+7.2}{2.5} \\
Distractor burst & \evse{+10.2}{2.5} & \evse{+5.7}{2.2} & \evse{+2.3}{2.1} & \evse{+6.2}{2.7} & \evse{+0.6}{2.0} & \evse{+5.1}{2.2} & \evse{+7.4}{2.8} & \evse{+4.0}{2.0} & \evse{+7.4}{2.0} & \evse{+0.6}{2.6} \\
\midrule
\multicolumn{11}{@{}l}{\tableheader Total effect, Mid $-$ Base (pp)} \\
\cmidrule{1-11}
Target relocation & \evse{\textbf{$-$9.2}}{2.3} & \evse{\textbf{$-$16.7}}{3.2} & \evse{\textbf{$-$13.8}}{3.0} & \evse{\textbf{$-$13.8}}{2.7} & \evse{\textbf{$-$9.8}}{3.0} & \evse{\textbf{$-$16.1}}{3.1} & \evse{\textbf{$-$10.3}}{2.7} & \evse{\textbf{$-$10.9}}{2.9} & \evse{\textbf{$-$12.6}}{2.7} & \evse{\textbf{$-$7.5}}{2.6} \\
Receptacle relocation & \evse{$-$4.2}{2.4} & \evse{$-$4.2}{2.9} & \evse{$-$2.8}{1.4} & \evse{+3.5}{2.9} & \evse{$-$2.8}{2.9} & \evse{+4.9}{3.7} & \evse{\textbf{$-$9.2}}{3.0} & \evse{$-$1.4}{2.2} & \evse{$-$2.1}{2.9} & \evse{$-$4.9}{3.5} \\
Obstacle insertion & \evse{\textbf{$-$4.6}}{1.6} & \evse{\textbf{$-$20.7}}{3.4} & \evse{$-$2.3}{2.1} & \evse{\textbf{$-$12.1}}{3.3} & \evse{\textbf{$-$8.6}}{2.4} & \evse{\textbf{$-$8.0}}{2.9} & \evse{\textbf{$-$9.8}}{2.8} & \evse{\textbf{$-$8.6}}{2.6} & \evse{\textbf{$-$10.3}}{2.4} & \evse{\textbf{$-$14.9}}{2.9} \\
Camera shift & \evse{$-$0.6}{0.6} & \evse{\textbf{$-$19.0}}{2.9} & \evse{$-$0.6}{1.5} & \evse{\textbf{$-$7.5}}{2.1} & \evse{\textbf{$-$16.7}}{3.4} & \evse{+0.6}{1.3} & \evse{$-$2.3}{3.2} & \evse{\textbf{$-$5.7}}{1.9} & \evse{$-$1.7}{1.0} & \evse{+1.7}{2.2} \\
Sensor-noise onset & \evse{$-$3.9}{1.8} & \evse{$-$4.5}{1.9} & \evse{\textbf{$-$8.9}}{2.4} & \evse{\textbf{$-$16.8}}{3.1} & \evse{\textbf{$-$15.6}}{3.1} & \evse{\textbf{$-$7.3}}{2.6} & \evse{\textbf{$-$14.0}}{3.2} & \evse{\textbf{$-$10.6}}{2.5} & \evse{\textbf{$-$6.7}}{1.9} & \evse{\textbf{$-$11.7}}{3.1} \\
Illumination switch & \evse{$-$1.1}{0.8} & \evse{$-$0.6}{1.0} & \evse{$-$2.8}{1.8} & \evse{\textbf{$-$17.4}}{3.0} & \evse{\textbf{$-$13.5}}{2.9} & \evse{$-$2.8}{1.5} & \evse{+0.0}{2.6} & \evse{\textbf{$-$6.7}}{2.0} & \evse{$-$1.7}{1.2} & \evse{$-$2.2}{2.5} \\
Visual-theme switch & \evse{\textbf{$-$11.7}}{2.5} & \evse{+0.0}{1.1} & \evse{$-$5.0}{1.9} & \evse{\textbf{$-$7.2}}{2.1} & \evse{\textbf{$-$8.3}}{2.4} & \evse{\textbf{$-$12.8}}{2.9} & \evse{$-$4.4}{2.5} & \evse{\textbf{$-$29.4}}{3.6} & \evse{\textbf{$-$7.8}}{2.2} & \evse{+4.4}{2.2} \\
Distractor burst & \evse{$-$1.1}{1.1} & \evse{$-$2.8}{1.2} & \evse{$-$3.4}{1.6} & \evse{$-$4.5}{1.9} & \evse{$-$0.6}{2.0} & \evse{$-$1.7}{1.3} & \evse{+0.6}{2.1} & \evse{$-$3.4}{1.6} & \evse{$-$0.6}{1.5} & \evse{$-$0.6}{1.9} \\
\midrule
\multicolumn{11}{@{}l}{\tableheader Conditional regression (\%)} \\
\cmidrule{1-11}
Target relocation & \evse{10.4}{2.3} & \evse{21.6}{3.7} & \evse{20.0}{3.4} & \evse{20.5}{3.6} & \evse{19.8}{3.8} & \evse{19.9}{3.3} & \evse{28.6}{5.8} & \evse{17.2}{3.5} & \evse{16.6}{3.0} & \evse{21.0}{4.4} \\
Receptacle relocation & \evse{6.0}{2.0} & \evse{8.5}{2.7} & \evse{3.7}{1.8} & \evse{7.4}{3.2} & \evse{13.3}{4.7} & \evse{9.9}{3.5} & \evse{33.3}{6.9} & \evse{5.7}{2.9} & \evse{6.8}{2.5} & \evse{25.5}{6.0} \\
Obstacle insertion & \evse{4.8}{1.7} & \evse{24.7}{3.7} & \evse{6.9}{2.1} & \evse{20.0}{4.0} & \evse{17.5}{3.7} & \evse{13.3}{2.8} & \evse{29.6}{5.7} & \evse{16.2}{3.3} & \evse{13.0}{2.7} & \evse{23.9}{4.2} \\
Camera shift & \evse{0.6}{0.6} & \evse{22.9}{3.4} & \evse{3.1}{1.5} & \evse{10.1}{2.4} & \evse{31.5}{4.6} & \evse{1.4}{1.0} & \evse{27.0}{5.8} & \evse{9.3}{2.6} & \evse{1.9}{1.1} & \evse{7.1}{2.5} \\
Sensor-noise onset & \evse{5.6}{1.7} & \evse{7.3}{2.2} & \evse{14.0}{2.9} & \evse{24.1}{3.5} & \evse{30.3}{4.8} & \evse{11.9}{2.7} & \evse{42.3}{5.7} & \evse{16.3}{3.4} & \evse{7.4}{2.1} & \evse{27.1}{4.7} \\
Illumination switch & \evse{1.2}{0.9} & \evse{1.3}{1.0} & \evse{6.1}{2.0} & \evse{24.8}{3.5} & \evse{25.7}{4.1} & \evse{4.3}{1.7} & \evse{14.3}{4.2} & \evse{10.9}{3.1} & \evse{2.6}{1.3} & \evse{11.8}{3.5} \\
Visual-theme switch & \evse{12.7}{2.7} & \evse{1.4}{1.0} & \evse{8.5}{2.4} & \evse{11.4}{2.9} & \evse{16.4}{3.2} & \evse{17.9}{3.3} & \evse{23.2}{5.2} & \evse{41.2}{4.5} & \evse{10.5}{2.4} & \evse{4.3}{2.1} \\
Distractor burst & \evse{1.9}{1.1} & \evse{3.5}{1.5} & \evse{5.3}{1.9} & \evse{7.5}{2.3} & \evse{7.2}{2.6} & \evse{2.9}{1.5} & \evse{9.0}{3.6} & \evse{6.5}{2.4} & \evse{2.9}{1.4} & \evse{7.7}{3.0} \\
 
\bottomrule
\end{tabular*}
\end{table*}

\subsection{Effects by chain position}
\label{sec:supp_position}
\begin{figure}[!tp]
\centering
\includegraphics[width=\linewidth]{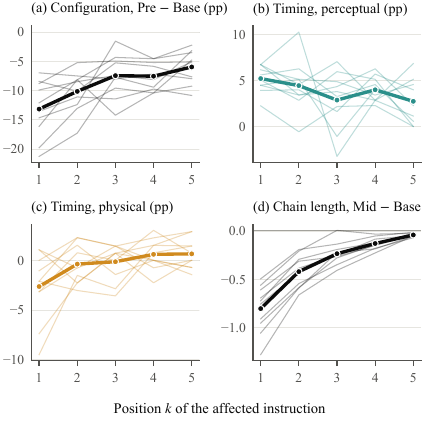}
\vspace{-3mm}
\caption{Effects by the position $k$ of the affected instruction, on the held-out core and embodiment cases. Thin lines show one policy each and the thick line the mean over the ten policies. Panel (a) gives the cost of the changed world, $\dconf$; panels (b) and (c) the timing effect $\dtim$ for perceptual and physical changes; and panel (d) the difference in chain length, \Mid$-$\Base, in instructions.}
\label{fig:supp_position}
\end{figure}

\Cref{fig:supp_position,tab:supp_position} break the \Track effects down by the position of the change (exploratory). A change at the first instruction has the largest configuration effect and costs the most chain length, and the perceptual timing benefit is largest there; the physical timing effect is slightly negative at the first instruction and near zero later. Part of the chain-length profile is mechanical, since a change at instruction $k$ can cost at most $6{-}k$ instructions. All effects here are over all assigned cases, with unrealized cases at zero difference, so the timing effects are smaller than the realized-case comparisons between channels and also reflect each position's coverage. Positions differ in task mix and \Base success, so the profile is descriptive.

\begin{table*}[!tp]
\caption{Effects by the position $k$ of the affected instruction, on the held-out cases, as the mean over the ten policies with its 95\% interval, resampling chains jointly across policies. Effects are in pp and chain length in instructions. A change at the first instruction has the largest cost of the changed world and costs the most chain length.}
\label{tab:supp_position}
\centering
\footnotesize
\setlength{\tabcolsep}{4pt}
\begin{adjustbox}{max width=\textwidth}
\begin{tabular}{@{}l ccccc@{}}
\toprule
\tableheader Measure & $k{=}1$ & $k{=}2$ & $k{=}3$ & $k{=}4$ & $k{=}5$ \\
\midrule
Pre $-$ Base (pp) & \tci{$-$13.2}{$-$15.6}{$-$10.9} & \tci{$-$10.1}{$-$12.4}{$-$8.0} & \tci{$-$7.4}{$-$9.1}{$-$5.8} & \tci{$-$7.5}{$-$9.2}{$-$6.0} & \tci{$-$6.0}{$-$7.5}{$-$4.5} \\
Mid $-$ Pre, perceptual (pp) & \tci{+5.2}{+3.5}{+6.9} & \tci{+4.5}{+2.4}{+6.7} & \tci{+2.9}{+1.2}{+4.5} & \tci{+4.0}{+2.5}{+5.6} & \tci{+2.7}{+1.2}{+4.3} \\
Mid $-$ Pre, physical (pp) & \tci{$-$2.6}{$-$5.0}{$-$0.3} & \tci{$-$0.4}{$-$2.0}{+1.2} & \tci{$-$0.1}{$-$2.1}{+1.9} & \tci{+0.6}{$-$1.1}{+2.2} & \tci{+0.6}{$-$0.6}{+1.9} \\
$\Delta$ chain length & \tci{$-$0.80}{$-$0.92}{$-$0.69} & \tci{$-$0.42}{$-$0.50}{$-$0.34} & \tci{$-$0.23}{$-$0.28}{$-$0.19} & \tci{$-$0.13}{$-$0.16}{$-$0.10} & \tci{$-$0.04}{$-$0.05}{$-$0.03} \\
 
\bottomrule
\end{tabular}\end{adjustbox}
\end{table*}

\section{Timing Mechanism}
\label{sec:supp_decomp}

\subsection{Diagnostic arms}
\label{sec:supp_arms}
\begin{table*}[!tp]
\caption{How the perceptual timing effect splits (H3), on the realized held-out perceptual core cases with \Pre, \Mid and rewind. Values are in pp with 95\% intervals except where noted. Of the 887 assigned perceptual core cases per policy, $n$ counts those realized with \Pre, \Mid and rewind; rewind is not applied when the robot touches a movable object at the onset, so $\dtim$ differs slightly from the realized-case value of \cref{tab:main}. Motion is \Mid$-$rewind and is significant after Holm correction for every policy. For policies that execute a queue it splits into Queue, \Mid$-$flush, and State, flush$-$rewind; -- marks a policy that executes one action per query ($Q{=}1$) and so has no queue to flush. Residual is rewind$-$\Pre with the 90\% interval of the equivalence test, and Matched is the same residual with \Pre held to rewind's remaining step budget (exploratory), also with its 90\% interval. The H3 column reads \emph{yes} where both parts of the criterion hold, a Holm-significant Motion and a residual whose 90\% interval lies inside $\pm$3 points without correction over policies, and \emph{gain only} where only Motion holds. Holm-correcting the equivalence tests as well leaves only X-VLA (\cref{sec:supp_prereg}).}
\label{tab:decomp}
\centering
\footnotesize
\setlength{\tabcolsep}{2pt}
\renewcommand{\arraystretch}{1.1}
\begin{adjustbox}{max width=\textwidth}
\begin{tabular}{@{}lrccccccl@{}}
\toprule
\tableheader Policy & \tableheader $n$ & \tableheader $\dtim$ & \tableheader Motion & \tableheader Queue & \tableheader State & \tableheader Residual & \tableheader Matched & \tableheader H3 \\
\midrule
FLOWER & 654 & \tci{+4.1}{+2.3}{+6.1} & \tci{+5.7}{+3.5}{+7.9} & \tci{+1.5}{+0.3}{+2.8} & \tci{+4.1}{+1.8}{+6.5} & \tci{$-$1.5}{$-$3.5}{+0.3} & \tci{$-$1.1}{$-$2.9}{+0.8} & gain only \\
X-VLA & 597 & \tci{+6.2}{+3.7}{+8.7} & \tci{+6.2}{+3.7}{+8.8} & \tci{+4.7}{+2.4}{+7.1} & \tci{+1.5}{$-$0.5}{+3.6} & \tci{+0.0}{$-$1.7}{+1.6} & \tci{+0.0}{$-$1.7}{+1.7} & yes \\
$\pi_{0.5}$ & 566 & \tci{+0.2}{$-$2.1}{+2.5} & \tci{+3.0}{+0.2}{+5.9} & \tci{$-$0.5}{$-$3.0}{+2.2} & \tci{+3.5}{+0.7}{+6.3} & \tci{$-$2.8}{$-$5.2}{$-$0.5} & \tci{$-$1.8}{$-$4.1}{+0.6} & gain only \\
UniVLA & 564 & \tci{+4.4}{+1.1}{+7.8} & \tci{+9.2}{+5.0}{+13.4} & \tci{+2.1}{$-$1.0}{+5.2} & \tci{+7.1}{+3.2}{+11.0} & \tci{$-$4.8}{$-$7.7}{$-$1.9} & \tci{$-$4.3}{$-$7.3}{$-$1.3} & gain only \\
OpenVLA-OFT & 476 & \tci{+3.8}{+0.0}{+7.9} & \tci{+6.5}{+2.5}{+10.6} & -- & -- & \tci{$-$2.7}{$-$6.1}{+0.7} & \tci{$-$1.3}{$-$4.7}{+2.1} & gain only \\
MoDE & 593 & \tci{+3.9}{+1.5}{+6.3} & \tci{+3.2}{+0.8}{+5.5} & \tci{+0.2}{$-$1.6}{+1.9} & \tci{+3.0}{+0.9}{+5.2} & \tci{+0.7}{$-$1.2}{+2.4} & \tci{+1.4}{$-$0.5}{+3.1} & yes \\
3D Diffuser Actor & 372 & \tci{+8.1}{+2.7}{+13.9} & \tci{+20.4}{+15.0}{+26.0} & \tci{+5.1}{$-$0.3}{+10.3} & \tci{+15.3}{+10.2}{+20.3} & \tci{$-$12.4}{$-$16.8}{$-$7.8} & \tci{$-$1.3}{$-$5.5}{+2.9} & gain only \\
GR-1 & 498 & \tci{+5.4}{+2.2}{+8.6} & \tci{+6.6}{+3.5}{+9.7} & -- & -- & \tci{$-$1.2}{$-$3.6}{+1.2} & \tci{+0.6}{$-$1.8}{+3.0} & gain only \\
Seer-Large & 627 & \tci{+3.2}{+1.4}{+5.0} & \tci{+4.0}{+2.1}{+5.9} & -- & -- & \tci{$-$0.8}{$-$2.2}{+0.6} & \tci{$-$0.5}{$-$2.0}{+1.1} & yes \\
RoboFlamingo & 467 & \tci{+5.6}{+1.8}{+9.4} & \tci{+8.1}{+4.4}{+11.8} & -- & -- & \tci{$-$2.6}{$-$5.6}{+0.5} & \tci{$-$1.5}{$-$4.7}{+1.7} & gain only \\
 
\bottomrule
\end{tabular}\end{adjustbox}
\end{table*}

\Mid-flush is \Mid with the policy's action queue emptied at the onset, so the first action after the change comes from a fresh query on the changed observation. \Mid-rewind applies the change at the onset and then restores the robot to its joint configuration at the start of instruction $k$ (zero velocity) and resets the policy, so the policy restarts the instruction in the changed world, as in \Pre, but after the world has evolved for the steps before the onset. The residual rewind$-$\Pre should therefore be near zero; a large residual indicates that something other than the motion already under way differs, such as a held block that cannot be rewound. One such difference is by design: rewind is a \Mid variant and keeps \Mid's remaining step budget, $360-(\text{onset}-\text{instruction start})$, while \Pre has the full 360, a median of \maxa{28} (OpenVLA-OFT) to \maxa{66} (3D Diffuser Actor) steps more. Flush and rewind differ only in the robot (joint configuration, velocity and gripper) and in the policy's internal state other than its queue, such as a proprioceptive input or a history of frames, which rewind returns to the start of the instruction; both keep the world as it evolved before the onset and \Mid's step budget. State is thus an operational contrast between two arms, not a measure of physical progress alone. The budget-matched residual of \cref{tab:decomp} counts a \Pre success as a failure when it took more steps than rewind had (exploratory). Rewind is not applied if the robot touches a movable object at the onset, so cases with a held block drop out. GR-1, Seer-Large, OpenVLA-OFT and RoboFlamingo execute one action per query ($Q{=}1$): they hold no queue at the onset, so flush equals \Mid and Motion is not split. \Cref{tab:supp_decomp_lite} repeats the split on Lite for both channels and three policies. For physical changes no Queue or State interval excludes zero; MoDE's residual interval does.

\begin{table*}[!tp]
\caption{The same split on Lite for both channels, for FLOWER, MoDE and X-VLA, on realized cases with all four arms. Queue is \Mid$-$flush, State flush$-$rewind and Residual rewind$-$\Pre, in pp with 95\% intervals. For physical changes no Queue or State term is separated from zero.}
\label{tab:supp_decomp_lite}
\centering
\footnotesize
\setlength{\tabcolsep}{4pt}
\begin{adjustbox}{max width=\textwidth}
\begin{tabular}{@{}llrcccc@{}}
\toprule
\tableheader Policy & \tableheader Channel & \tableheader $n$ & \tableheader $\dtim$ & \tableheader Queue & \tableheader State & \tableheader Residual \\
\midrule
FLOWER & perceptual & 224 & \tci{+5.8}{+2.2}{+9.7} & \tci{+0.4}{$-$1.8}{+2.8} & \tci{+8.9}{+5.4}{+12.7} & \tci{$-$3.6}{$-$7.6}{+0.4} \\
FLOWER & physical & 79 & \tci{$-$2.5}{$-$10.3}{+5.1} & \tci{$-$1.3}{$-$9.0}{+6.4} & \tci{$-$1.3}{$-$7.1}{+3.8} & \tci{+0.0}{$-$6.2}{+6.3} \\
MoDE & perceptual & 210 & \tci{+6.7}{+2.3}{+11.4} & \tci{+3.8}{+1.0}{+6.8} & \tci{+2.9}{$-$1.4}{+7.3} & \tci{+0.0}{$-$4.3}{+4.2} \\
MoDE & physical & 67 & \tci{$-$3.0}{$-$12.1}{+6.1} & \tci{+3.0}{$-$4.6}{+11.9} & \tci{+7.5}{$-$2.9}{+17.9} & \tci{$-$13.4}{$-$23.5}{$-$4.3} \\
X-VLA & perceptual & 214 & \tci{+7.9}{+3.8}{+12.2} & \tci{+6.5}{+1.9}{+11.2} & \tci{+3.3}{$-$0.5}{+7.2} & \tci{$-$1.9}{$-$5.7}{+1.9} \\
X-VLA & physical & 69 & \tci{$-$5.8}{$-$14.5}{+3.0} & \tci{$-$5.8}{$-$16.4}{+4.4} & \tci{+5.8}{$-$5.8}{+17.1} & \tci{$-$5.8}{$-$16.2}{+4.4} \\
 
\bottomrule
\end{tabular}\end{adjustbox}
\end{table*}

\subsection{Strong sensor noise}
\label{sec:supp_strongnoise}
We repeated the split of \cref{tab:supp_decomp_lite} at 2$\times$ and 4$\times$ sensor noise on the Lite cases of the strength sweep (exploratory). It uses the 41 to 44 cases per policy and level realized with all four arms, fewer than the realized cases of \cref{tab:supp_magnitude}, so its \Mid$-$\Pre differs from the values there. Where the benefit is large, most of it is State: at 4$\times$, FLOWER's \Mid$-$\Pre of \lite{+37.2} \lite{[+20.9, +53.5]} splits into State \lite{+30.2} \lite{[+14.3, +45.2]} and Queue \lite{+11.6} \lite{[+2.4, +21.4]}, and MoDE's \lite{+36.6} \lite{[+20.9, +52.5]} into State \lite{+41.5} \lite{[+25.0, +58.5]} and Queue \lite{+7.3} \lite{[0.0, +15.0]}; at 2$\times$, X-VLA's \lite{+13.6} \lite{[+2.3, +25.6]} is State \lite{+11.4} \lite{[+2.3, +20.9]} with Queue \lite{0.0} \lite{[$-$11.1, +11.4]}. The other three pairs have smaller benefits with wide intervals (FLOWER at 2$\times$ \lite{+16.3}, State \lite{+7.0}, Queue \lite{+4.7}; MoDE at 2$\times$ \lite{+12.2}; X-VLA at 4$\times$ \lite{+4.5}). Under milder corruption queued actions carry part of the benefit, as for X-VLA at the default strength (\cref{sec:mechanism}). These sweeps cover three policies on Lite only; Max-A does not repeat them.

\subsection{Changes that are undone}
\label{sec:supp_transient}
Each Lite block relocation has a transient counterpart in which the block returns to its original pose after a delay. If the robot is touching the block at that step, the return is skipped and not retried, and the case is kept with the block where it was moved, so every mean below includes blocked returns. \Cref{tab:supp_transient} gives the counterparts with a 10-step delay. For FLOWER a briefly moved block costs \lite{0.0} \lite{[$-$6.2, +6.2]} points, and \lite{+2.2 / $-$2.2 / 0.0 / 0.0} across delays of 2, 5, 10 and 20 steps. For X-VLA most of the cost comes from the change \emph{staying}; the undone change is estimated at \lite{$-$11.6} \lite{[$-$23.8, 0.0]} at a 10-step delay, but at \lite{$-$2.4 / +4.8 / $-$11.6 / $-$4.8} across the four delays, so the 10-step value is not a trend. Each delay has 42 to 48 cases per policy; a return is blocked in at most 3 of them up to 10 steps and in 7 to 10 at 20 steps. MoDE is the only one of the three whose undone-change cost grows with exposure (\lite{$-$4.4 / $-$4.4 / $-$8.9 / $-$15.6}), carried by its 13 receptacle cases (\lite{$-$38.5} at 20 steps). Undoing the change recovers \lite{+18.6} points for X-VLA \lite{[+2.4, +34.9]}, the only interval that excludes zero; for FLOWER (\lite{+4.2}) and MoDE (\lite{+6.7}) it includes zero. The number of cases $n$ is small, so these results are exploratory.

\begin{table*}[!tp]
\caption{Relocations that are undone after 10 steps, on Lite (exploratory). $n$ counts the cases, with blocked returns in parentheses, which stay in every mean, and values are in pp with 95\% intervals. The last column gives what undoing the change recovers in \Mid, the transient minus the persistent effect.}
\label{tab:supp_transient}
\centering
\footnotesize
\setlength{\tabcolsep}{4pt}
\begin{adjustbox}{max width=\textwidth}
\begin{tabular}{@{}llccc@{}}
\toprule
\tableheader Policy & \tableheader $n$ & \tableheader Persistent \Mid$-$\Base & \tableheader Transient \Mid$-$\Base & \tableheader Trans.$-$pers.\ (\Mid) \\
\midrule
FLOWER & 48 (2) & \tci{$-$4.2}{$-$12.2}{+4.2} & \tci{+0.0}{$-$6.2}{+6.2} & \tci{+4.2}{$-$4.3}{+12.2} \\
MoDE & 45 (1) & \tci{$-$15.6}{$-$31.1}{+0.0} & \tci{$-$8.9}{$-$22.7}{+4.5} & \tci{+6.7}{$-$10.6}{+22.7} \\
X-VLA & 43 (3) & \tci{$-$30.2}{$-$45.2}{$-$14.3} & \tci{$-$11.6}{$-$23.8}{+0.0} & \tci{+18.6}{+2.4}{+34.9} \\
 
\bottomrule
\end{tabular}\end{adjustbox}
\end{table*}

\section{Strength and Onset Phase}
\label{sec:supp_dose}

\subsection{Strength of the change}
\label{sec:supp_ladders}
\begin{figure}[!htbp]
\centering
\includegraphics[width=\linewidth]{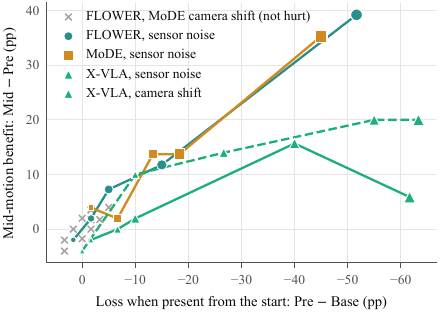}
\vspace{-5mm}
\caption{The benefit of a mid-motion arrival against the damage a change does, at five strengths (exploratory). Larger markers are stronger, and each pair of policy and change has 60 assigned cases per level, the Lite cases of that change. \Pre$-$\Base uses all assigned cases and \Mid$-$\Pre realized cases, and both contain \Pre, so the relation is descriptive. For FLOWER and MoDE the benefit grows with the damage, and a change that does not hurt a policy (grey) gives no benefit.}
\label{fig:dose}
\vspace{-3mm}
\end{figure}
For the camera shift and sensor noise the levels are 0.25, 0.5, 1, 2 and 4 times variant A: the yaw and pitch of the shift, and both the $\sigma$ and the occluded area of the noise (from 0.0125 and 2.5\% to 0.2 and 40\%). Target relocation uses moves of 2, 4, 6, 9 and 12\,cm. \Pre and \Mid run at every level on the Lite cases of each change (60 assigned cases per level; the realized count is lower and differs by policy). \Cref{tab:supp_magnitude} lists the values plotted in \cref{fig:dose} and those for target relocation, which the figure leaves out. Target relocation differs: a mid-motion arrival adds cost for small moves on X-VLA and large moves on FLOWER, and only 29 to 41 of the 60 cases are realized at 9 and 12\,cm.

\begin{table}[!htbp]
\caption{The values plotted in \cref{fig:dose} and those for target relocation, on Lite: \Pre$-$\Base on all assigned cases and \Mid$-$\Pre on realized cases, in pp. The levels are 0.25, 0.5, 1, 2 and 4 times variant A for the camera shift and sensor noise, and 2, 4, 6, 9 and 12\,cm for target relocation.}
\label{tab:supp_magnitude}
\centering
\footnotesize
\setlength{\tabcolsep}{1pt}
\begin{adjustbox}{max width=\columnwidth}
\begin{tabular}{@{}lll@{}}
\toprule
\tableheader Policy, change &\tableheader \Pre$-$\Base & \tableheader \Mid$-$\Pre \\
\midrule
X-VLA, camera & \lite{0 / $-$10 / $-$27 / $-$55 / $-$63} & \lite{$-$4 / +10 / +14 / +20 / +20} \\
FLOWER, camera & \lite{$-$2 / 0 / $-$2 / $-$2 / $-$3} & \lite{+4 / $-$2 / +2 / 0 / +2} \\
MoDE, camera & \lite{+3 / 0 / +3 / +2 / $-$5} & \lite{$-$4 / +2 / $-$2 / 0 / +4} \\
FLOWER, noise & \lite{+2 / $-$2 / $-$5 / $-$15 / $-$52} & \lite{$-$2 / +2 / +7 / +12 / +39} \\
MoDE, noise & \lite{$-$2 / $-$7 / $-$13 / $-$18 / $-$45} & \lite{+4 / +2 / +14 / +14 / +35} \\
X-VLA, noise & \lite{$-$2 / $-$7 / $-$10 / $-$40 / $-$62} & \lite{$-$2 / 0 / +2 / +16 / +6} \\
X-VLA, target & \lite{+2 / $-$3 / $-$17 / $-$15 / $-$3} & \lite{$-$8 / $-$9 / 0 / $-$3 / $-$7} \\
FLOWER, target & \lite{$-$3 / $-$2 / $-$5 / $-$13 / $-$5} & \lite{+2 / +2 / 0 / $-$5 / $-$10} \\
MoDE, target & \lite{+2 / $-$2 / 0 / $-$13 / $-$10} & \lite{$-$2 / +2 / $-$4 / +2 / +9} \\
\bottomrule
\end{tabular}\end{adjustbox}
\end{table}

\begin{figure}[!htbp]
\centering
\includegraphics[width=\linewidth]{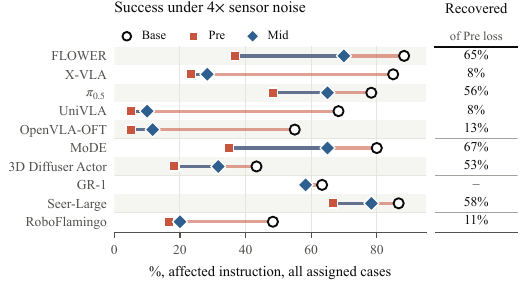}
\vspace{-5mm}
\caption{Success on the affected instruction under sensor noise at four times the default strength, for all ten policies (Lite, exploratory). The column on the right gives the share of the loss in \Pre that \Mid recovers on realized cases; GR-1 loses too little for a share.}
\label{fig:noise4x}
\vspace{-3mm}
\end{figure}

\paragraph{4$\times$ sensor noise on every policy.}
\label{sec:supp_noise4x}
To test whether the largest benefit generalizes beyond the three policies of the sweep, the other policies ran the strongest sensor noise (4$\times$ variant A, the same 60 cases; exploratory). \Cref{fig:noise4x} shows the success rates, and \cref{tab:supp_noise4x} reports both differences on realized cases, so the share of the loss that \Mid recovers has one denominator. The point estimates of five policies (FLOWER, MoDE, $\pi_{0.5}$, Seer-Large, 3D Diffuser Actor) recover \lite{53--67\%} of the loss. Four (X-VLA, UniVLA, OpenVLA-OFT, RoboFlamingo) lose \lite{58--81} points and recover \lite{8--13\%}. GR-1 is barely affected (\lite{$-$9}), too little for a recovery share. The noise corrupts only the static camera: UniVLA and OpenVLA-OFT see nothing else and fall to \lite{5\%} under \Pre, while X-VLA and RoboFlamingo also have the gripper camera yet recover little. A larger loss thus does not by itself imply a larger benefit across policies. Within FLOWER and MoDE it does (\cref{fig:dose}). X-VLA's noise benefit peaks at 2$\times$ (\lite{+16}) and falls to \lite{+6} at 4$\times$.

\begin{table*}[!tp]
\caption{Sensor noise at four times the default strength on every policy (Lite cases, exploratory). Real.\ gives the realized and assigned cases. Success (\%) is over all assigned cases, and \Pre$-$\Base and \Mid$-$\Pre (pp) over realized cases. Rec.\ is the share of the loss that \Mid recovers (\%), shown where \Pre loses at least 10 points and -- otherwise. Every value has its 95\% bootstrap interval over chains. The point estimates of five policies recover more than half of the loss.}
\label{tab:supp_noise4x}
\centering
\footnotesize
\setlength{\tabcolsep}{1pt}
\begin{tabular*}{\textwidth}{@{\extracolsep{\fill}}l r ccc cc c@{}}
\toprule
\tableheader Policy & Real. & \Base & \Pre & \Mid & \Pre$-$\Base & \Mid$-$\Pre & Rec. \\
\midrule
FLOWER & 51/60 & \tcis{88}{80}{95} & \tcis{37}{25}{49} & \tcis{70}{58}{81} & \tci{$-$60.8}{$-$74.5}{$-$47.2} & \tci{+39.2}{+25.0}{+54.0} & \tcis{65}{46}{82} \\
X-VLA & 51/60 & \tcis{85}{75}{93} & \tcis{23}{13}{34} & \tcis{28}{17}{41} & \tci{$-$72.5}{$-$84.6}{$-$58.8} & \tci{+5.9}{$-$4.0}{+15.7} & \tcis{8}{$-$6}{22} \\
$\pi_{0.5}$ & 46/60 & \tcis{78}{67}{88} & \tcis{48}{36}{61} & \tcis{65}{53}{77} & \tci{$-$39.1}{$-$53.3}{$-$25.0} & \tci{+21.7}{+8.7}{+35.6} & \tcis{56}{27}{80} \\
UniVLA & 47/60 & \tcis{68}{56}{80} & \tcis{5}{0}{11} & \tcis{10}{3}{18} & \tci{$-$80.9}{$-$91.5}{$-$69.6} & \tci{+6.4}{+0.0}{+14.3} & \tcis{8}{0}{18} \\
OpenVLA-OFT & 38/60 & \tcis{55}{43}{67} & \tcis{5}{0}{12} & \tcis{12}{5}{20} & \tci{$-$78.9}{$-$92.1}{$-$65.8} & \tci{+10.5}{+0.0}{+23.7} & \tcis{13}{0}{29} \\
MoDE & 51/60 & \tcis{80}{69}{90} & \tcis{35}{23}{47} & \tcis{65}{53}{77} & \tci{$-$52.9}{$-$67.9}{$-$37.3} & \tci{+35.3}{+20.0}{+50.0} & \tcis{67}{46}{84} \\
3D Diffuser Actor & 34/60 & \tcis{43}{31}{56} & \tcis{18}{10}{29} & \tcis{32}{20}{44} & \tci{$-$44.1}{$-$63.6}{$-$21.2} & \tci{+23.5}{+0.0}{+45.7} & \tcis{53}{0}{92} \\
GR-1 & 35/60 & \tcis{63}{51}{75} & \tcis{58}{46}{70} & \tcis{58}{46}{70} & \tci{$-$8.6}{$-$18.9}{+0.0} & \tci{+0.0}{$-$17.1}{+16.2} & -- \\
Seer-Large & 51/60 & \tcis{87}{78}{95} & \tcis{67}{55}{78} & \tcis{78}{68}{88} & \tci{$-$23.5}{$-$36.0}{$-$11.5} & \tci{+13.7}{+2.0}{+25.5} & \tcis{58}{11}{100} \\
RoboFlamingo & 33/60 & \tcis{48}{36}{61} & \tcis{17}{8}{27} & \tcis{20}{10}{31} & \tci{$-$57.6}{$-$75.8}{$-$39.3} & \tci{+6.1}{$-$9.1}{+21.2} & \tcis{11}{$-$21}{37} \\
 
\bottomrule
\end{tabular*}
\end{table*}

\subsection{Onset phase}
\label{sec:supp_phase}
\begin{figure}[!htbp]
\centering
\includegraphics[width=\linewidth]{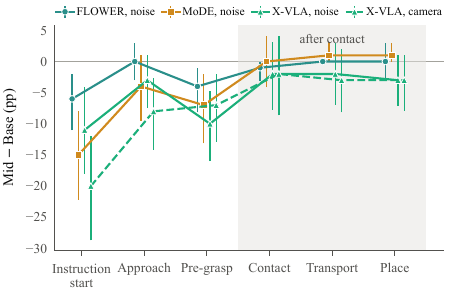}
\vspace{-5mm}
\caption{The total effect \Mid$-$\Base when sensor noise begins at each of six onsets, for three policies (exploratory). Each onset has 100 cases, all assigned, with 95\% intervals that resample whole chains. Three curves use the camera-shift cases of the phase track with sensor noise in place of the shift, and one shows X-VLA under the camera shift itself. Of the 100 cases per onset, 49 to 96 are realized, fewest at transport, and the others count at their \Base outcome. Onsets differ in task mix, and a later onset leaves fewer steps under the changed view, so the profile is descriptive. Most of the loss falls before contact.}
\label{fig:phase}
\vspace{-3mm}
\end{figure}
We apply variant-A sensor noise at six onsets, 100 cases per onset, on the 600 camera-shift cases of the phase track, with the noise in place of the camera shift, which barely affects FLOWER and MoDE (\cref{tab:supp_magnitude}), and only \Mid run (\cref{fig:phase}). The three policies lose \lite{6} to \lite{15} points when it begins at the start of the instruction, where \Mid starts from the same state as \Pre (\cref{tab:supp_onsets}), and \lite{4} to \lite{10} just before the grasp, but at most \lite{3} after contact. Restricted to block tasks, the profile is the same (FLOWER \lite{$-$6.8 / +3.1 / $-$6.7 / $-$2.4 / 0.0}; MoDE \lite{$-$22.7 / 0.0 / $-$11.1 / $-$2.4 / +1.0}; X-VLA \lite{$-$25.0 / $-$3.1 / $-$8.9 / $-$4.8 / $-$2.0} from the start of the instruction to transport; the place onset has only place and stack tasks, which are not block tasks, so it has no value here). Onsets differ in task mix and chain set, only \Mid is run, so these are total effects, and a later onset leaves fewer steps under the corrupted view, so differences between phases are descriptive. Sweeps for target relocation, receptacle relocation and arm push are flat or uniformly negative (target relocation $-$6 to $-$9 at every phase for FLOWER). The sweep covers three policies; Max-A fixes the onset of each case and does not repeat it.

\section{Consequences Down the Chain}
\label{sec:supp_chain}

Among realized \Mid cases in which the affected instruction was completed in both \Base and \Mid, \cref{tab:supp_aftershock} compares success on instruction $k{+}1$ on the held-out cases (H6). Intervals resample whole chains; $p$ is the one-sided bootstrap $p$ for \Mid$<$\Base, with Holm correction over the policies. This is a conditional paired analysis, not a population causal effect. The cases are core and embodiment changes at positions 1 to 4; the changes to the robot, which act once rather than persist, are \maxa{679} of the \maxa{6,518} cases over the ten policies. On core changes alone (exploratory) the cost is \maxa{$-$18.4} to \maxa{$-$5.3} points and Holm-significant for every policy. \Cref{fig:aftershock} in the main text shows it per policy, with the odds ratios.

\paragraph{Changed world or timing.} Every core change stays in place to the end of the chain (the embodiment changes, which act once, are included too), so the difference above could be the cost of the changed world on instruction $k{+}1$ or a trace of having met the change mid-motion. \Pre helps to tell them apart (exploratory, not among the hypotheses fixed in advance). We keep the cases in which the affected instruction was completed in \Base, \Pre and \Mid (\maxa{266} to \maxa{859} per policy) and compare success on instruction $k{+}1$ across the three arms, as an equal-weight mean over the ten policies with chains resampled jointly. It is \maxa{85.5\%} in \Base, \maxa{78.0\%} in \Pre and \maxa{77.5\%} in \Mid, so \Pre$-$\Base is \maxa{$-$7.4} \maxa{[$-$8.8, $-$6.1]} points and \Mid$-$\Pre is \maxa{$-$0.5} \maxa{[$-$1.6, +0.5]}. For perceptual changes \Pre$-$\Base is \maxa{$-$9.7} \maxa{[$-$11.5, $-$7.9]} and \Mid$-$\Pre \maxa{+0.3} \maxa{[$-$1.0, +1.7]}; for physical changes they are \maxa{$-$3.0} \maxa{[$-$5.0, $-$1.0]} and \maxa{$-$2.2} \maxa{[$-$4.0, $-$0.3]}. The cost on the next instruction is thus about the same in \Pre and \Mid, which is consistent with a cost of the changed world that a policy pays again after coping with the change once, with a small additional cost of timing for physical changes. The three arms end instruction $k$ in different robot and scene states, which this comparison does not separate from the change that persists.

\paragraph{Hazard over the rest of the chain.} To ask whether the cost stops at the next instruction, we fit a discrete-time hazard model (exploratory) on the H6 cases of \cref{tab:supp_aftershock}, in which \Base and \Mid both completed instruction $k$, not on the three-arm cases above. Each instruction $j=k{+}1..5$ that the \Base or \Mid arm reaches is one observation, failing it is the outcome, and $\operatorname{logit} h = \alpha_{j-k} + \beta_\text{next}\,\Mid\,[j{=}k{+}1] + \beta_\text{later}\,\Mid\,[j{>}k{+}1]$, where \Mid is 1 for the \Mid arm and 0 for \Base and a bracket is 1 when its condition holds and 0 otherwise. Pooled, $\alpha$ is specific to the policy and the lag $j{-}k$, and the two $\beta$ are shared. The analysis plan named a chain random effect; we resample whole chains instead (1,000 draws), which gives intervals that are robust to the same within-chain dependence. Pooled, the odds of failing are 1.95 [1.77, 2.15] times higher on the next instruction and 1.67 [1.52, 1.85] times higher on each later instruction. Per policy, the next-instruction odds ratio is 1.32 to 3.13, every interval above 1, and the interval of the later odds ratio excludes 1 for every policy except 3D Diffuser Actor (1.04 [0.73, 1.47]) and RoboFlamingo (1.10 [0.85, 1.42]). This is consistent with a cost that persists down the chain, not only on the next instruction. The position profile is in \cref{sec:supp_position}.

\begin{table}[!htbp]
\caption{Success (\%) on the next instruction, $k{+}1$, among held-out cases in which both \Base and \Mid completed instruction $k$ (H6). $\Delta$ is \Mid$-$\Base in pp, with its 95\% interval and its one-sided bootstrap $p$ after Holm correction over the ten policies. Success is lower after the change for every policy.}
\label{tab:supp_aftershock}
\centering
\footnotesize
\setlength{\tabcolsep}{2.5pt}
\begin{tabular}{@{}lrrrrcr@{}}
\toprule
\tableheader Policy & \tableheader Cases & \tableheader \Mid & \tableheader \Base & \tableheader $\Delta$ & \tableheader 95\% CI & \tableheader $p_\text{Holm}$ \\
\midrule
FLOWER & 915 & 88.1 & 94.8 & $-$6.7 & \tciint{$-$8.7}{$-$4.8} & $<$0.001 \\
X-VLA & 760 & 83.2 & 92.0 & $-$8.8 & \tciint{$-$11.5}{$-$6.0} & $<$0.001 \\
$\pi_{0.5}$ & 743 & 85.3 & 91.7 & $-$6.3 & \tciint{$-$9.1}{$-$3.6} & $<$0.001 \\
UniVLA & 648 & 72.2 & 89.0 & $-$16.8 & \tciint{$-$20.1}{$-$13.4} & $<$0.001 \\
OpenVLA-OFT & 481 & 64.9 & 77.5 & $-$12.7 & \tciint{$-$18.0}{$-$7.5} & $<$0.001 \\
MoDE & 753 & 83.0 & 90.0 & $-$7.0 & \tciint{$-$9.7}{$-$4.4} & $<$0.001 \\
3D Diffuser Actor & 336 & 54.2 & 61.0 & $-$6.8 & \tciint{$-$13.3}{+0.0} & 0.026 \\
GR-1 & 565 & 74.5 & 85.3 & $-$10.8 & \tciint{$-$14.1}{$-$7.4} & $<$0.001 \\
Seer-Large & 816 & 88.6 & 93.6 & $-$5.0 & \tciint{$-$7.2}{$-$2.9} & $<$0.001 \\
RoboFlamingo & 501 & 69.1 & 76.2 & $-$7.2 & \tciint{$-$11.7}{$-$2.8} & 0.001 \\
 
\bottomrule
\end{tabular}
\end{table}

\section{Behaviour and Queue-Level Responses}
\label{sec:supp_behaviour}

\subsection{Behaviour metrics}
\label{sec:supp_behav_metrics}
All metrics use privileged state after the onset, on realized \Mid arms, and are exploratory. \emph{Retargeting latency}: steps from the onset to the first of $m{=}5$ consecutive steps in which the tool centre moves more than 1\,mm and its displacement projects more onto the unit vector from the tool centre toward the object's new position than onto the one toward its old position; an arm that never meets this within the instruction has no latency. \emph{Ghost reach}: the tool centre comes within 3\,cm of the object's old position, defined when the old and new positions are at least 6\,cm apart. \emph{Stale commitment}: path length travelled toward the old pose before retargeting. \emph{Response class}, in order of precedence: wrong task (fails while CALVIN's task checker reports another task completed); frozen (tool path under 2\,cm); among successes, stale-then-recovered (ghost reach), late-recovered (latency over 30 steps) or adapted; every other failure is \emph{failed after moving}. \Cref{tab:supp_behaviour} reports target and receptacle relocation on the held-out cases for all ten policies. Every arm, failed or successful, meets the latency criterion within the instruction, and no arm is frozen; ghost reach is about as common among failures (\maxa{27} of \maxa{150}) as among successes (\maxa{61} of \maxa{463}), so most losses are not a policy reaching for the old pose. A latency of 0 means that the criterion holds from the first step after the onset, which can happen by geometry when the new pose lies ahead of the motion, so the latency describes the path and is not a measure of detection. Under receptacle relocation, which moves the stack base while the policy carries a block toward it, every \Mid arm also retargets, within a median of \maxa{0--8} steps (3D Diffuser Actor \maxa{17}). Here, unlike for the target, ghost reach is more common among failures (\maxa{18} of \maxa{122}) than among successes (\maxa{11} of \maxa{273}), and \maxa{41} of the \maxa{395} arms complete another task instead, most often for MoDE and OpenVLA-OFT (\maxa{8} each). Side-effect tasks, another CALVIN task completed during the affected instruction, occur in \maxa{2--10}\% of \Mid arms and \maxa{3--12}\% of \Pre arms on the core changes, with no consistent difference between the two. The other changes relocate no target or receptacle, so they have no old and new pose for these metrics. Contacts with objects other than the target are not analysed here.

\begin{table*}[!tp]
\caption{How \Mid responds when a block is relocated, on the realized held-out cases (exploratory). Target relocation moves the block to be grasped, and receptacle relocation moves the stack-base block as the policy brings a block to it; drawer and slider relocations, as targets or receptacles, have no single pose and are left out. Succ.\ is the success rate, Lat.\ the median steps to retarget, Ghost the share of cases that reach the old pose, and Stale the median path toward the old pose before retargeting (cm). The last four columns count the response classes: Adapt.\ for adapted, Recov.\ for stale-then-recovered and late-recovered together, Failed for failed after moving, and Other for a wrong task or a frozen arm; no arm was frozen.}
\label{tab:supp_behaviour}
\centering
\footnotesize
\setlength{\tabcolsep}{4pt}
\begin{adjustbox}{max width=\textwidth}
\begin{tabular}{@{}llrrrrr rrrr@{}}
\toprule
\tableheader Family & \tableheader Policy & $n$ & \tableheader Succ.\,\% &\tableheader Lat. & \tableheader Ghost \% & \tableheader Stale & \tableheader Adapt. & \tableheader Recov. & \tableheader Failed & \tableheader Other \\
\midrule
\multicolumn{11}{@{}l}{\textcolor{phys}{\textbf{Target relocation}}} \\
\cmidrule{1-11}
VLA & FLOWER & 77 & 94.8 & 3 & 15.6 & 1.8 & 62 & 11 & 4 & 0 \\
 & X-VLA & 67 & 77.6 & 3 & 17.9 & 1.6 & 43 & 9 & 11 & 4 \\
 & $\pi_{0.5}$ & 65 & 81.5 & 0 & 6.2 & 0.0 & 50 & 3 & 10 & 2 \\
 & UniVLA & 62 & 64.5 & 6 & 17.7 & 2.8 & 35 & 5 & 18 & 4 \\
 & OpenVLA-OFT & 56 & 67.9 & 0 & 3.6 & 0.0 & 35 & 3 & 15 & 3 \\
\midrule
Diffusion & MoDE & 73 & 75.3 & 3 & 13.7 & 1.2 & 49 & 6 & 13 & 5 \\
 & 3D Diffuser Actor & 38 & 52.6 & 16 & 34.2 & 3.6 & 12 & 8 & 14 & 4 \\
\midrule
Predictive & GR-1 & 57 & 73.7 & 3 & 8.8 & 1.0 & 37 & 5 & 8 & 7 \\
 & Seer-Large & 72 & 83.3 & 3 & 15.3 & 1.2 & 52 & 8 & 10 & 2 \\
\midrule
VLM+LSTM & RoboFlamingo & 46 & 65.2 & 2 & 17.4 & 0.5 & 25 & 5 & 13 & 3 \\
\midrule[\heavyrulewidth]
\multicolumn{11}{@{}l}{\textcolor{phys}{\textbf{Receptacle relocation (block receptacles)}}} \\
\cmidrule{1-11}
VLA & FLOWER & 50 & 86.0 & 4 & 4.0 & 1.2 & 40 & 3 & 6 & 1 \\
 & X-VLA & 45 & 75.6 & 8 & 17.8 & 1.6 & 30 & 4 & 7 & 4 \\
 & $\pi_{0.5}$ & 49 & 87.8 & 0 & 4.1 & 0.0 & 41 & 2 & 3 & 3 \\
 & UniVLA & 34 & 61.8 & 2 & 2.9 & 0.2 & 21 & 0 & 10 & 3 \\
 & OpenVLA-OFT & 29 & 58.6 & 0 & 6.9 & 0.0 & 16 & 1 & 4 & 8 \\
\midrule
Diffusion & MoDE & 51 & 66.7 & 2 & 9.8 & 0.5 & 33 & 1 & 9 & 8 \\
 & 3D Diffuser Actor & 25 & 28.0 & 17 & 16.0 & 2.8 & 5 & 2 & 14 & 4 \\
\midrule
Predictive & GR-1 & 30 & 76.7 & 2 & 0.0 & 0.7 & 23 & 0 & 4 & 3 \\
 & Seer-Large & 48 & 77.1 & 4 & 6.2 & 0.8 & 36 & 1 & 9 & 2 \\
\midrule
VLM+LSTM & RoboFlamingo & 34 & 41.2 & 2 & 5.9 & 0.3 & 14 & 0 & 15 & 5 \\
 
\bottomrule
\end{tabular}\end{adjustbox}
\end{table*}

\subsection{Oracle flush and disagreement gate}
\label{sec:supp_gate}
\paragraph{Design.} The flush arm of \cref{sec:supp_arms} is an oracle, because it empties the queue at the true moment of the change. The gate must decide by itself when to do so. Every two steps the policy is queried again on the current observation. If the first five actions of the fresh chunk differ from the queued ones by more than $\tau$ (mean L2 over the arm dimensions), the queue is replaced. $\tau{=}0.872$ is the 99th percentile of the disagreement on 9,759 probes without a change, so the gate fires on about 1\% of probes when nothing changes. It does not use any flag that marks the change. This is a policy-side version of the latent-disagreement gate \lmax proposes.

\paragraph{Effects.} \Cref{tab:supp_gate} gives FLOWER's recovery over the unmodified \Mid. On 200 \Base chains, the average chain length is \lite{4.550} with the gate against \lite{4.545} without it, so these chains show no cost in success, though this comparison is not a test of equivalence; wall time is about five times higher because of the extra queries. The interval of the flush is wide, so the quarter of the physical loss it recovers is a point estimate, not a ceiling for replanning. FLOWER also holds little of its timing effect in its queue (\cref{tab:decomp}), which limits what this test can show.

\begin{table}[!tp]
\caption{The oracle flush and the disagreement gate for FLOWER on Lite. Recovery is the success of each variant minus that of the unmodified \Mid, in pp with 95\% intervals.}
\label{tab:supp_gate}
\centering
\footnotesize
\setlength{\tabcolsep}{3pt}
\begin{adjustbox}{max width=\columnwidth}
\begin{tabular}{@{}lrrrcc@{}}
\toprule
\tableheader Channel & $n$ & \tableheader \Base & \tableheader \Mid & \tableheader Oracle flush & \tableheader Gate \\
\midrule
Physical & 130 & 97.7 & 89.2 & \tci{+2.3}{$-$2.3}{+7.1} & \tci{+1.5}{$-$2.4}{+5.6} \\
Perceptual & 272 & 95.2 & 91.2 & \tci{+0.0}{$-$2.3}{+2.3} & \tci{$-$1.8}{$-$3.8}{+0.0} \\
 
\bottomrule
\end{tabular}\end{adjustbox}
\end{table}

\section{Changes to the Robot and the Instruction}
\label{sec:supp_newch}

\subsection{Changes to the robot}
\label{sec:supp_embodiment}
Arm push physically displaces the arm: joints are reset to the inverse kinematics of the displaced pose with zero velocity, the relative controller's hold point moves with it, and a held block is carried along. This makes the push act on policies that output absolute poses as well. Grasp loss releases a held block during transport. \Cref{tab:supp_embodiment} gives both changes on \Track for every policy. Arm push changes success by \PushTot{} \PushTotCI{} points on average over the ten policies, with chains resampled jointly, and its $\dtim$ over all assigned cases is \PushTim{} \PushTimCI{}: \Mid is within \maxa{11} points of \Base for every policy, and $\dtim$ is at most \maxa{6} points either way. Grasp loss is realized only when a block is held at the onset (\maxa{11--26} of \maxa{49} cases per policy; the rest count with their \Base outcome), and no policy loses more than \maxa{6.1} points over its assigned cases (\GraspTot{} \GraspTotCI{} on average). On realized cases only, arm push costs \maxa{$-$5.6} \maxa{[$-$8.8, $-$2.7]} points and grasp loss \maxa{$-$4.9} \maxa{[$-$8.5, $-$1.7]}, with timing effects of \maxa{$-$0.2} \maxa{[$-$4.6, +4.0]} and \maxa{$-$1.4} \maxa{[$-$6.1, +2.7]} (equal-weight means over the ten policies, chains resampled jointly; exploratory).

\begin{table*}[!tp]
\caption{The two changes to the robot on the held-out cases, over all assigned cases. Real.\ gives the cases with the change realized out of those assigned; success is in \%, and $\dtim$, \Mid$-$\Pre, in pp with its 95\% bootstrap interval over chains; an interval of [0, 0] means that no case changed outcome between \Pre and \Mid. Under an arm push, \Mid is within 11 points of \Base for every policy.}
\label{tab:supp_embodiment}
\centering
\footnotesize
\setlength{\tabcolsep}{1pt}
\begin{adjustbox}{max width=\textwidth}
\begin{tabular*}{\textwidth}{@{\extracolsep{\fill}}l rrrrc rrrrc@{}}
\toprule
& \multicolumn{5}{c}{\tableheader Arm push} & \multicolumn{5}{c}{\tableheader Grasp loss} \\
\cmidrule(lr){2-6}\cmidrule(l){7-11}
\tableheader Policy & Real. & \Base & \Pre & \Mid & $\dtim$ & Real. & \Base & \Pre & \Mid & $\dtim$ \\
\midrule
FLOWER & 85/100 & 89 & 86 & 87 & \tci{+1.0}{$-$4.1}{+6.1} & 26/49 & 94 & 94 & 94 & \tci{+0.0}{+0.0}{+0.0} \\
X-VLA & 87/100 & 87 & 85 & 87 & \tci{+2.0}{$-$3.2}{+7.3} & 22/49 & 73 & 69 & 73 & \tci{+4.1}{+0.0}{+10.4} \\
$\pi_{0.5}$ & 79/100 & 82 & 77 & 76 & \tci{$-$1.0}{$-$7.0}{+4.7} & 19/49 & 63 & 63 & 61 & \tci{$-$2.0}{$-$6.4}{+0.0} \\
UniVLA & 73/100 & 74 & 73 & 72 & \tci{$-$1.0}{$-$7.1}{+5.1} & 21/49 & 71 & 71 & 69 & \tci{$-$2.0}{$-$8.3}{+4.2} \\
OpenVLA-OFT & 65/100 & 66 & 57 & 58 & \tci{+1.0}{$-$7.1}{+9.1} & 13/49 & 47 & 45 & 47 & \tci{+2.0}{+0.0}{+6.4} \\
MoDE & 84/100 & 85 & 81 & 81 & \tci{+0.0}{$-$6.1}{+6.1} & 23/49 & 84 & 80 & 78 & \tci{$-$2.0}{$-$10.6}{+6.1} \\
3D Diffuser Actor & 55/100 & 41 & 38 & 44 & \tci{+6.0}{$-$2.0}{+13.0} & 11/49 & 33 & 31 & 29 & \tci{$-$2.0}{$-$9.8}{+4.2} \\
GR-1 & 63/100 & 67 & 61 & 60 & \tci{$-$1.0}{$-$7.6}{+5.9} & 13/49 & 67 & 67 & 65 & \tci{$-$2.0}{$-$6.0}{+0.0} \\
Seer-Large & 85/100 & 87 & 82 & 76 & \tci{$-$6.0}{$-$14.4}{+2.9} & 22/49 & 82 & 82 & 82 & \tci{+0.0}{+0.0}{+0.0} \\
RoboFlamingo & 64/100 & 52 & 51 & 47 & \tci{$-$4.0}{$-$10.8}{+3.0} & 12/49 & 43 & 43 & 43 & \tci{+0.0}{+0.0}{+0.0} \\
 
\bottomrule
\end{tabular*}\end{adjustbox}
\end{table*}

\subsection{Revised instructions}
\label{sec:supp_intent}
Instruction revision swaps the goal of the instruction for another task feasible in the same scene, chosen with CALVIN's symbolic conditions. Success is judged on the new task. All ten policies ran Lite's 30 revision cases (\cref{tab:supp_intent}; exploratory). Given from the start of the instruction, the new task is completed in \lite{77--100\%} of cases. Given mid-motion, it is completed \lite{9.4} points less often on average (\Mid$-$\Pre of \lite{$-$9.4} \lite{[$-$16.4, $-$3.6]}), and the cost is uneven: six policies stay within \lite{9} points of \Pre, while 3D Diffuser Actor, GR-1, Seer-Large and UniVLA lose \lite{14} to \lite{25}. With 13 to 29 realized cases per policy, no single policy's exact McNemar test falls below 0.05. The case-bootstrap intervals of Seer-Large and UniVLA exclude zero while their exact two-sided tests give $p$ = 0.0625 and 0.125: all of their 5 and 4 discordant cases go the same way, and an approximate percentile interval on so few discordant pairs need not agree with an exact test. The 30 revision cases lie on 29 chains, so resampling cases is nearly the same as resampling chains. The pooled interval places \Mid below \Pre. Other pooled timing effects have negative point estimates too, and the physical one at the first position also has an interval below zero (\cref{tab:supp_position}). The losses are rarely a policy carrying on with the old task: the original task is completed in \lite{9} of the 230 \Mid runs, but in \lite{5} of them the policy then completes the new task as well, and only \lite{4} of the \lite{38} failed \Mid runs finish the old task instead.
\begin{table}[!tp]
\caption{Revised instructions on Lite (exploratory). $n$ counts the cases realized in both arms. Success (\%) is on the revised task, and \Mid$-$\Pre (pp) comes with its 95\% case bootstrap interval; the bottom row is the equal-weight mean over the policies, resampling cases jointly across them, and its $n$ and Orig.\ are sums. Orig.\ counts the runs that also complete the original task, in \Pre and \Mid. Pooled over the policies, \Mid completes the revised task less often than \Pre.}
\label{tab:supp_intent}
\centering
\footnotesize
\setlength{\tabcolsep}{2.4pt}
\begin{adjustbox}{max width=\columnwidth}
\begin{tabular}{@{}llrrrcc@{}}
\toprule
\tableheader Family & \tableheader Policy & $n$ & \tableheader \Pre & \tableheader \Mid & \tableheader \Mid$-$\Pre & \tableheader Orig. \\
\midrule
VLA & FLOWER & 29 & 96.6 & 93.1 & \tci{$-$3.4}{$-$10.7}{+0.0} & 0 / 0 \\
 & X-VLA & 28 & 89.3 & 89.3 & \tci{+0.0}{$-$10.4}{+10.3} & 0 / 0 \\
 & $\pi_{0.5}$ & 24 & 95.8 & 91.7 & \tci{$-$4.2}{$-$19.0}{+8.7} & 0 / 0 \\
 & UniVLA & 23 & 95.7 & 78.3 & \tci{$-$17.4}{$-$33.3}{$-$4.2} & 0 / 0 \\
 & OpenVLA-OFT & 23 & 82.6 & 73.9 & \tci{$-$8.7}{$-$26.1}{+8.3} & 0 / 0 \\
\midrule
Diffusion & MoDE & 28 & 92.9 & 89.3 & \tci{$-$3.6}{$-$18.5}{+11.5} & 0 / 1 \\
 & 3D Diffuser Actor & 14 & 92.9 & 78.6 & \tci{$-$14.3}{$-$36.4}{+0.0} & 0 / 0 \\
\midrule
Predictive & GR-1 & 20 & 95.0 & 70.0 & \tci{$-$25.0}{$-$47.8}{+0.0} & 1 / 2 \\
 & Seer-Large & 28 & 100.0 & 82.1 & \tci{$-$17.9}{$-$33.3}{$-$4.0} & 1 / 5 \\
\midrule
VLM+LSTM & RoboFlamingo & 13 & 76.9 & 76.9 & \tci{+0.0}{$-$21.4}{+22.2} & 1 / 1 \\
\midrule
\multicolumn{2}{@{}l}{\textbf{All ten policies}, equal weight} & 230 & & & \tci{$-$9.4}{$-$16.4}{$-$3.6} & 3 / 9 \\
 
\bottomrule
\end{tabular}\end{adjustbox}
\end{table}

\section{Audit Trail and Pre-specified Analysis Plan}
\label{sec:supp_auditprereg}

\subsection{Corrections found by audit}
\label{sec:supp_audit}
We audited our own pipeline repeatedly, and every correction was logged with its date and quantified (\cref{tab:supp_audit}). The first six corrections predate the freeze of the analysis plan on 2026-11-04 (\cref{sec:supp_prereg}), so the frozen tests already include them; the last two, on 2026-11-08, corrected exploratory Lite results after Max-A had run. Runs affected by the first two were rerun. The 16 cases of the third, all run before Max-A, keep the validity decided when they ran, as recorded at the time; excluding FLOWER's six core cases among them moves its Lite difference between channels from +8.9 to +9.9 points. The largest artifact was a $-$63-point arm-push loss at the transport onset, which disappeared once the reference scene was updated correctly.

\begin{table}[!htbp]
\caption{Corrections found by auditing our own pipeline, with the date each was logged (2026). The first six predate the freeze on Nov.~4; the last two corrected exploratory Lite results after Max-A.}
\label{tab:supp_audit}
\centering
\footnotesize
\setlength{\tabcolsep}{2pt}
\begin{tabularx}{\linewidth}{@{}lYY@{}}
\toprule
\tableheader Date & \tableheader Issue & \tableheader Effect after correction \\
\midrule
Oct.~30 & Reference scene not updated for some moved objects & Affected runs rerun; physical $\dtim$ roughly halved; FLOWER arm push at transport $-$63 $\rightarrow$ $-$1 \\
Oct.~30 & Onset on the final action of an instruction & Such cases count as unreached; affected runs rescored with the reruns above \\
Oct.~30 & Checks before evaluation read contacts before a collision pass & 13 of 867 accepted FLOWER, 1 of 186 MoDE and 2 of 190 X-VLA cases, all run before Max-A, would be rejected; they keep their run-time validity, and Max-A uses the corrected check \\
Oct.~30 & A resumed run could skip a failed check & Failures made final: the case is unrealizable with its \Base outcome, and arms it completed are not used \\
Oct.~30 & Missing \Base outcomes in early result files & Recovered from saved summaries with checksums \\
Oct.~30 & Intervals for the next-instruction cost resampled cases, not chains & Bootstrap resamples whole chains, as in the frozen test \\
Nov.~8 & Gate table read results computed before the reruns with the corrected reference scene & FLOWER flush recovery $+3.0$ of 12.1 $\rightarrow$ $+2.3$ of 8.5 points \\
Nov.~8 & Behaviour class ``stale-failed'' counted every failure after moving, not only reaches to the old pose & X-VLA (Lite) 15 $\rightarrow$ 5 of 16 failures reach the old pose; class renamed \\
\bottomrule
\end{tabularx}
\end{table}

\subsection{Confirmatory hypotheses for Max}
\label{sec:supp_prereg}
The hypotheses, tests and analysis code were fixed in a dated internal record on 2026-11-04, before the first Max-A case ran (\cref{tab:supp_prereg}); the record is not a public registration. It names the ten policies, OpenVLA-OFT included, whose Lite run finished the next day. Max-A is the first third of Max in each combination of track, change, variant and position, in the order of the case list (ascending chain number, itself a random draw), without the Lite cases: per policy, 1,526 held-out cases, 1,377 core and 149 embodiment; 887 of the core cases are perceptual and also receive the rewind arm. As the frozen code does, H1 and H2 count five changes as physical: target relocation, receptacle relocation, obstacle insertion, arm push and grasp loss. On the core changes alone (exploratory), the physical $\dtim$ is \maxa{$-$0.8} \maxa{[$-$2.5, +0.9]} and the difference between channels \maxa{+6.2} \maxa{[+4.2, +8.1]}.

The frozen code computes H1 and H2 on the held-out cases, on them without 3D Diffuser Actor, and on the union of Lite and Max-A, and the per-policy rows of H3 and H6. A reporting script, written on 2026-11-07 after the freeze, while Max-A was still running and once interim results for eight policies were in, adds the Holm-adjusted per-policy $p$ for H1, H3 and H6 and the H3 equivalence test, from the same bootstrap (\cref{sec:supp_metrics}); it has not changed since. The plan also reports H3 and H6 on the union, which we computed for this supplement with the same code. On the union, H6 holds for all ten policies (\maxa{$-$16.0} to \maxa{$-$5.3} points, Holm-adjusted $p\le0.005$). H3 has Lite rewind arms only for FLOWER, MoDE and X-VLA, and the union keeps their held-out verdicts (90\% intervals of the residual \maxa{[$-$3.8, $-$0.3]}, \maxa{[$-$1.1, +2.2]} and \maxa{[$-$2.0, +1.0]}). The held-out result is the one that confirms. Power uses the Lite standard error divided by $\sqrt{\text{Max-A}/\text{Lite}}\approx\sqrt{3.1}$ with one-sided $\alpha{=}0.05$: about 0.8 per policy for a 9-point difference between channels and 0.25 for 5 points. H4 and H5 need the sweeps over strength and onset phase, which Max-A does not run; they were moved to exploratory at the freeze and are reported on Lite. Results are in \cref{tab:prereg} and \cref{tab:supp_prereg_res}. The plan states the equivalence part of H3 for each policy as a 90\% interval of the residual inside $\pm$3 points, a TOST at $\alpha{=}0.05$, and separately asks for Holm correction within each family of per-policy tests without saying whether the equivalence tests form such a family. The reporting script Holm-corrects the positive gain over the ten policies and reads the equivalence from the uncorrected 90\% interval, which gives MoDE, X-VLA and Seer-Large; Holm-correcting the equivalence $p$-values as well leaves only X-VLA. The budget-matched residual (\cref{sec:supp_arms}) is a post-hoc check and does not change any verdict.

\begin{table*}[!tp]
\caption{Each test fixed in advance, policy by policy, on the held-out cases. For H1 the table gives the timing effect by channel, their difference (pp) and the one-sided $p$ for the difference after Holm correction. For H3 it gives the verdict on the perceptual core from \cref{tab:decomp}: \emph{yes} where both parts of the criterion hold and \emph{gain only} where only \Mid$-$rewind $>0$ does. For H6 it gives the cost to the next instruction, $\Delta$ (pp), and its $p$ after Holm correction. Effects come with their 95\% bootstrap intervals over chains.}
\label{tab:supp_prereg_res}
\centering
\footnotesize
\setlength{\tabcolsep}{1pt}
\begin{adjustbox}{max width=\textwidth}
\begin{tabular*}{\textwidth}{@{\extracolsep{\fill}}lcccrccr@{}}
\toprule
& \multicolumn{4}{c}{\tableheader H1} & \tableheader H3 & \multicolumn{2}{c}{\tableheader H6} \\
\cmidrule(lr){2-5}\cmidrule(lr){6-6}\cmidrule(l){7-8}
\tableheader Policy & \tableheader Perc. & \tableheader Phys. & \tableheader Difference & $p_\text{Holm}$ & & $\Delta$ & $p_\text{Holm}$ \\
\midrule
FLOWER & \tci{+5.1}{+3.3}{+6.9} & \tci{+2.3}{$-$0.5}{+5.1} & \tci{+2.7}{$-$0.4}{+6.1} & 0.139 & gain only & \tci{$-$6.7}{$-$8.7}{$-$4.8} & $<$0.001 \\
X-VLA & \tci{+6.8}{+4.4}{+9.3} & \tci{+1.0}{$-$2.7}{+4.7} & \tci{+5.8}{+1.3}{+10.3} & 0.032 & yes & \tci{$-$8.8}{$-$11.5}{$-$6.0} & $<$0.001 \\
$\pi_{0.5}$ & \tci{+1.6}{$-$0.5}{+3.9} & \tci{+0.3}{$-$3.1}{+3.8} & \tci{+1.3}{$-$2.8}{+5.4} & 0.272 & gain only & \tci{$-$6.3}{$-$9.1}{$-$3.6} & $<$0.001 \\
UniVLA & \tci{+5.0}{+1.7}{+8.3} & \tci{$-$2.9}{$-$7.5}{+1.6} & \tci{+7.9}{+2.5}{+13.4} & 0.021 & gain only & \tci{$-$16.8}{$-$20.1}{$-$13.4} & $<$0.001 \\
OpenVLA-OFT & \tci{+5.6}{+1.7}{+9.4} & \tci{+1.0}{$-$4.0}{+6.1} & \tci{+4.6}{$-$1.9}{+11.0} & 0.165 & gain only & \tci{$-$12.7}{$-$18.0}{$-$7.5} & $<$0.001 \\
MoDE & \tci{+5.1}{+2.8}{+7.6} & \tci{$-$1.0}{$-$5.7}{+3.5} & \tci{+6.1}{+1.1}{+11.2} & 0.052 & yes & \tci{$-$7.0}{$-$9.7}{$-$4.4} & $<$0.001 \\
3D Diffuser Actor & \tci{+7.8}{+2.9}{+12.8} & \tci{$-$5.0}{$-$12.2}{+2.1} & \tci{+12.8}{+5.0}{+20.6} & 0.004 & gain only & \tci{$-$6.8}{$-$13.3}{+0.0} & 0.026 \\
GR-1 & \tci{+6.6}{+3.8}{+9.4} & \tci{+1.0}{$-$3.3}{+5.4} & \tci{+5.6}{+0.4}{+10.8} & 0.076 & gain only & \tci{$-$10.8}{$-$14.1}{$-$7.4} & $<$0.001 \\
Seer-Large & \tci{+4.4}{+2.6}{+6.3} & \tci{$-$1.9}{$-$4.9}{+1.1} & \tci{+6.4}{+2.8}{+10.0} & $<$0.001 & yes & \tci{$-$5.0}{$-$7.2}{$-$2.9} & $<$0.001 \\
RoboFlamingo & \tci{+5.5}{+1.9}{+9.1} & \tci{$-$1.0}{$-$5.7}{+3.7} & \tci{+6.6}{+0.7}{+12.4} & 0.076 & gain only & \tci{$-$7.2}{$-$11.7}{$-$2.8} & 0.001 \\
 
\bottomrule
\end{tabular*}\end{adjustbox}
\end{table*}

\begin{table}[!htbp]
\caption{The confirmatory hypotheses for Max, as worded at the freeze. H1 is primary. In H4 the axis \Pre$-$\Base is signed, so a benefit that grows with the damage is a negative slope on it, and the frozen wording, kept here, has the opposite sign to its heading. H4 and H5 were set aside as exploratory at the freeze, and neither is tested on Max-A.}
\label{tab:supp_prereg}
\centering
\footnotesize
\setlength{\tabcolsep}{2pt}
\begin{tabularx}{\linewidth}{@{}lY@{}}
\toprule
\# & \tableheader Hypothesis and test \\
\midrule
H1 & A mid-motion arrival helps perceptual changes more than physical ones: pooled difference between channels $>0$; per policy with Holm correction. \\
H2 & Physical $\dtim$ alone, two-sided; sign reported whatever it is; within $\pm$1 point is reported as no detectable effect. \\
H3 & The perceptual benefit comes from motion already under way: \Mid$-$rewind $>0$ and the residual within $\pm$3 points, per policy. \\
H4 & The benefit tracks degradation: positive slope of \Mid$-$\Pre on \Pre$-$\Base across strengths. \\
H5 & Perceptual loss occurs before contact: mean \Mid$-$\Base over start, approach and pre-grasp below the mean over contact, transport and place. \\
H6 & Cost to the next instruction: success on the next instruction is lower after the change, given that the affected instruction was completed in both arms. \\
\bottomrule
\end{tabularx}
\end{table}

\paragraph{Decision rules.} If H1 fails pooled, the paper is reported as a benchmark and measurement study, with H3--H6 as properties of the policies tested. If H1 holds pooled but fails for a policy, that policy is reported as an exception with its profile across changes. If the H3 residual is not within $\pm$3 points for a policy, the reading that the benefit comes from motion already under way does not hold for that policy. If H1 holds pooled only with 3D Diffuser Actor, this is reported as such. On Max-A, H1 holds pooled, at \maxa{+6.0} \maxa{[+4.1, +7.9]} held out, \maxa{+5.2} \maxa{[+3.4, +7.0]} without 3D Diffuser Actor and \maxa{+6.8} \maxa{[+5.2, +8.4]} on the union; per policy it is significant only for X-VLA, Seer-Large, UniVLA and 3D Diffuser Actor, so the other policies are reported with their profiles across changes (\cref{tab:supp_events}) and no claim of generality is made per policy. With the uncorrected interval, the H3 residual passes the $\pm$3-point equivalence test only for X-VLA, MoDE and Seer-Large, so under the rule above the reading is confirmed for these three (for X-VLA alone under the stricter reading above). For the other seven, the first half of H3 holds, since rewinding removes the benefit for all ten, but the rewound run falls short of \Pre by as much as \maxa{12.4} points. The main text therefore states that reading only for the three, and reports the rewind result for the other seven as an observation. The step-budget check of \cref{sec:supp_arms}, which brings every residual except UniVLA's within \maxa{1.8} points of zero, is post hoc and does not change any verdict. For H2, the pooled physical $\dtim$ is \maxa{$-$0.6} \maxa{[$-$2.2, +0.9]}. Its point estimate lies inside $\pm$1 point, which the plan reports as no detectable physical timing effect; the interval does not establish equivalence within $\pm$1 point. Without 3D Diffuser Actor it is \maxa{$-$0.1} \maxa{[$-$1.7, +1.4]}, and on the union \maxa{$-$1.0} \maxa{[$-$2.4, +0.3]}, just outside the band before rounding ($-$1.01). No single policy has a physical interval that excludes zero. H6 holds for all ten policies after Holm correction, with costs from \maxa{$-$16.8} to \maxa{$-$5.0} points. Any change to these tests after the first Max result is dated, and results are reported under both versions. Two analyses depart from the plan without replacing a confirmatory test: the reporting script described above, and the hazard model, which resamples chains instead of fitting the planned chain random effect (\cref{sec:supp_chain}).

\section{Discussion and Future Directions}
\label{sec:supp_discussion}

\subsection{Interpreting progress on the benchmark}
\label{sec:supp_interp}
A policy can improve on \maxbench by losing less in the changed world ($\dconf$), by turning the motion already under way into an advantage for perceptual changes, or by abandoning it quickly for physical ones ($\dtim$). These are different abilities, and a single gap would conflate them. A response method should be judged on all three arms and on \Base, since a method that replans often may protect against physical changes and give up the perceptual benefit.

\subsection{Scope and limitations}
\label{sec:supp_limits}
\maxbench is a simulation benchmark with CALVIN's rendering, contact model and task set. The channel grouping is a design choice; obstacle insertion, for example, is both visible and physical. Comparisons between channels use cases realized in both arms, which depend on each policy's \Base trajectory and on checks that reject far more physical than perceptual cases (\cref{sec:supp_outcomes}), so they are conditional on realization rather than a pure causal isolation of onset time. Comparisons between phases and between positions differ in task mix. The rewind decomposition and the analyses of the next-instruction cost, hazard, position and behaviour cover all ten policies on Max-A, but the sweeps over strength and onset phase, the decomposition for physical changes and the disagreement gate cover three policies (one for the gate) on sets outside Max-A and are exploratory, as are the revised-instruction and null-control cases, which run on Lite. The policies differ in training data, scale and serving protocol, so family comparisons describe released checkpoints. We evaluate one training seed per policy on Max-A; MoDE's three released seeds also run Lite. No policy is run on all 7,828 cases: the confirmatory results use Max-A, and the rest of Max is released for future evaluation. A human audit of 400 realized FLOWER cases from Max-A, drawn mostly from the physical changes most likely to block a task, judged 398 feasible and none infeasible from the state just after the change (feasible share 100\%, 95\% interval 99.0--100\%, over the 398 definite judgments); two cases were unclear because CALVIN's instructions do not say what contact between objects is allowed (\cref{sec:supp_humanaudit}). The audit covers one policy's onset states and cases already accepted by the automated checks, and the annotator saw both cameras, whereas UniVLA and OpenVLA-OFT see only the static one. The held-out cases share no case with Lite, but 204 of their 385 chains also carry Lite cases; on the other 181, the difference between channels is smaller, with an interval still above zero (\cref{sec:supp_lite_agree}).

\subsection{Learning and evaluating responses to change}
\label{sec:supp_future}
The results suggest testing a selective response: keep the progress made before the change through a corrupted view, and drop it when the world has really moved. A gate that conditions on whether a change is task-relevant, rather than on raw disagreement, is the natural next method, evaluated against the oracle flush and rewind arms measured here. Training on generated trajectories with changes, as scene editing already does for static distractors~\cite{pakdamansavoji2026nice}, compound and recurrent changes, and transfer to real robots remain open.

\subsection{Compute and release}
\label{sec:supp_compute}
All runs use NVIDIA L4 GPUs on Google Colab. We will release the cases, the onset calibration, the policy adapters, the runner, the per-case results and the analysis code.

\end{document}